\documentclass[nohyperref]{article}

\usepackage{bbm}

\usepackage{listings}

\newcommand{\bx}{\mathbf{x}}

\newcommand{\bc}{\mathbf{c}}

\newcommand{\bs}{\mathbf{s}}

\newcommand{\bI}{\mathbf{I}}

\newcommand{\bz}{\mathbf{z}}

\newcommand{\br}{\mathbf{r}}

\newcommand{\cC}{\mathcal{C}}

\newcommand{\cS}{\mathcal{S}}

\usepackage{dsfont}
\usepackage{enumitem}

\usepackage{soul}
\usepackage{subcaption}
\usepackage{multirow}
\usepackage{booktabs}
\usepackage{csquotes}
\usepackage{graphicx}
\usepackage{mathtools}

\newcommand{\image}{\mathbf{x}}
\newcommand{\mask}{\mathbf{m}}

\newcommand{\rep}{\mathbf{r}}
\newcommand{\weight}{\mathbf{w}}
\newcommand{\dataset}{\mathcal{D}}

\newcommand{\class}{i}
\newcommand{\feature}{j}

\newcommand{\causal}{\mathcal{C}}
\newcommand{\spurious}{\mathcal{S}}
\newcommand{\allclasses}{\mathcal{Y}}
\newcommand{\oldclasses}{\mathcal{T}}

\newcommand{\causalmask}{\mathcal{M}^{c}}
\newcommand{\spuriousmask}{\mathcal{M}^{s}}

\newcommand{\RCS}{\mathrm{RCS}}
\newcommand{\CFV}{\mathrm{CFV}}
\newcommand{\SFV}{\mathrm{SFV}}
\newcommand{\causalaccabv}{\mathrm{acc^{(\cC)}}}
\newcommand{\spuriousaccabv}{\mathrm{acc^{(\cS)}}}

\newcommand{\fm}{neural activation map}

\newcommand{\importance}{\mathrm{IV}}

\newcommand{\zeros}{\boldsymbol{0}}
\newcommand{\ones}{\boldsymbol{1}}

\usepackage{enumitem}

\setlist[description]{leftmargin=*,labelindent=*}
\setlist[itemize]{leftmargin=*}
\setlist[enumerate]{leftmargin=*}
\usepackage{microtype}
\usepackage{graphicx}
\usepackage{booktabs} % for professional tables

\usepackage{hyperref}

\usepackage[accepted]{icml2022}

\usepackage{amsmath}
\usepackage{amssymb}
\usepackage{mathtools}
\usepackage{amsthm}
\usepackage[capitalize,noabbrev]{cleveref}

\theoremstyle{plain}
\newtheorem{theorem}{Theorem}[section]

\theoremstyle{definition}
\newtheorem{definition}[theorem]{Definition}

\theoremstyle{remark}

\usepackage[textsize=tiny]{todonotes}

\begin{document}

\twocolumn[
\icmltitle{Core Risk Minimization using Salient ImageNet}

\icmlsetsymbol{equal}{*}

\begin{icmlauthorlist}
\icmlauthor{Sahil Singla}{equal,umd}
\icmlauthor{Mazda Moayeri}{equal,umd}
\icmlauthor{Soheil Feizi}{umd}
\end{icmlauthorlist}

\icmlaffiliation{umd}{University of Maryland, College Park, United States}

\icmlcorrespondingauthor{Sahil Singla}{ssingla@umd.edu}
\icmlcorrespondingauthor{Mazda Moayeri}{mmoayeri@umd.edu}

\icmlkeywords{Machine Learning, ICML}

\vskip 0.3in
]

\pagestyle{plain}

\printAffiliationsAndNotice{\icmlEqualContribution}

\begin{abstract}

% \looseness=-1
% The use of spurious features by deep neural networks can make them unreliable in the wild. Recently, \citet{salientimagenet2021} introduced the Salient Imagenet dataset, annotating and localizing core (i.e. part of the object) and spurious (i.e. likely to co-occur with the object but not a part of it) features for ${\sim}52$k samples from $232$ classes of Imagenet. This dataset is useful for evaluating the reliance of models on spurious features, but its small size limits its utility for training large models. In this work, we first introduce the Salient Imagenet-1M dataset with soft masks localizing core and spurious features for \emph{all 1000 Imagenet classes}. We obtain core/spurious annotations of ${\sim}1.05$M images in the training and $226$k images in the test set. Next, we introduce a training paradigm called Core Risk Minimization (CoRM) whose objective is to ensure that the model predicts a class using the image regions containing core features for the class. We use our dataset to evaluate the reliance of many ($42$) Imagenet pretrained models on spurious features, finding: (i) transformers are more sensitive to spurious features compared to Convnets, (ii) zero-shot CLIP transformers are highly susceptible to spurious features. Then, we assess different approaches for solving relaxations of CoRM and achieve significantly higher ($+ 12\%$) \textit{core accuracy} (accuracy with non-core regions corrupted using noise) with no drop in clean accuracy compared to Empirical Risk Minimization trained models.
 
Deep neural networks can be unreliable in the real world especially when they heavily use spurious features for their predictions. Recently, \citet{salientimagenet2021} introduced the Salient Imagenet dataset by annotating and localizing core and spurious features of ${\sim}52$k samples from $232$ classes of Imagenet. While this dataset is useful for evaluating the reliance of pretrained models on spurious features, its small size limits its usefulness for training models. In this work, we first introduce the {\bf Salient Imagenet-1M dataset} with more than $1$ million soft masks localizing core and spurious features for \emph{all 1000 Imagenet classes}. Using this dataset, we first evaluate the reliance of several Imagenet pretrained models ($42$ total) on spurious features and observe: (i) transformers are more sensitive to spurious features compared to Convnets, (ii) zero-shot CLIP transformers are highly susceptible to spurious features. Next, we introduce a new learning paradigm called {\bf Core Risk Minimization (CoRM)} whose objective ensures that the model predicts a class using its core features. We evaluate different computational approaches for solving CoRM and achieve significantly higher ($+ 12\%$) \textit{core accuracy} (accuracy when non-core regions corrupted using noise) with no drop in clean accuracy compared to models trained via Empirical Risk Minimization. \looseness-1

\looseness=-1
\end{abstract}
\begin{figure}[t]
\centering
\includegraphics[trim=0cm 0cm 0cm 0cm, clip, width=\linewidth]{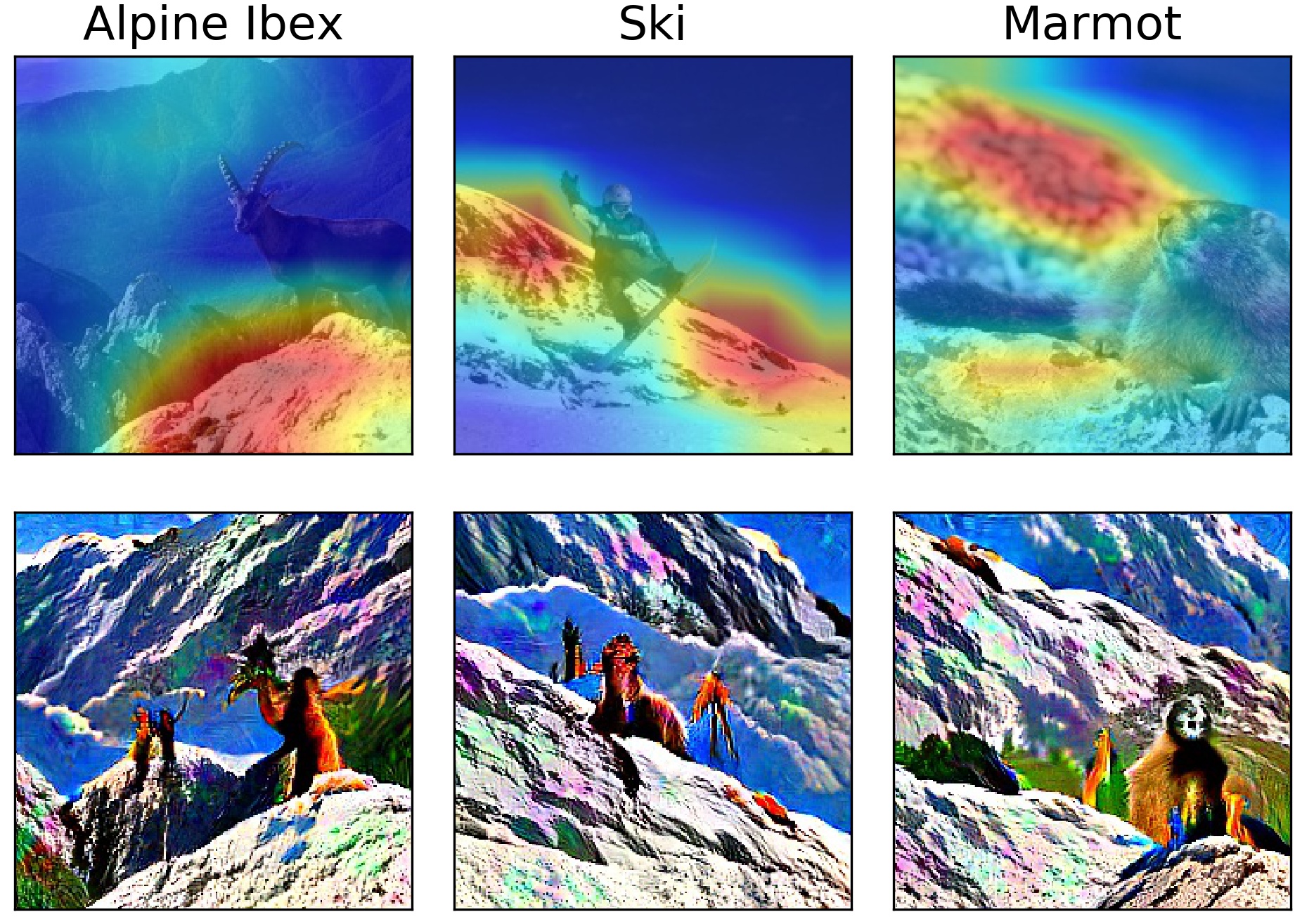}
\caption{({\bf Top}) Regions activating the feature highlighted in red. ({\bf Bottom}) Images perturbed to visually amplify the feature. The feature (likely {\it mountain}) is {\it spurious} for $3$ Imagenet classes: Alpine Ibex, Ski, Marmot. Surprisingly, the feature is also \textit{core (essential)} for $3$ different classes: Mountains, Valley, Volcano.} 
\label{fig:motivating}
\end{figure}

\section{Introduction}\label{introduction}
Decision making in high-stakes applications such as medicine, finance, autonomous driving, law enforcement and criminal justice is increasingly driven by deep learning models, thereby raising concerns about the trustworthiness and reliability of these systems in the real world. A root cause for the lack of reliability of deep models is their heavy reliance on {\it spurious} input features (i.e., features that are not essential to the true label) in their inferences. For example, \citet{deGrave2021aa} discovered that a convolutional neural network (CNN) trained to detect COVID-19 from chest radiographs uses spurious text-markers for its predictions. Similarly, \citet{Zech2018VariableGP} observed that a CNN trained to detect pneumonia from Chest-X rays had unexpectedly learned to identify particular hospital systems with
near-perfect accuracy (e.g. by detecting a hospital-specific metal token on the scan) with poor generalization to novel hospital systems. The list of such examples goes on \citep{terraincognita2018, causalconfusion2019, Bissoto2020DebiasingSL}.

% To highlight this issue, in Figure \ref{fig:motivating}, we show example of a spurious feature common across $3$ classes in Imagenet (Alpine Ibex, Ski, Marmot). Surprisingly, we also observe that this feature is core (essential) for $3$ other classes (Mountains, Valley, Volcano) showing that while deep models excel at pattern recognition, they can struggle in discerning which patterns are core for a class, at times incorrectly making use of spurious patterns recognized elsewhere.

To highlight the complexity of this issue, in Figure \ref{fig:motivating}, we show example of a spurious feature that is common across $3$ classes. Surprisingly, this feature is also core (essential) for $3$ other classes showing that while deep models excel at pattern recognition, they can struggle in discerning which patterns are core for a class, at times incorrectly making use of spurious patterns recognized elsewhere.

% The growing use of deep learning in sensitive applications such as medicine, autonomous driving, law enforcement and finance raises concerns about the trustworthiness and reliability of these systems in the real world. 

%Another limitation of deep models is that the standard Empirical Risk Minimization (ERM) paradigm of deep learning is brittle when the test distribution is different from the training distribution because of spurious features. Recently, \citet{arjovsky2020invariant} proposed a framework called Invariant Risk Minimization (IRM) to address this problem. IRM and its variants \cite{Krueger2021OutofDistributionGV, Xie2020RiskVP, mahajan2020domain} posit the existence of a feature embedder such that the optimal classifier on top of these features is the same for every environment (from which data can be drawn). However, \citet{rosenfeld2021the} show that IRM can fail catastrophically unless the test distribution is sufficiently similar to the training distribution. But in such a scenario, IRM would no longer be required; we would expect the standard ERM to perform just as well. 

The standard Empirical Risk Minimization (ERM) paradigm for training deep neural networks is brittle when the test distribution is different from the training distribution because of spurious features. Recently, \citet{arjovsky2020invariant} proposed a framework called Invariant Risk Minimization (IRM) to address this problem. IRM and its variants \cite{Krueger2021OutofDistributionGV, Xie2020RiskVP, mahajan2020domain} posit the existence of a feature embedder such that the optimal classifier on top of these features is the same for every environment from which data can be drawn. However, \citet{rosenfeld2021the} show that IRM can fail catastrophically unless the test distribution is sufficiently similar to the training distribution. In such cases, however, IRM would no longer be required; we would expect ERM to perform just as well. %Similarly, \citet{gulrajani2021in} show that when carefully implemented, ERM shows improved performance compared to many domain generalization methods. 

\begin{figure}[t]
\centering
\begin{subfigure}{0.9\linewidth}
\centering
\includegraphics[trim=0cm 0cm 10cm 0cm, clip, width=\linewidth]{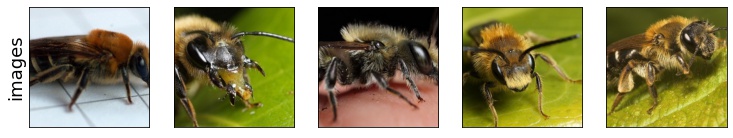}
\end{subfigure}\\
\begin{subfigure}{0.9\linewidth}
\centering
\includegraphics[trim=0cm 0cm 10cm 0cm, clip, width=\linewidth]{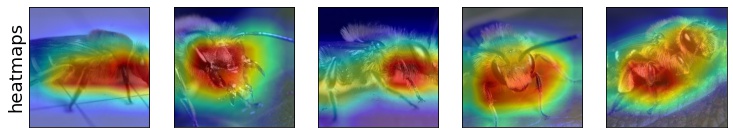}
\end{subfigure}\\
\begin{subfigure}{0.9\linewidth}
\centering
\includegraphics[trim=0cm 0cm 10cm 0cm, clip, width=\linewidth]{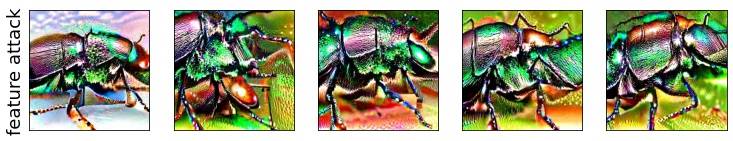}
\end{subfigure}
\caption{Example of \textbf{core feature} for the class \textbf{bee}. %MTurk worker description: 'focus is on legs and face of bee.'
}
\label{fig:core_example}
\end{figure}

The above methods use single-label supervision (i.e. an image is labeled only by class index). One can argue that such limited annotations may restrict the model's ability to learn from meaningful features in its predictions since the model is not given the information regarding which features are essential/core and which ones are redundant/spurious. Much of the prior work on discovering spurious features \cite{pandora1809, zhang2018manifold, chung2019slice, xiao2021noise} require expensive human-guided labeling of visual attributes, which is not scalable for datasets with a large number of classes and images such as Imagenet. However, \citet{salientimagenet2021} recently introduced an approach for discovering spurious features {\it at scale} using the neurons of robust models as visual attribute detectors. An application of their approach on a subset of Imagenet \cite{5206848} resulted in a dataset called {\it the Salient Imagenet} whose samples, in addition to class labels, are annotated by two sets of masks: {\it core masks} that highlight core/essential attributes (with respect to the true class) and {\it spurious masks} that highlight attributes co-occurring with the object but not a part of it. While valuable for evalation, the size of the Salient ImageNet dataset ($232$ classes%\footnote{$\oldclasses$ denotes the set of $232$ classes analyzed by \citet{salientimagenet2021}.}
, ${\sim}52$k images) limits its utility for training models.

In this work, we significantly expand the size of the Salient Imagenet dataset in two steps (Section \ref{sec:scaling_salient_imagenet}). First, for each class $\class \in \allclasses - \oldclasses$ \footnote{$\allclasses$ denotes the set of all $1000$ Imagenet classes and $\oldclasses$ the set of classes analyzed by \citet{salientimagenet2021}.} (the set of remaining $1000-232 = 768$ classes), we identify the top-$5$ penultimate layer neurons of a robust model highly predictive of $\class$. We then conduct a Mechanical Turk (MTurk) study for each of these (class, neuron) pairs to determine whether the neuron is core or spurious for the class, resulting in new annotations for $768 \times 5 = 3840$ pairs, for a total of $5000$ core/spurious annotations ($4370$ core and $630$ spurious), when combined with those of \citet{salientimagenet2021}. Second, for each (class=$\class$,\ feature=$\feature$) pair, we conduct another MTurk study to validate that the neural activation maps (NAMs) for these neurons highlight the same visual attribute for a large number of images. For images with label $\class$, we select the subset with the top-$260$ values of feature $\feature$. Next, we ask workers to validate whether the NAMs of $15$ images (randomly selected from $260$) focus on the same visual attribute. {\bf We validate that for $95.26\%$ of pairs, NAMs indeed focus on the same visual attribute.} The resulting dataset, called {\it Salient Imagenet-1M}, contains more than $1$ million core/spurious mask annotations.
%${\sim}1.05$M training and ${\sim}226k$ test images. 

\begin{figure}[t]
\centering
\begin{subfigure}{0.9\linewidth}
\centering
\includegraphics[trim=0cm 0cm 10cm 0cm, clip, width=\linewidth]{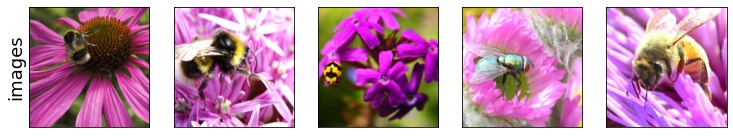}
\end{subfigure}\\
\begin{subfigure}{0.9\linewidth}
\centering
\includegraphics[trim=0cm 0cm 10cm 0cm, clip, width=\linewidth]{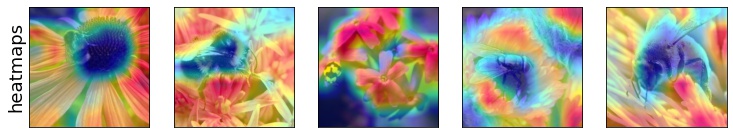}
\end{subfigure}\\
\begin{subfigure}{0.9\linewidth}
\centering
\includegraphics[trim=0cm 0cm 10cm 0cm, clip, width=\linewidth]{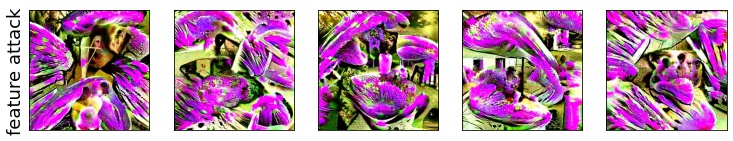}
\end{subfigure}
\caption{Example of \textbf{spurious feature} for the class \textbf{bee}. %MTurk worker description: 'focus is on petals of flower rather than bee.'
}
\label{fig:spurious_example}
\end{figure}

\looseness-1

Using our test set, we study a diverse set of pretrained Imagenet models and training paradigms ($42$ models total) in Section \ref{sec:benchmarking}. We find that: (i) transformers are more sensitive to spurious features compared to Convnets, (ii) adversarial training makes Resnets more sensitive to spurious features, (iii) zero-shot CLIP transformers are highly susceptible to spurious features, (iv) models with the same clean accuracy can have vastly different \textit{core accuracy} (i.e., accuracy when the non-core regions are corrupted using noise).

%We next aim to train models with reduced reliance on spurious features. 
We next aim to train models that {\it mainly use core features} in their predictions. To this end, we propose a training paradigm called {\it Core Risk Minimization (CoRM)} in Section \ref{sec:corm}. We note that in some cases, spurious features can be \textit{useful}. Consider the class ``matchstick'': the brightness of a flame (spurious) may obscure core features beyond recognition, but the flame itself provides evidence for the presence of a matchstick (see Appendix \ref{sec:motivation_appendix}). Thus, when core features are absent or not known to be in the image, we want our objective to reduce to standard ERM. Based on this desideratum, we formulate our CoRM objective as follows: \looseness-1
%We next aim to train models that mainly use core features in their predictions. To this end, we introduce a training paradigm called {\it Core Risk Minimization (CoRM)} in Section \ref{sec:corm}. The objective of CoRM is to ensure that the model predicts some class \textit{using mainly the core features} for the class. However, in some cases this may be too restrictive. For example, an image of ``flame" provides evidence for the presence of ``lighter" (even if lighter is not explicitly visible in the image). Thus, when the core features are absent or not known to be present in the image, we want our objective to reduce to the standard ERM objective. Based on this desideratum, we formulate our CoRM objective as follows:
\begin{align}
&\min_\theta\ \mathbb{E}_{(\bx,\bc,y)}\ \left[ \mathbb{E}_{ \bz \sim \mathbb{N}(\zeros, \sigma^{2}\bI)} \ \  \ell (f_{\theta}(\bx^{*}), y) \right]\label{eq: corm_eq_intro} \\
&\text{where } \bx^{*} = \bx + \bz \odot (\ones - \bc) \nonumber
\end{align}
Here, $\bc$ denotes the core mask (i.e., $\bc_{i,j}=1$ iff $\bx_{i,j}$ is a core pixel), $y$ is the ground truth label for $\bx$, $f_{\theta}(\bx^{*})$ denotes the logits, $\ell$ is the cross entropy loss and $\theta$ is the model parameters. Note that $\bx^{*}_{i,j} = \bx_{i,j}\ \ \forall\ i,j: \bc_{i,j} = 1$ which ensures that $\bx^*$ will have the same core features as $\bx$. However, the \textit{non-core}\footnote{We can also corrupt the spurious mask $\bs$ (instead of the non-core mask: $\ones-\bc$) in our formulation. However, we choose to use non-core masks because the Salient Imagenet dataset contains significantly larger number of core masks than spurious masks.} regions are corrupted using Gaussian noise with variance $\sigma^{2}$. Note that one can easily modify our CoRM formulation to have corruptions using other noise distributions or even adversarial corruptions on non-core regions. We use Gaussian noise because of its simplicity and because it easily allows us to control the degree of contextual information from non-core regions in the image using the parameter $\sigma$. When the core masks are unknown, we can set $\bc_{i,j} = 1\ \ (\forall\ i,j)$ so that $\bx^{*} = \bx$. This allows us to recover the standard ERM objective. Finally, note that our CoRM objective can also be used to train models when the core masks $\bc$ are soft (e.g., $0 \leq \bc_{i,j} \leq 1$ not $\in \{0, 1\})$.

In our CoRM objective \eqref{eq: corm_eq_intro}, we need to compute the inner expectation over the Gaussian distribution in a high dimensional space which can be difficult. Thus, to train models using the Salient Imagenet-1M dataset, we evaluate different variations for training using CoRM: (i) randomly adding Gaussian noise to the non-core regions during training, (ii) saliency regularization that penalizes the gradient norm in the non-core (i.e., $\ones-\bc$) regions. We show that by combining these two techniques, we achieve significantly higher $(+12\%)$ \emph{core accuracy}, while improving the clean accuracy compared to ERM trained models.

In summary, we make the following contributions:
%\vskip -1pt
\begin{itemize}
%\begin{itemize}[leftmargin=7pt]
%\setlength{\itemsep}{2pt}
%\setlength{\parskip}{-2pt}
%\setlength{\parsep}{-2pt}
\item We introduce the {\bf Salient Imagenet-1M} dataset with core and spurious masks for more than a million images in \textit{all Imagenet classes}.
%    \item \textit{Salient Imagenet-1M}: dataset with core/spurious masks for large number of images for \textit{all Imagenet classes}.
    
        \item We comprehensively study the reliance on spurious features for {\bf 42 pretrained Imagenet models} and training procedures, discovering interesting trends.
        \item We introduce {\bf Core Risk Minimization (CoRM)}, a new learning paradigm to train models that mainly rely on core features in their predictions, leading to strong empirical results ($+12\%$ core, $+0.58\%$ clean accuracy in Table \ref{tab:corm_results}).
        
%    \item \textit{Core Risk Minimization (CoRM)}: framework to train models to predict using core features with strong empirical results ($+12\%$ core, $+0.58\%$ clean accuracy in Table \ref{tab:corm_results}).
%    \item Comprehensive study of the reliance on spurious features for a large number of Imagenet pretrained models and training procedures ($42$ models in total)
\end{itemize}

% Second, we introduce a training paradigm called Core Risk Minimization (CoRM) where the objective (informally) is to ensure that the model predicts a class (say $\class$) \textit{using only the core features for} $\class$. However, in some cases this objective may be too restrictive. For example, an image of a flame provides evidence for the presence of lighter (though the lighter may not be explicitly visible in the image). Thus, when the core features are absent or not known to be present in an image with label $\class$, we want our objective to reduce to the standard Empirical Risk Minimization (ERM) objective. 

% And use spurious features only when 

%introduce a significantly expanded version of the Salient Imagenet dataset covering all $1000$ classes of Imagenet with $226,946$ images in the test set and $421,794$ images in the training set. 

% and (b) a risk minimization paradigm. 

%it is possible 

%i.e. the invariance of $\mathbb{E}\left[y | \Phi(\bx) \right]$. Follow-up works \cite{Krueger2021OutofDistributionGV, Xie2020RiskVP, mahajan2020domain} have proposed variations on this objective based on the strictly stronger desiredatum of the invariance of $p\left(y | \Phi(\bx) \right)$. 

%suggest penalizing the variance of the risks, while give the same objective but taking the square root of the variance. Many papers have suggested similar alternatives (Jin et al., 2020; Mahajan et al., 2020; Bellot & van der Schaar, 2020). 

\section{Related work}\label{sec:related_work}
\noindent \textbf{Interpretability}: Most of the existing works on post-hoc interpretability techniques focus on inspecting the decisions for a single image \citep{ZeilerF13, MahendranV15, DosovitskiyB15, YosinskiCNFL15, NguyenYC16, sanitychecks2018, zhou2018interpreting, chang2018explaining, olah2018the, Yeh2019OnT, carter2019activation, oshaughnessy2020generative, sturmfels2020visualizing, verma2020counterfactual}. These include saliency maps \citep{Simonyan2013DeepIC, Sundararajan2017AxiomaticAF, Smilkov2017SmoothGradRN, Singla2019UnderstandingIO}, class activation maps \citep{ZhouKLOT15,SelvarajuDVCPB16, Bau30071, Ismail2019AttentionDL, Ismail2020BenchmarkingDL}, surrogate models to interpret local decision boundaries such as LIME \citep{ribeiro2016}, methods to maximize neural activation values by optimizing input images \citep{NguyenYC14, mahendran15} and finding influential \citep{pangweikoh2021} or counterfactual inputs \citep{deviparikhcounterfactual19}. 

\noindent \textbf{Failure explanation}: Recent works \citep{tsipras2019robustness, Engstrom2019LearningPR} provide evidence that robust models \citep{Madry2017TowardsDL} are more interpretable than standard models. Thus, some recent works use the penultimate layer neurons of robust models as visual attribute detectors for discovering failure modes. Among these, \citet{sparsewong21b} can only analyze the failures of robust models which achieve lower accuracy than standard (non-robust) models. Barlow \citep{singlaCVPR2021} can analyze the failures of \textit{any model} (standard/robust) but is not useful for highly accurate models. The framework of \citet{salientimagenet2021} addresses these limitations and is discussed in Section \ref{sec:review_sal_imagenet}.
% \noindent \textbf{Failure explanation using robust models}: Recent works \citep{tsipras2019robustness, Engstrom2019LearningPR} provide evidence that robust models \citep{Madry2017TowardsDL} are more interpretable than standard models. Based on this observation, \citet{sparsewong21b, singlaCVPR2021, salientimagenet2021} use the penultimate layer neurons of robust models as the visual attribute detectors for discovering failure modes. \citet{sparsewong21b} can only analyze failures of robust models which achieve lower accuracy than standard (non-robust) models. Barlow \citep{singlaCVPR2021} can analyze the failures of any trained model with a sufficiently large number of misclassified instances in the benchmark dataset and thus is not useful for models with few misclassified examples. %highly accurate models. %(with few misclassified examples).
% The framework of \citet{salientimagenet2021} circumvents these limitations and is discussed in Section \ref{sec:review_sal_imagenet}.
%We discuss the framework of \citet{salientimagenet2021} in detail in Section \ref{sec:review_sal_imagenet}.
%\citet{salientimagenet2021} introduced an approach for discovering the failure modes of highly accurate models. %An application of their approach on a subset of Imagenet (with $232$ classes) resulted in the {\it Salient Imagenet} dataset which can be used to evaluate the sensitivity of any pretrained model to spurious features. However, due to its small size ($\sim 52$k images), this dataset is not useful for training models. 

\noindent \textbf{Domain generalization}:  In this setting, we aim to learn predictors which generalize to test distributions different from training data. The frameworks studied either make assumptions about covariate/label shifts \cite{Widmer2004LearningIT, JMLR:v10:bickel09a, pmlr-v80-lipton18a}, or that test distribution is in some set around training data \cite{Bagnell2005RobustSL, Rahimian2019DistributionallyRO}, or that training data is sampled from distinct distributions \cite{NIPS2011_b571ecea, pmlr-v28-muandet13, Sagawa*2020Distributionally}. Many works provide formal guarantees by assuming invariance in the causal structure of the data \cite{10.5555/2074022.2074085, 10.5555/3020419.3020437, noauthororeditor, HeinzeDeml2018InvariantCP, HeinzeDeml2021ConditionalVP, Christiansen2021ACF}. IRM \cite{arjovsky2020invariant} was also designed for this setting but lacked strong theoretical guarantees. Since then, there have been many works on improving the IRM objective \cite{Xie2020RiskVP, pmlr-v119-chang20c, invariantgamesahuja2020, Krueger2021OutofDistributionGV, mahajan2020domain} and comparing ERM and IRM from theoretical \cite{ahuja2021empirical, rosenfeld2021the, Rosenfeld2021AnOL} and empirical perspectives \cite{gulrajani2021in}.

\begin{figure*}[t]
\centering
\includegraphics[trim=0cm 0cm 0cm 0cm,  width=\linewidth]{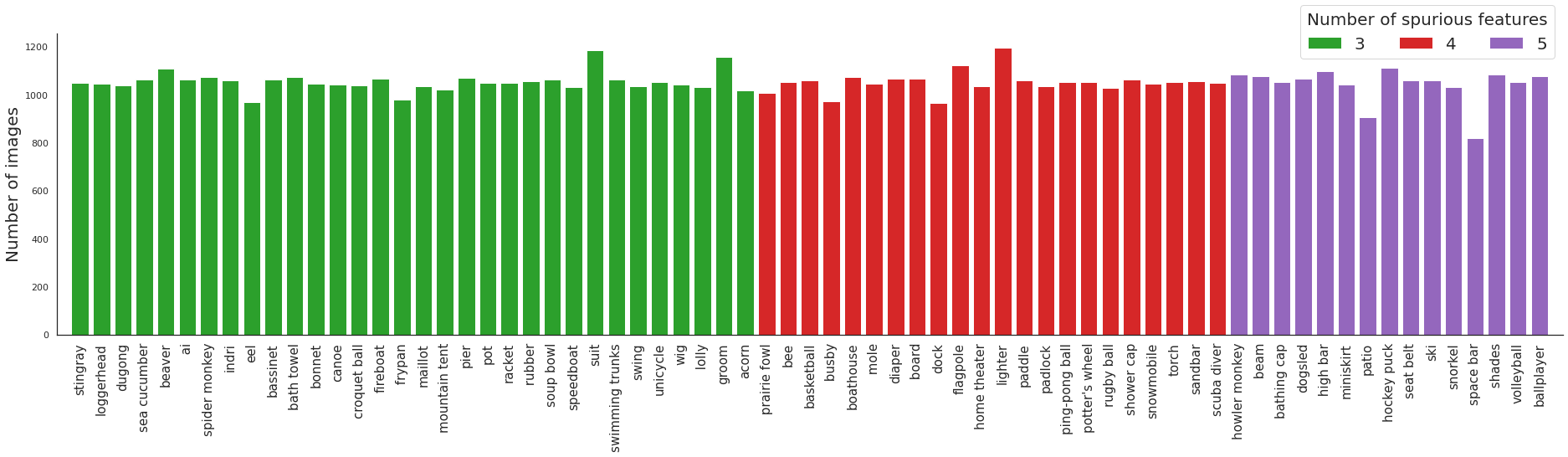}
\caption{Number of images for classes with at least $3$ spurious features in the Salient Imagenet-1M training set. %For $15$ classes, all $5$ features were found to be spurious.
}
\label{fig:hist_salient_imagenet}
\end{figure*}

% In contrast, our approach circumvents these limitations and discovers the spurious features for standard models even when they achieve high accuracy.
\section{Notation and Definitions}\label{sec:notation}
The activation vector in the penultimate layer (after global average pooling) of a trained neural network is called the \textit{neural feature vector}. Each element of this vector is called a \emph{neural feature}. For an image $\bx$ and neural feature $\feature$, we can obtain the \textit{Neural Activation Map} or \textbf{NAM} (similar to CAM by \citet{ZhouKLOT15}) that provides a soft mask for the highly activating pixels in $\bx$ for the feature $\feature$. The corresponding \textbf{heatmap} can be obtained by overlaying the NAM on top of $\bx$ so that the red region highlights the highly activating pixels. The \textbf{feature attack} \cite{Engstrom2019LearningPR} is generated by optimizing the image $\bx$ to increase the value of feature $\feature$. These methods are discussed in more detail in Appendix \ref{sec:appendix_viz_neurons}. For each class $\class \in \allclasses$, we define {\bf core features} $(\text{denoted by } \causal(\class))$ as the set of features that are always a part of the object $\class$, {\bf spurious features} $(\text{denoted by }\spurious(\class))$ are the ones that are likely to {\it co-occur} with $\class$, but not a part of it. Example visualizations for core and spurious features (using heatmaps, feature attack) are in Figures \ref{fig:core_example} and \ref{fig:spurious_example} respectively.

\section{Expanding the Salient Imagenet dataset}\label{sec:scaling_salient_imagenet}
The original Salient Imagenet dataset introduced by \citet{salientimagenet2021} is limited to $232$ classes (denoted by $\oldclasses$) with a total size of $52,521$ images ($\approx 226$ images per class). Each instance in the dataset is of the form $(\bx, y, \causalmask, \spuriousmask)$ where $y$ is the ground truth label and $\causalmask / \spuriousmask$ denote the set of core/spurious masks for the image $\bx$ respectively. By adding noise to the core/spurious regions using these masks and observing the drop in the accuracy, this dataset can be used to test the sensitivity of \emph{any pretrained model} to different visual features. %by corrupting images using their relevant masks and 
While useful for evaluating models, the relatively small size of this dataset limits its usefulness for training large models. Thus, our first goal is to significantly expand the size of the Salient Imagenet dataset so that we can use it for training deep models that mainly rely on core features for their inferences. %We also believe this expanded Salient Imagenet dataset can itself be a useful resource to the community.  %\Sahil{Is this fine?}

\subsection{Review of the Salient Imagenet Framework}\label{sec:review_sal_imagenet}
% Consider the Imagenet classification problem: $\cX \to \allclasses$. Our goal is to predict the ground truth label $y \in \cY$ for inputs $\bx \in \cX$. 

%\SF{can we, in a high level, explain what salient imagenet samples/dataset represent before going into details of how this dataset was collected? you can use either $\cD_{i,j}$ notation or just show that for each sample, potentially two sets of core and spurious masks are provided.}

For each class $\class \in \oldclasses$, using an adversarially trained (robust) model, \citet{salientimagenet2021} first identified the $5$ neural features that are most predictive of $\class$ using the \textit{Neural Feature Importance} scores (details in Appendix \ref{sec:appendix_select_neurons}) resulting in $232 \times 5 = 1160$ (class, feature) pairs. Next for each class $\class \in \oldclasses$, they annotated each of these $5$ features as core or spurious using a Mechanical Turk (MTurk) study.

To obtain the core/spurious annotation for each (class=$\class$, feature=$\feature$) pair,  they showed the MTurk workers two panels: one describes the class $\class$ while the other visualizes the feature $\feature$. To describe the class $\class$, they showed the object names \citep{Miller95wordnet}, object supercategory \cite{tsipras2020imagenet}, object definition, wikipedia links and $3$ images with label $\class$ from the Imagenet validation set. The feature $\feature$ is visualized using $5$ images with predicted class $\class$ that maximally activate the feature $\feature$, their heatmaps and feature attack visualizations (Examples in Figures \ref{fig:core_example} and \ref{fig:spurious_example}). 

Next, they asked the workers to determine whether the visual attribute (inferred by visualizing $\feature$) is a part of the main object (i.e. class $\class$), some separate objects, or the background. They also required the workers to provide reasons for their answers and rate their confidence on a likert scale from $1$ to $5$. The design for this study is shown in Appendix Figure \ref{fig:discover_spurious_study}. Each of these \textit{Human Intelligence Tasks} (or HITs) were evaluated by $5$ workers. The HITs for which majority of the workers (i.e. $ \geq 3$) voted for either separate object or background were deemed to be spurious and the ones with main object as the majority vote were deemed to be core.

\subsection{Mechanical Turk studies for Salient Imagenet-1M}\label{sec:mturk_studies}
\textbf{Discovering core/spurious features for 768 classes}: We used the same procedure discussed in Section \ref{sec:review_sal_imagenet} to obtain core/spurious annotations for the remaining $768$ classes (denoted by $\allclasses - \oldclasses$). Out of the total $768 \times 5 = 3840$ (class, feature) pairs that we evaluated, $3,372$ are deemed to be core and $468$ to be spurious by the workers. On merging the annotations from \citet{salientimagenet2021}, we obtain such annotations for all $1,000$ classes of Imagenet. In total, we obtain $4,370$ core and $630$ spurious class, feature pairs. For $357$ classes, we discover at least $1$ spurious feature. For $15$ classes, all $5$ features were found to be spurious (shown in Figure \ref{fig:hist_salient_imagenet}). 
We visualize several spurious features in Appendix \ref{sec:background_spurious_appendix}
(background) and \ref{sec:foreground_spurious_appendix}
(foreground). %\Sahil{Is this fine?} 

\textbf{Discovering large sets of images containing core/spurious features}: To further expand the Salient Imagenet dataset, we validate that for the (class=$\class$, feature=$\feature$) pair, the visual attribute inferred by visualizing the feature $\feature$ (using top-$5$ images with prediction $\class$) is also highlighted by the NAMs for the images with ground truth label $\class$ and top-$k$ ($k \gg 5$) values of feature $\feature$. This is expected because in the standard ERM paradigm for training deep models, a model will learn to associate a visual attribute with the class $\class$ only if the dataset contains a sufficiently large number of images with ground truth label $\class$ containing the same attribute. 

%\red{To further expand the Salient Imagenet dataset, we validate that the visual attribute inferred by visualizing the feature $\feature$ for some (class=$\class$, feature=$\feature$) pair, is also highlighted by the NAMs for the images with ground truth label $\class$ and top-$k$ ($k \gg 5$) values of feature $\feature$.} \SF{I don't understand this part} 

% is also present in the images with ground truth label $\class$ and top-$k$ ($k \gg 5$) values of the feature $\feature$ and the NAMs for these images (for feature $\feature$) highlight the same attribute. 

To validate that NAMs indeed focus on the desired attributes, we conducted another MTurk study. For the (class=$\class$, feature=$\feature$) pair, we first obtain the training images with label $\class$ ($\approx 1300$ images/label in Imagenet) and top-$260$ activations ($20\%$ of $1300$) of the feature $\feature$. From this set of $260$ images, we selected $5$ images with the lowest activations of feature $\feature$ and randomly selected $10$ images from the remaining set (excluding the already selected images). We show the workers three panels. The first panel shows images and heatmaps with the highest $5$ activations, the second with next $5$ highest activations and the third with lowest $5$ activations. For each heatmap, workers were asked to determine if the highlighted attribute looked different from at least $3$ other heatmaps in the \emph{same panel}. Next, they were asked to determine if the heatmaps in the $3$ \emph{different panels} focused on the same visual attribute, different attributes or if the visualization in any of the panels was unclear. The design of the study is shown in Appendix Figure \ref{fig:mturk_validation_study}. For all (class, feature) pairs ($1000 \times 5 = 5000$ total), we obtained answers from $5$ workers each. For $4763$ pairs (i.e. $95.26\%$), majority of workers selected {\it same} as the answer to both questions. {\it For core features, we observe significantly higher validation rate of $96.48\%\ (4216/4370)$ than for spurious features $86.83\%\ (547/630)$.} These results indicate the quality of our annotated masks (specially the core ones) is high.

\section{The Salient Imagenet-1M dataset}\label{sec:salient_imagenet}
In Section \ref{sec:mturk_studies}, for each class $\class \in \allclasses$, we obtain a set of core and spurious features denoted by $\causal(\class)$ and $\spurious(\class)$, respectively. We also validated that the NAMs highlight the same visual attribute for a large number of images in $95.26\%$ of all (class=$\class$, feature=$\feature$) pairs. These results enable us to significantly expand the size of the Salient Imagenet dataset and use it for training reliable deep models. 

%the high validation rate (especially for core features: $96.48\%$) suggests that for each (class=$\class$, feature=$\feature$) pair, we can use significantly more NAMs than the ones for top-$260$ images as the soft masks to highlight the visual attribute in $\feature$. 

%If $\feature \in \causal(\class)$, these are called \emph{core masks} and if $\feature \in \spurious(\class)$ then \emph{spurious masks}. 

\subsection{Train and Test sets of Salient Imagenet-1M}
For the Imagenet dataset,  the test set was constructed by selecting $50$ images per class resulting in the test set of $50 \times 1000 = 50,000$ images. However, such a test set may not be adequate for testing the sensitivity of a trained model to spurious features because for each class $\class \in \allclasses$, we want our test set to include: (i) a large number of images per spurious feature that the class $\class$ is vulnerable to, and (ii) masks for core/spurious regions in these images so that by adding noise to these regions, we can test the sensitivity of the model to these features for making its predictions. 

\textbf{Test set.} To construct the test set, for each $\class \in \allclasses, \feature \in \causal(\class) \cup \spurious(\class)$, we first define $\dataset(\class, \feature)$ as the set of images with label $\class$ and top-$65$ activations of $\feature$, and their NAMs. The NAMs for these images act as the soft masks that highlight the visual attribute encoded in $\feature$. If $\feature \in \causal(\class)$, these are called \emph{core masks} and if $\feature \in \spurious(\class)$ then \emph{spurious masks}. By taking the union of these sets, i.e., $\cup_{\feature \in \causal(\class) \cup \spurious(\class)} \dataset(\class, \feature)$, we obtain the desired test set for class $\class$. In total, the test set contains $226,946$ images across all $1,000$ classes.  %Each instance in the test set is of the form $(\bx, y, \causalmask, \spuriousmask)$ where $y$ is the ground truth label while $\spuriousmask$ and $\causalmask$ represent the set of spurious and core masks for the image $\bx$, respectively. 

\textbf{Training set.} To construct the training set for class $\class \in \allclasses$, we follow the same procedure as above. However, to keep the training and test sets disjoint, we only select images with label $\class$ that have \textit{not already been included in the test set for class $\class$.} This results in $1,054,221$ training images for $1000$ classes. We plot the number of images in the training set for classes with at least $3$ spurious features in Figure \ref{fig:hist_salient_imagenet}. We note that the NAM validation procedure discussed in Section \ref{sec:mturk_studies} has been performed for top-$260$ images per (class,feature) pair and remaining masks in the training set may not have the same level of quality. However, one can easily specify a constant $k$ to select masks for each (class=$\class$, feature=$\feature$) pair, only for the images with label $\class$ and top-$k$ values of $\feature$. %and all ones ($\ones$) for the remaining images. 

%We acknowledge that because we only validated the NAMs for top-$260$ images per pair in Section \ref{sec:mturk_studies}, the masks for all images in the training set may not be of high quality. However, we can easily specify a constant $k$ so that for each (class=$\class$, feature=$\feature$) pair, only for the images with ground truth label $\class$ with top-$k$ values of feature $\feature$, we use NAM as the mask and all ones ($\ones$) as the mask for remaining images. 

% The union of these training and test sets is what we call the Salient Imagenet-1M dataset (summarized in Table \ref{tab:sal_imagenet_summary}, 1M because it contains masks for more than $1$ million images).

The union of training and test sets is the Salient Imagenet-1M dataset. We use 1M because the \textit{validated dataset} contains more than $1$M mask annotations (see Appendix \ref{sec:appendix_validated_dataset}).

\section{Benchmarking pretrained models}\label{sec:benchmarking}

In this section, we use Salient Imagenet-1M's test set to measure the sensitivity of several pretrained models to core/spurious features by computing the degradation in model performance due to Gaussian noise in core/spurious image regions. The premise here is that if a model does not use the content of a region, then adding noise to the region should have no effect. Contrapositively, if adding noise to a region degrades performance, then the model does make use of the region. \citet{salientimagenet2021} conducted a similar analysis, introducing the concepts of {\it core accuracy} and {\it spurious accuracy}, where core/spurious accuracy is (informally) defined to be the model accuracy on images with added noise in the spurious/core regions, respectively. However, their core and spurious accuracy are evaluated only on images that contain the required mask (i.e. only images with spurious masks were included for measuring the core accuracy). In their dataset (as well as our test set), a sample $\image$ contains a mask for feature $\feature$ only if its activation of $\feature$ is among the top-$65$ for images from its class, resulting in significantly different data over which core and spurious accuracy were computed. %\looseness-1

This results in the following problems: (i) unequal datasets where core/spurious accuracies are computed ($357$ classes have at least $1$ spurious feature while $985$ have at least $1$ core feature), resulting in incomparable numbers, (ii) spurious masks tend to generalize worse than core masks (Section \ref{sec:mturk_studies}), thus the computed core accuracy may not be reliable (iii) core and spurious masks can overlap since they are computed using NAM as the soft segmentation masks. To address these limitations, we evaluate each metric only on images with at least $1$ core mask, and compute the spurious mask as the complement of (i.e. $\ones-$) the mask used for core regions. Furthermore, we employ a new metric, the {\it Relative Core Sensitivity}, that combines core and spurious accuracy to quantify model reliance on core features, while controlling for general noise robustness. Lastly, our analysis is significantly larger than that of \citet{salientimagenet2021}, both in the number of classes and models considered. \looseness-1 

% (ii) only $357$ classes have at least one spurious feature while $985$ have at least one core resulting in significantly smaller dataset for computing core accuracy, 
% the complement of core mask (complement means replacing each pixel with $1$ minus its value). 

%Furthermore, the core masks were observed to generalize better, as mentioned in \ref{sec:mturk_studies}. %Note that by complement, we refer to replacing each pixel with $1$ minus its value.  

%In our analysis, we modify these metrics so that the datasets over which we compute core and spurious accuracies are equal. 

%Further, we employ a new metric, the {\it relative core accuracy}, that combines core and spurious accuracy to quantify model reliance on core features, while controlling for general noise robustness. Lastly, our analysis is significantly larger than that of \citet{salientimagenet2021}, both in the number of classes and models considered.  

\subsection{Revised Core and Spurious Accuracy}
\begin{figure}
    \centering
    \includegraphics[width=\linewidth]{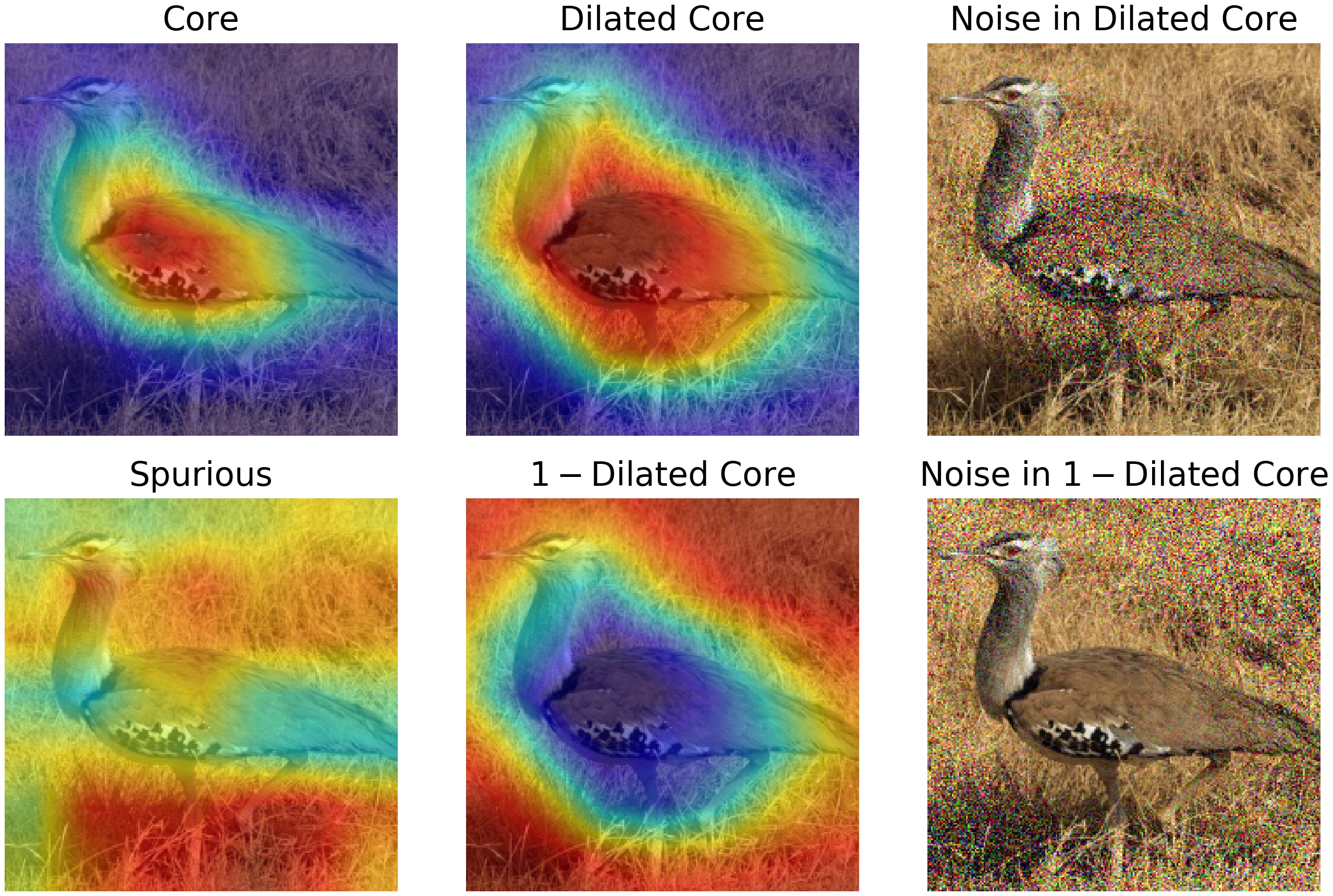}
    \caption{Raw and dilated core masks. First column corresponds to masks used by \citet{salientimagenet2021} to compute spurious and core accuracies, while second column corresponds to masks used in this work. Third column shows noise $(\sigma=0.25)$ applied.}
    \label{fig:dilation}
\end{figure}

Each image in the Salient Imagenet-1M test set may have up to five NAMs for core features. Similar to \citet{salientimagenet2021}, for each image, we take the elementwise maximum of the NAMs for its core features to come up with a single consolidated core mask per image (referred to as $\bc$). We also observe that in practice, the core masks often do not cover the entirety of the core region (Figure \ref{fig:dilation} top left). To ameliorate this, we apply a dilation transform that iteratively replaces each pixel value with the maximum pixel value within a small square kernel (Figure \ref{fig:dilation}, second column). %\looseness-1
%a consequence of inspecting only five images per class. 

\begin{definition}{(\textbf{Dilated Core Mask})}
For mask $\mask$, $1$ iteration of {\it dilation} with square kernel of side $2k+1$ is defined:
$$[\text{dil}(\mask)]_{p,q} = \max_{u, v\ \in\  [-k,k]} \mask_{p + u, q + v}$$
The {\it dilated core mask}, denoted as $\tilde{\bc}$, is obtained by applying 15 iterations of dilation using $k=2$ on the core mask $\bc$. 
\end{definition}

In Figure \ref{fig:dilation}, we visualize the difference in core/spurious mask computation procedures between our work and \citet{salientimagenet2021}. In the left column, we see that using spurious masks obtained in the same manner as core masks (i.e. by taking the max over NAMs of spurious features) %\red{using spurious masks directly} \SF{not clear} 
may introduce an incongruity in what is considered core and spurious. Specifically, while the spurious mask in the bottom left focuses on the background, it also covers much of the core region. However, the $\ones - \tilde{\bc}$ mask (middle column) by design has low overlap with the dilated core mask $(\tilde{\bc})$.

%We corrupt image regions using the same procedure as in \citet{salientimagenet2021}. For an image $\image$, we generate $\image + \sigma (\mask \odot \noise)$ where $\mask$ is the mask (either $\tilde{\bc}$ or $\ones-\tilde{\bc}$), $\noise$ is the noise vector $\noise \sim \mathcal{N}(\mathbf{0}, \mathbf{I})$ and $\sigma$ is a hyperparameter controlling the strength of the corruption. We use $\sigma=0.25$ for all experiments (Figure \ref{fig:dilation} right).

Using these dilated core masks for each sample in the Salient Imagenet-1M {\it test set}, we can now define our revised versions of core and spurious accuracy as follows: 

\begin{definition}{(\textbf{Core and Spurious Accuracy})}
\label{def:causal_spurious_acc}
The Core Accuracy, $\causalaccabv$ for a model $h$ is defined as follows:
\begin{align*}
&\causalaccabv\ = \frac{1}{|\allclasses|} \sum_{\class \in \allclasses} \frac{1}{|\dataset\causal(\class)|} \sum_{\bx \in \dataset\causal(\class)} \mathbbm{1}(h(\bx^{*})=i) \\
& \text{where } \bx^{*} = \bx + \sigma (\bz \odot (\ones-\tilde{\bc})),\quad \bz \sim \mathcal{N(\zeros, \bI)}
% &\causalaccabv\ = \frac{1}{|\allclasses|} \sum_{\class \in \allclasses} \causalaccabv(\class),\qquad\text{where } \causalaccabv \left(\class\right) = \\
% &\frac{1}{|\dataset\causal(\class)|} \sum_{\bx \in \dataset\causal(\class)} \mathbbm{1}(h(\bx + \sigma (\bz \odot (\ones-\tilde{\bc})))=i) 
\end{align*}
where $\dataset\causal(\class) = \cup_{\feature \in \causal(\class)}\dataset(\class, \feature)$. The Spurious Accuracy,  $\spuriousaccabv$ is defined similarly using $\bx^{*} = \bx + \sigma (\bz \odot \tilde{\bc})$.
\end{definition}
We use $\sigma=0.25$ for all experiments (Figure \ref{fig:dilation} last column).

\begin{figure} 
    \centering
    \includegraphics[width=0.8\linewidth]{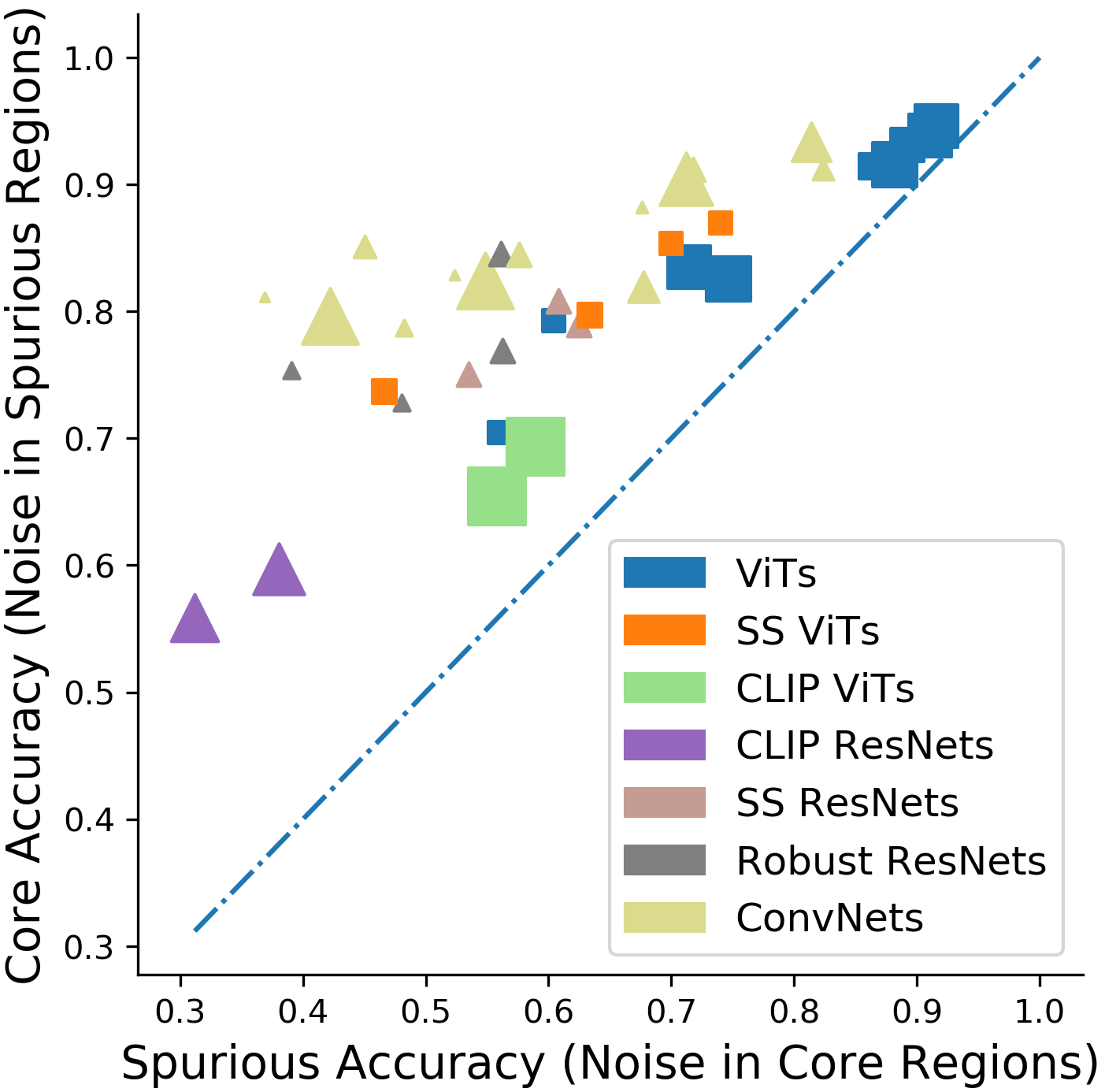}
    \caption{Core/Spurious accuracy evaluated with $\sigma=0.25$. Models with transformer (convolution) architectures have square (triangle) markers. Marker size is proportional to parameter count.}
    \label{fig:core_spur_scatter}
\end{figure}

\subsection{Relative Core Sensitivity}
A limitation of using noise to measure model sensitivity to different image regions is that models extremely robust to noise corruptions will have high core and spurious accuracy and thus less gap between the two (regardless of its use of spurious features). Thus, we introduce a new metric: \emph{Relative Core Sensitivity} that quantifies the model reliance on core features while controlling for general noise robustness (adapted from a similar metric in \citet{mazda}):
%\SF{cite mazda's cvpr work}:
%\Mazda{how about adding 'adapted from a similar metric in \cite{mazda}' and then removing the sentence below.}. 
%it introduces the confounding variable of general noise robustness. For example, a model that is extremely robust to Gaussian noise corruption would have high core and spurious accuracy, regardless of its reliance on spurious features. We can still discern the relative sensitivity to core and spurious regions by comparing core and spurious accuracy to one another (i.e., a large gap between core and spurious accuracy is good). However, models with very high or very low noise robustness afford less room for such gap. To this end, we propose a new metric called the \textbf{R}elative \textbf{C}ore \textbf{S}ensitivity ($\RCS$) as follows: 
% A limitation of using Gaussian noise to measure model sensitivity to different image regions is that it introduces the confounding variable of general noise robustness. For example, a model that is extremely robust to Gaussian noise corruption would have high core and spurious accuracy, regardless of its reliance on spurious features. We can still discern the relative sensitivity to core and spurious regions by comparing core and spurious accuracy to one another (i.e., a large gap between core and spurious accuracy is good). However, models with very high or very low noise robustness afford less room for such gap. To this end, we propose a new metric called the \textbf{R}elative \textbf{C}ore \textbf{S}ensitivity ($\RCS$) as follows: 
\begin{definition}{(\textbf{Relative Core Sensitivity})}
We define the Relative Core Sensitivity or $\RCS$ as follows:
\begin{align*}
    \RCS := \frac{\causalaccabv-\spuriousaccabv}{2 \min(\bar{a}, 1-\bar{a})}, \ \text{where } \bar{a} = \frac{(\causalaccabv +\spuriousaccabv)}{2}
\end{align*}
\end{definition}
Here, $\bar{a}$ acts as a proxy for the general noise robustness of the model under inspection. We can show that for all models with $\bar{a}$ noise robustness, 2$\min(\bar{a}, 1-\bar{a})$ is the maximum possible gap between the core/spurious accuracy. Thus, $\RCS$ normalizes the gap in the current model by the total possible gap. In Figure \ref{fig:core_spur_scatter}, higher $\RCS$ corresponds to lying higher above the diagonal (high core and low spurious accuracy). %A similar metric was also derived in \citet{mazda}. 
We derive this metric in more detail in Appendix \ref{sec:appendix_rcs}. 

%Intuitively, the more $p$ lies above the $y=x$ line, higher is the core and lower is the spurious accuracy (as in Figure \ref{fig:core_spur_scatter}), leading to higher $\RCS$ values. $\RCS$ values for several models are shown in Figure \ref{fig:rca_bar}. 

 %over the total distance possible. %We visualize the $\RCS$ values of the pretrained models in Figure \ref{fig:rca_bar}.

\begin{figure}
    \centering
    \includegraphics[width=0.9\linewidth]{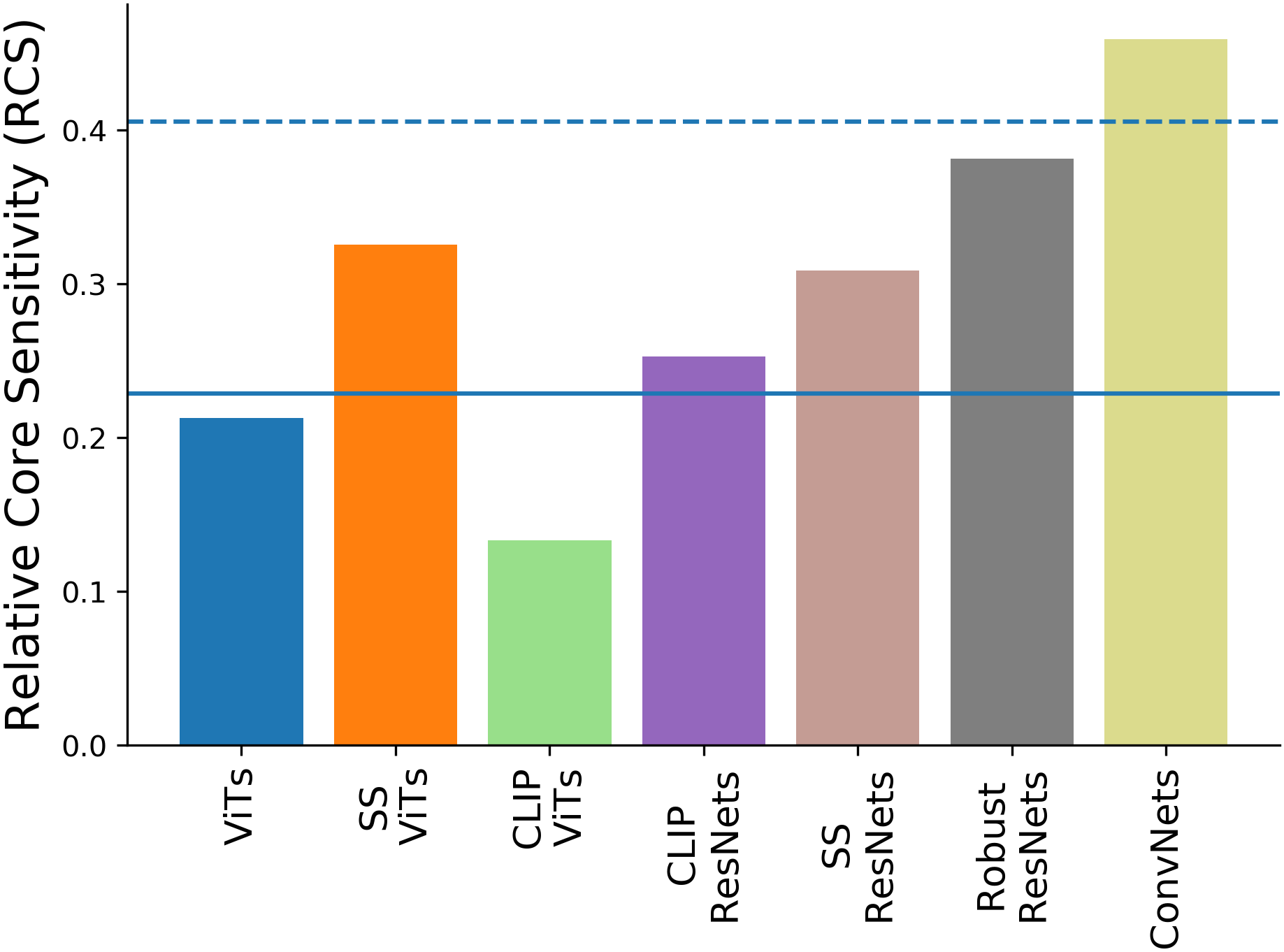}
    \caption{Average $\RCS$ values grouped by model architecture and training procedure. Solid line is average for transformer models ($\RCS=0.23$), dashed for convolutional models ($\RCS=0.41$).}
    \label{fig:rca_bar}
\end{figure}

\subsection{Findings} \label{sec:pretrained_findings_sahil}
We study a large and diverse set of pretrained Imagenet models and training paradigms ($42$ models in total), namely ConvNets \cite{vgg, resnet, wide_resnet, mobilenetv2, resnext, inceptionv2, efficientnet, mnasnet}, ViTs \cite{vit, deit, convit, swin}, Robust ResNets \cite{robust_resnets}, self-supervised (SS) models on ViT and ResNet arches \cite{MoCov3, dino, simclr}, zero-shot models such as CLIP ResNets, CLIP ViTs \cite{clip}. Details in Appendix \ref{sec:appendix_pretrained_models}.

% I'm going to have one paragraph on observations, and next on conjectures. or should it be observation followed by conjecture per observation?
Figure \ref{fig:core_spur_scatter} shows the core and spurious accuracy for all model categories evaluated. We observe that transformer models (squares) lie closer to the diagonal than convolutional models (triangles) suggesting they rely more on spurious features. We hypothesize that the lack of a proper inductive bias in transformers may lead to this phenomenon. We show the average $\RCS$ values in Figure \ref{fig:rca_bar} (high $\RCS$ implies high core and low spurious accuracy). We again validate that the \emph{transformer models have significantly lower $\RCS$} ($0.23$ compared to $0.41$ for convolutional models). 
In Appendix Table \ref{tab:pretrained_all}, we observe that {\it zero-shot CLIP ViTs yield the lowest $\RCS$ value}. We conjecture that the use of text tokens in zero-shot CLIP models may introduce an additional source of spurious vulnerabilities. We also observe that {\it adversarial training in ResNets decreases $\RCS$} from $0.44$ to $0.38$. A similar result was also observed by \citet{mazda} in a different setup (details in Appendix \ref{sec:appendix_comparing_to_rival10}). %We hypothesize that adversarial training makes \red{it easier to learn backgrounds} \SF{I don't understand this part} \Mazda{other conjectures: (i) increased robustness is obtained by relying on more content (i.e. spur ftrs) to predict (ii)AT decreases sensitivity to corruption of core region more than spur region bc adv attacks may target core regions more} leads to this phenomenon. 
%Our dataset and analysis can help the ML community to investigate the root causes of these phenomena in future research. 
Moreover, our analysis indicates that the {\it standard accuracy is not sufficient to fully characterize model quality}; i.e., different models may have different core accuracy even with the same standard accuracy. For example, EfficientNet-B4 and Inception-V4 (Appendix Table \ref{tab:pretrained_all}) have almost the same clean accuracy ($0.37\%$ gap) but vastly different core accuracy ($9.23\%$ gap).

\begin{table*}[ht!]
    \centering
    \begin{tabular}{l|llll} \toprule
        Training Procedure &  Clean Accuracy ($\uparrow$) & Core Accuracy ($\uparrow$)  & Spurious Accuracy ($\downarrow$) & $\RCS$ ($\uparrow$)  \\ \midrule
        Baseline (ERM) & 74.37\% $\pm$ 0.67 &52.02\% $\pm$ 1.95 & 12.12\% $\pm$ 1.75 & 0.624\ $\pm$ 0.035 \\
        %$1-\bc$ Noising & 22.59\% & 56.24\% & 2.95\%& 90.51\\ 
        Random $\ones-\bc$ Noising &73.74\% $\pm$ 0.69 &59.23\% $\pm$ 6.36 & 11.49\% $\pm$ 5.67 & 0.692\ $\pm$ 0.102 \\
        Saliency Regularization &{\bf 75.06\%} $\pm$ 0.11 &54.37\% $\pm$ 1.66 & 12.37\% $\pm$ 2.39 & 0.633 $\pm$ 0.053\\ 
        \midrule
        Rand Noising + Sal Reg & {\bf 74.95\%} $\pm$ 0.19 &{\bf 63.86\%} $\pm$ 2.92 & {\bf 8.97\%} $\pm$ 2.95 & {\bf 0.759} $\pm$ 0.063\\
        \bottomrule
    \end{tabular}
    \caption{Model performance using various approaches to CoRM. The final row shows a combination of Random $\ones-\bc$ noising and Saliency regularization results in $11.84\%$ increase in core accuracy and $0.58\%$ increase in clean accuracy. Results averaged over $4$ trials.}
    \label{tab:corm_results}
\end{table*}

\section{Core Risk Minimization (CoRM)}\label{sec:corm}

%Models optimized with ERM classify accurately, but not necessarily for the right reasons. Our pretrained model analysis shows that ERM alone does not guarantee that models are significantly more sensitive to core regions than spurious ones. In appendix \ref{sec:motivation_appendix}, we demonstrate the danger of reliance on spurious features: often, samples that are misclassified to a class $\class$ activate core features $\causal(\class)$ less than true objects of the class, but instead activate spurious features much higher. Intuitively, a model that {\it faithfully} learns concepts 
%\looseness=-1
ERM yields classifiers that achieve impressive accuracy but it cannot guide models to learn that certain image regions should inform the class label more than others. 
%ERM yields classifiers that achieve impressive accuracy. However, ERM makes no considerations to the important fact that certain image regions should inform the class label more than others. 
Models that use spurious features can give a false of sense of performance, as accuracy can drop dramatically in a new domain where correlations between class labels and spurious features are broken. A model that faithfully learns concepts should rely more on core features than on spurious ones.

We formalize this notion in  {\bf Co}re {\bf R}isk {\bf M}inimization, defined in optimization (\ref{eq: corm_eq_intro}). CoRM seeks to minimize the expected loss over samples with Gaussian noise added in non-core regions. When all image regions are deemed to be core, CoRM reduces to ERM. However, when informative core masks are available, CoRM requires that the optimal classifier remains accurate in spite of corruption in the spurious regions. Cost of data collection previously inhibited the pursuit of CoRM-like learning. Salient Imagenet-1M's rich core/spurious annotations have the potential to enable training of models that make predictions while avoiding spurious shortcuts. We outline our relaxations to the CoRM objective that lead to significant increases in core accuracy and $\RCS$. \looseness -1 
%We demonstrate that using Salient Imagenet-1M, a few relaxations of the CoRM objective, and simple methods, 
%the introduction of Salient ImageNet-1M's large scale instance-wise annotations of core and spurious regions opens the door to train models that make predictions while avoiding spurious shortcuts. We outline relaxations to the CoRM objective and empirical methods that lead to significant increases in core accuracy and $\RCS$. 
% classify correctly and {\it for the right reasons}.   
\subsection{Relaxing CoRM}

% Solving the CoRM optimization involves computing the expected loss over a high dimensional Gaussian distribution, which is computationally expensive for deep models. \red{The outer expectation can also only be approximated due to lack of knowledge of the underlying data distribution. ERM relaxes this constraint.} \Sahil{I believe this part can be removed or shortened. People know about this stuff already}, as it operates under the assumption that empirical risk over a finite training set represents the true (population) risk sufficiently closely, and so minimizing empirical risk yields models that generalize well. 

First, we approximate the inner expectation of CoRM with a single sample. That is, for each input $(\bx, \bc)$ with label $y$, we draw a random noise vector $\bz \sim \mathcal{N}(\mathbf{0},\ \sigma^2 \bI)$, and use $$\mathbb{E}_{\bx^*\sim\mathbb{P}(\bx, \bc, \sigma)} [\ell (f_\theta(\bx^*), y)] \approx \ell(f_\theta(\bx + (\ones-\bc) \odot \bz), y)$$
Next, to obtain $\bc$ for {\it any} sample $\image \in \class$ in Salient Imagenet, we use the NAMs for all core features $\causal(\class)$, regardless of the activation of $\image$ on the feature. This differs from the test-set computation of $\bc$, and introduces noisier core masks, but facilitates a massive increase in training set size. We note that $\bc$ can be dilated (or eroded) to any degree, introducing a hyperparameter allowing the practitioner to choose the amount of surrounding context the model can use without penalty. While $\sigma$ alters the magnitude of corruption, dilation alters the {\it region} of corruption, introducing a spatial bias in favor of spurious features that are near the core ones. In our experiments, we train on masks $\bc$ with no dilation applied.

\subsection{Methods}
% other names for the method: partial noise augmentation, 
We explore two efficient approaches to perform CoRM and ultimately reduce the reliance on spurious features: (i) random $\ones-\bc$ noising and (ii) saliency regularization. Following directly from the relaxed formulation of CoRM, $\ones-\bc$ noising minimizes risk on samples augmented with additive Gaussian noise scaled by $\ones-\bc$. In practice, we find that deep networks overfit to noise in the non-core regions, leading to degraded clean accuracy. However, witholding the additive noise for randomly selected batches (e.g., with probability with $p=0.5$) during training leads to a better trade-off between clean and core accuracies.

%{\it Non-core noising} comes directly from the relaxed formulation of CoRM, and augments every sample with additive Gaussian noise scaled by the complement of the soft core segmentation mask. In practice, we find that deep networks can overfit to noise in the spurious regions, leading to degraded clean accuracy. However, witholding the additive noise for randomly selected batches (e.g., with probability with $p=0.5$) during training leads to a better trade-off between clean and core accuracies. We call this approach {\it random non-core noising} \Sahil{Call it random non-core noising or 1-core noising (the latter seems to be better to me)??}. 
A second approach utilizes gradient information to perform {\it saliency regularization}. Such regularization has been shown to improve generalization \cite{simpson2019gradmask}, robustness to distribution shift in non-core regions \cite{chang2021towards}, and model interpretability \cite{aya}. Formally, for a sample $\bx$ with core region $\bc$ and label $y$, saliency regularization introduces the following loss term:
$$\mathcal{L}_{sal} = \|(\ones-\bc) \odot \nabla_{\bx} \ell(f_\theta(\bx), y) \|^2 $$
where $\ell$ is the classification loss. We compute the saliency penalty after a full forward and backward pass, as the input gradients are then readily available. Model parameters $\theta$ are then updated to minimize $\mathcal{L}_{sal}$. Notice that because saliency regularization and spurious noising affect opposite ends of the training pipeline (pre-forward pass vs. post-backward pass), they can be combined easily. 

\subsection{Findings}

\looseness=-1
We train Resnet-50 models on Salient Imagenet-1M, employing the two aforementioned methods and their combinations, as well as a baseline model that uses the standard paradigm, ERM. We seek to demonstrate the feasibility of methods toward achieving CoRM's objective, not to obtain highest possible accuracies. Thus, we do not perform data augmentation. We provide details about how data augmentation can be used with Salient Imagenet-1M masks in Appendix \ref{sec:appendix_corm_training}. %Complete training details are provided in appendix \ref{sec:appendix_corm_training}. 
In addition to clean accuracy, we present core and spurious accuracy, as well as $\RCS$, each computed over the test set of Salient Imagenet-1M, following the evaluation protocol of Section \ref{sec:benchmarking}. 

%In experiments of this section, we train Resnet-50 model architectures  on Salient ImageNet-1M, employing the two aforementioned methods, as well as their combination, and a baseline model that uses the standard ERM. We stress that the intention of these experiments is to catalog the effects of each approach, {\it not} to achieve the highest possible accuracies. For this reason, and also to avoid uncontrolled effects, we do {\it not} perform augmentation. Further, random cropping, a common augmentation technique, is less feasible for Salient ImageNet-1M, as the masks are obtained for images after undergoing the standard ImageNet test transformations of resizing and taking a square center crop. \red{However, by generating such masks during training on the processed image (using the NAM of robust model), we can indeed use data augmentation for training models and this approach can be explored in future works.} \Sahil{@Mazda: Is this fine???}
%\Sahil{@Mazda: We should mention that our Salient Imagenet dataset can be easily modified to work with random resize cropping.}

%We note that one can still additionally augment both the mask and processed image in tandem, though content discarded with the first transform cannot be used in conjunction with the masks \SF{can data augmentation be used in future studies on salient imagenet?} \Sahil{Yes, we can generate the salient imagenet dataset for original Imagenet images and then perform data augmentation}. 

Table \ref{tab:corm_results} summarizes the results. We find that random $\ones-\bc$ noising increases core accuracy by $7.21\%$ relative to baseline. Saliency regularization has a more modest improvement in core accuracy, but improves clean accuracy. Combining the two methods yields a $11.84\%$ increase in core accuracy, while also marginally improving clean accuracy. Moreover, spurious accuracy decreases significantly, causing a large improvement of $0.13$ in $\RCS$. These preliminary results suggest that the goals of achieving high clean accuracy while also maintaining high core accuracy and having a low spurious accuracy can be made feasible through Salient Imagenet-1M. We hope that our introduced dataset and training methods will lead to the development of deep models that mainly rely on core features for their inferences. %and thus more reliable deep learning. %and shed more light on the reliance of spurious features in deep learning. %Further development of methods and evaluation of the effects of training models that strictly classify based on core regions is paramount in building and understanding the future of trustworthy deep image classifiers. 

\bibliography{main}

\begin{thebibliography}{93}
\providecommand{\natexlab}[1]{#1}
\providecommand{\url}[1]{\texttt{#1}}
\expandafter\ifx\csname urlstyle\endcsname\relax
  \providecommand{\doi}[1]{doi: #1}\else
  \providecommand{\doi}{doi: \begingroup \urlstyle{rm}\Url}\fi

\bibitem[Adebayo et~al.(2018)Adebayo, Gilmer, Muelly, Goodfellow, Hardt, and
  Kim]{sanitychecks2018}
Adebayo, J., Gilmer, J., Muelly, M., Goodfellow, I.~J., Hardt, M., and Kim, B.
\newblock Sanity checks for saliency maps.
\newblock In \emph{NeurIPS}, 2018.

\bibitem[Ahuja et~al.(2020)Ahuja, Shanmugam, Varshney, and
  Dhurandhar]{invariantgamesahuja2020}
Ahuja, K., Shanmugam, K., Varshney, K., and Dhurandhar, A.
\newblock Invariant risk minimization games.
\newblock In III, H.~D. and Singh, A. (eds.), \emph{Proceedings of the 37th
  International Conference on Machine Learning}, volume 119 of
  \emph{Proceedings of Machine Learning Research}, pp.\  145--155. PMLR, 13--18
  Jul 2020.
\newblock URL \url{https://proceedings.mlr.press/v119/ahuja20a.html}.

\bibitem[Ahuja et~al.(2021)Ahuja, Wang, Dhurandhar, Shanmugam, and
  Varshney]{ahuja2021empirical}
Ahuja, K., Wang, J., Dhurandhar, A., Shanmugam, K., and Varshney, K.~R.
\newblock Empirical or invariant risk minimization? a sample complexity
  perspective.
\newblock In \emph{International Conference on Learning Representations}, 2021.
\newblock URL \url{https://openreview.net/forum?id=jrA5GAccy_}.

\bibitem[Arjovsky et~al.(2020)Arjovsky, Bottou, Gulrajani, and
  Lopez-Paz]{arjovsky2020invariant}
Arjovsky, M., Bottou, L., Gulrajani, I., and Lopez-Paz, D.
\newblock Invariant risk minimization, 2020.

\bibitem[Bagnell(2005)]{Bagnell2005RobustSL}
Bagnell, J.~A.
\newblock Robust supervised learning.
\newblock In \emph{AAAI}, 2005.

\bibitem[Bau et~al.(2020)Bau, Zhu, Strobelt, Lapedriza, Zhou, and
  Torralba]{Bau30071}
Bau, D., Zhu, J.-Y., Strobelt, H., Lapedriza, A., Zhou, B., and Torralba, A.
\newblock Understanding the role of individual units in a deep neural network.
\newblock \emph{Proceedings of the National Academy of Sciences}, 117\penalty0
  (48):\penalty0 30071--30078, 2020.
\newblock ISSN 0027-8424.
\newblock \doi{10.1073/pnas.1907375117}.
\newblock URL \url{https://www.pnas.org/content/117/48/30071}.

\bibitem[Beery et~al.(2018)Beery, Horn, and Perona]{terraincognita2018}
Beery, S., Horn, G.~V., and Perona, P.
\newblock Recognition in terra incognita.
\newblock \emph{CoRR}, abs/1807.04975, 2018.
\newblock URL \url{http://arxiv.org/abs/1807.04975}.

\bibitem[Bickel et~al.(2009)Bickel, Br{{\"u}}ckner, and
  Scheffer]{JMLR:v10:bickel09a}
Bickel, S., Br{{\"u}}ckner, M., and Scheffer, T.
\newblock Discriminative learning under covariate shift.
\newblock \emph{Journal of Machine Learning Research}, 10\penalty0
  (75):\penalty0 2137--2155, 2009.
\newblock URL \url{http://jmlr.org/papers/v10/bickel09a.html}.

\bibitem[Bissoto et~al.(2020)Bissoto, Valle, and Avila]{Bissoto2020DebiasingSL}
Bissoto, A., Valle, E., and Avila, S.
\newblock Debiasing skin lesion datasets and models? not so fast.
\newblock \emph{2020 IEEE/CVF Conference on Computer Vision and Pattern
  Recognition Workshops (CVPRW)}, pp.\  3192--3201, 2020.

\bibitem[Blanchard et~al.(2011)Blanchard, Lee, and Scott]{NIPS2011_b571ecea}
Blanchard, G., Lee, G., and Scott, C.
\newblock Generalizing from several related classification tasks to a new
  unlabeled sample.
\newblock In Shawe-Taylor, J., Zemel, R., Bartlett, P., Pereira, F., and
  Weinberger, K.~Q. (eds.), \emph{Advances in Neural Information Processing
  Systems}, volume~24. Curran Associates, Inc., 2011.
\newblock URL
  \url{https://proceedings.neurips.cc/paper/2011/file/b571ecea16a9824023ee1af16897a582-Paper.pdf}.

\bibitem[Caron et~al.(2021)Caron, Touvron, Misra, J\'egou, Mairal, Bojanowski,
  and Joulin]{dino}
Caron, M., Touvron, H., Misra, I., J\'egou, H., Mairal, J., Bojanowski, P., and
  Joulin, A.
\newblock Emerging properties in self-supervised vision transformers.
\newblock In \emph{Proceedings of the International Conference on Computer
  Vision (ICCV)}, 2021.

\bibitem[Carter et~al.(2019)Carter, Armstrong, Schubert, Johnson, and
  Olah]{carter2019activation}
Carter, S., Armstrong, Z., Schubert, L., Johnson, I., and Olah, C.
\newblock Activation atlas.
\newblock \emph{Distill}, 2019.
\newblock \doi{10.23915/distill.00015}.
\newblock https://distill.pub/2019/activation-atlas.

\bibitem[Chang et~al.(2019)Chang, Creager, Goldenberg, and
  Duvenaud]{chang2018explaining}
Chang, C.-H., Creager, E., Goldenberg, A., and Duvenaud, D.
\newblock Explaining image classifiers by counterfactual generation.
\newblock In \emph{International Conference on Learning Representations}, 2019.
\newblock URL \url{https://openreview.net/forum?id=B1MXz20cYQ}.

\bibitem[Chang et~al.(2021)Chang, Adam, and Goldenberg]{chang2021towards}
Chang, C.-H., Adam, G.~A., and Goldenberg, A.
\newblock Towards robust classification model by counterfactual and invariant
  data generation.
\newblock \emph{arXiv preprint arXiv:2106.01127}, 2021.

\bibitem[Chang et~al.(2020)Chang, Zhang, Yu, and Jaakkola]{pmlr-v119-chang20c}
Chang, S., Zhang, Y., Yu, M., and Jaakkola, T.
\newblock Invariant rationalization.
\newblock In III, H.~D. and Singh, A. (eds.), \emph{Proceedings of the 37th
  International Conference on Machine Learning}, volume 119 of
  \emph{Proceedings of Machine Learning Research}, pp.\  1448--1458. PMLR,
  13--18 Jul 2020.
\newblock URL \url{https://proceedings.mlr.press/v119/chang20c.html}.

\bibitem[Chen et~al.(2020)Chen, Kornblith, Norouzi, and Hinton]{simclr}
Chen, T., Kornblith, S., Norouzi, M., and Hinton, G.
\newblock A simple framework for contrastive learning of visual
  representations.
\newblock \emph{arXiv preprint arXiv:2002.05709}, 2020.

\bibitem[Chen* et~al.(2021)Chen*, Xie*, and He]{MoCov3}
Chen*, X., Xie*, S., and He, K.
\newblock An empirical study of training self-supervised vision transformers.
\newblock \emph{arXiv preprint arXiv:2104.02057}, 2021.

\bibitem[Christiansen et~al.(2021)Christiansen, Pfister, Jakobsen, Gnecco, and
  Peters]{Christiansen2021ACF}
Christiansen, R., Pfister, N., Jakobsen, M.~E., Gnecco, N., and Peters, J.
\newblock A causal framework for distribution generalization.
\newblock \emph{IEEE transactions on pattern analysis and machine
  intelligence}, PP, 2021.

\bibitem[Chung et~al.(2019)Chung, Kraska, Polyzotis, Tae, and
  Whang]{chung2019slice}
Chung, Y., Kraska, T., Polyzotis, N., Tae, K.~H., and Whang, S.~E.
\newblock Slice finder: Automated data slicing for model validation.
\newblock In \emph{2019 IEEE 35th International Conference on Data Engineering
  (ICDE)}, pp.\  1550--1553. IEEE, 2019.

\bibitem[d'Ascoli et~al.(2021)d'Ascoli, Touvron, Leavitt, Morcos, Biroli, and
  Sagun]{convit}
d'Ascoli, S., Touvron, H., Leavitt, M., Morcos, A., Biroli, G., and Sagun, L.
\newblock Convit: Improving vision transformers with soft convolutional
  inductive biases.
\newblock \emph{arXiv preprint arXiv:2103.10697}, 2021.

\bibitem[de~Haan et~al.(2019)de~Haan, Jayaraman, and
  Levine]{causalconfusion2019}
de~Haan, P., Jayaraman, D., and Levine, S.
\newblock Causal confusion in imitation learning.
\newblock \emph{CoRR}, abs/1905.11979, 2019.
\newblock URL \url{http://arxiv.org/abs/1905.11979}.

\bibitem[DeGrave et~al.(2021)DeGrave, Janizek, and Lee]{deGrave2021aa}
DeGrave, A.~J., Janizek, J.~D., and Lee, S.-I.
\newblock Ai for radiographic covid-19 detection selects shortcuts over signal.
\newblock \emph{Nature Machine Intelligence}, 3\penalty0 (7):\penalty0
  610--619, 2021.

\bibitem[Deng et~al.(2009)Deng, Dong, Socher, Li, Li, and Fei-Fei]{5206848}
Deng, J., Dong, W., Socher, R., Li, L., Li, K., and Fei-Fei, L.
\newblock Imagenet: A large-scale hierarchical image database.
\newblock \emph{2009 IEEE Conference on Computer Vision and Pattern
  Recognition}, pp.\  248--255, 2009.

\bibitem[Didelez et~al.(2006)Didelez, Dawid, and
  Geneletti]{10.5555/3020419.3020437}
Didelez, V., Dawid, A.~P., and Geneletti, S.
\newblock Direct and indirect effects of sequential treatments.
\newblock In \emph{Proceedings of the Twenty-Second Conference on Uncertainty
  in Artificial Intelligence}, UAI'06, pp.\  138–146, Arlington, Virginia,
  USA, 2006. AUAI Press.
\newblock ISBN 0974903922.

\bibitem[Dosovitskiy \& Brox(2016)Dosovitskiy and Brox]{DosovitskiyB15}
Dosovitskiy, A. and Brox, T.
\newblock Inverting visual representations with convolutional networks.
\newblock In \emph{The IEEE Conference on Computer Vision and Pattern
  Recognition (CVPR)}, June 2016.

\bibitem[Dosovitskiy et~al.(2020)Dosovitskiy, Beyer, Kolesnikov, Weissenborn,
  Zhai, Unterthiner, Dehghani, Minderer, Heigold, Gelly, Uszkoreit, and
  Houlsby]{vit}
Dosovitskiy, A., Beyer, L., Kolesnikov, A., Weissenborn, D., Zhai, X.,
  Unterthiner, T., Dehghani, M., Minderer, M., Heigold, G., Gelly, S.,
  Uszkoreit, J., and Houlsby, N.
\newblock An image is worth 16x16 words: Transformers for image recognition at
  scale.
\newblock \emph{CoRR}, abs/2010.11929, 2020.
\newblock URL \url{https://arxiv.org/abs/2010.11929}.

\bibitem[Engstrom et~al.(2019)Engstrom, Ilyas, Santurkar, Tsipras, Tran, and
  Madry]{Engstrom2019LearningPR}
Engstrom, L., Ilyas, A., Santurkar, S., Tsipras, D., Tran, B., and Madry, A.
\newblock Adversarial robustness as a prior for learned representations, 2019.

\bibitem[Goyal et~al.(2019)Goyal, Wu, Ernst, Batra, Parikh, and
  Lee]{deviparikhcounterfactual19}
Goyal, Y., Wu, Z., Ernst, J., Batra, D., Parikh, D., and Lee, S.
\newblock Counterfactual visual explanations.
\newblock In \emph{ICML}, 2019.

\bibitem[Gulrajani \& Lopez-Paz(2021)Gulrajani and Lopez-Paz]{gulrajani2021in}
Gulrajani, I. and Lopez-Paz, D.
\newblock In search of lost domain generalization.
\newblock In \emph{International Conference on Learning Representations}, 2021.
\newblock URL \url{https://openreview.net/forum?id=lQdXeXDoWtI}.

\bibitem[He et~al.(2016)He, Zhang, Ren, and Sun]{resnet}
He, K., Zhang, X., Ren, S., and Sun, J.
\newblock Deep residual learning for image recognition.
\newblock In \emph{Proceedings of the IEEE conference on computer vision and
  pattern recognition}, pp.\  770--778, 2016.

\bibitem[Heinze-Deml \& Meinshausen(2021)Heinze-Deml and
  Meinshausen]{HeinzeDeml2021ConditionalVP}
Heinze-Deml, C. and Meinshausen, N.
\newblock Conditional variance penalties and domain shift robustness.
\newblock \emph{Mach. Learn.}, 110:\penalty0 303--348, 2021.

\bibitem[Heinze-Deml et~al.(2018)Heinze-Deml, Peters, and
  Meinshausen]{HeinzeDeml2018InvariantCP}
Heinze-Deml, C., Peters, J., and Meinshausen, N.
\newblock Invariant causal prediction for nonlinear models.
\newblock \emph{Journal of Causal Inference}, 6, 2018.

\bibitem[Ismail et~al.(2019)Ismail, Gunady, Pessoa, Bravo, and
  Feizi]{Ismail2019AttentionDL}
Ismail, A.~A., Gunady, M.~K., Pessoa, L., Bravo, H.~C., and Feizi, S.
\newblock Input-cell attention reduces vanishing saliency of recurrent neural
  networks.
\newblock In \emph{NeurIPS}, 2019.

\bibitem[Ismail et~al.(2020)Ismail, Gunady, Bravo, and
  Feizi]{Ismail2020BenchmarkingDL}
Ismail, A.~A., Gunady, M.~K., Bravo, H.~C., and Feizi, S.
\newblock Benchmarking deep learning interpretability in time series
  predictions.
\newblock In \emph{NeurIPS}, 2020.

\bibitem[Ismail et~al.(2021)Ismail, Corrada~Bravo, and Feizi]{aya}
Ismail, A.~A., Corrada~Bravo, H., and Feizi, S.
\newblock Improving deep learning interpretability by saliency guided training.
\newblock \emph{Advances in Neural Information Processing Systems}, 34, 2021.

\bibitem[Koh \& Liang(2017)Koh and Liang]{pangweikoh2021}
Koh, P.~W. and Liang, P.
\newblock Understanding black-box predictions via influence functions.
\newblock In \emph{Proceedings of the 34th International Conference on Machine
  Learning - Volume 70}, ICML'17, pp.\  1885–1894. JMLR.org, 2017.

\bibitem[Krueger et~al.(2021)Krueger, Caballero, Jacobsen, Zhang, Binas, Priol,
  and Courville]{Krueger2021OutofDistributionGV}
Krueger, D., Caballero, E., Jacobsen, J.-H., Zhang, A., Binas, J., Priol,
  R.~L., and Courville, A.~C.
\newblock Out-of-distribution generalization via risk extrapolation (rex).
\newblock In \emph{ICML}, 2021.

\bibitem[Lipton et~al.(2018)Lipton, Wang, and Smola]{pmlr-v80-lipton18a}
Lipton, Z., Wang, Y.-X., and Smola, A.
\newblock Detecting and correcting for label shift with black box predictors.
\newblock In Dy, J. and Krause, A. (eds.), \emph{Proceedings of the 35th
  International Conference on Machine Learning}, volume~80 of \emph{Proceedings
  of Machine Learning Research}, pp.\  3122--3130. PMLR, 10--15 Jul 2018.
\newblock URL \url{https://proceedings.mlr.press/v80/lipton18a.html}.

\bibitem[Liu et~al.(2021)Liu, Lin, Cao, Hu, Wei, Zhang, Lin, and Guo]{swin}
Liu, Z., Lin, Y., Cao, Y., Hu, H., Wei, Y., Zhang, Z., Lin, S., and Guo, B.
\newblock Swin transformer: Hierarchical vision transformer using shifted
  windows.
\newblock \emph{International Conference on Computer Vision (ICCV)}, 2021.

\bibitem[Madry et~al.(2018)Madry, Makelov, Schmidt, Tsipras, and
  Vladu]{Madry2017TowardsDL}
Madry, A., Makelov, A., Schmidt, L., Tsipras, D., and Vladu, A.
\newblock Towards deep learning models resistant to adversarial attacks.
\newblock In \emph{ICLR}, 2018.

\bibitem[Mahajan et~al.(2021)Mahajan, Tople, and Sharma]{mahajan2020domain}
Mahajan, D., Tople, S., and Sharma, A.
\newblock Domain generalization using causal matching.
\newblock In \emph{ICML}, 2021.

\bibitem[Mahendran \& Vedaldi(2015)Mahendran and Vedaldi]{mahendran15}
Mahendran, A. and Vedaldi, A.
\newblock Understanding deep image representations by inverting them.
\newblock In \emph{2015 IEEE Conference on Computer Vision and Pattern
  Recognition (CVPR)}, pp.\  5188--5196, 2015.
\newblock \doi{10.1109/CVPR.2015.7299155}.

\bibitem[Mahendran \& Vedaldi(2016)Mahendran and Vedaldi]{MahendranV15}
Mahendran, A. and Vedaldi, A.
\newblock Visualizing deep convolutional neural networks using natural
  pre-images.
\newblock \emph{International Journal of Computer Vision}, 120:\penalty0
  233--255, 2016.

\bibitem[Miller(1995)]{Miller95wordnet}
Miller, G.~A.
\newblock Wordnet: A lexical database for english.
\newblock \emph{COMMUNICATIONS OF THE ACM}, 38:\penalty0 39--41, 1995.

\bibitem[Moayeri et~al.(2022)Moayeri, Pope, Balaji, and Feizi]{mazda}
Moayeri, M., Pope, P., Balaji, Y., and Feizi, S.
\newblock A comprehensive study of image classification model sensitivity to
  foregrounds, backgrounds, and visual attributes, 2022.

\bibitem[Muandet et~al.(2013)Muandet, Balduzzi, and
  Schölkopf]{pmlr-v28-muandet13}
Muandet, K., Balduzzi, D., and Schölkopf, B.
\newblock Domain generalization via invariant feature representation.
\newblock In Dasgupta, S. and McAllester, D. (eds.), \emph{Proceedings of the
  30th International Conference on Machine Learning}, volume~28 of
  \emph{Proceedings of Machine Learning Research}, pp.\  10--18, Atlanta,
  Georgia, USA, 17--19 Jun 2013. PMLR.
\newblock URL \url{https://proceedings.mlr.press/v28/muandet13.html}.

\bibitem[Nguyen et~al.(2015)Nguyen, Yosinski, and Clune]{NguyenYC14}
Nguyen, A., Yosinski, J., and Clune, J.
\newblock Deep neural networks are easily fooled: High confidence predictions
  for unrecognizable images.
\newblock In \emph{2015 IEEE Conference on Computer Vision and Pattern
  Recognition (CVPR)}, pp.\  427--436, 2015.
\newblock \doi{10.1109/CVPR.2015.7298640}.

\bibitem[Nguyen et~al.(2016)Nguyen, Yosinski, and Clune]{NguyenYC16}
Nguyen, A.~M., Yosinski, J., and Clune, J.
\newblock Multifaceted feature visualization: Uncovering the different types of
  features learned by each neuron in deep neural networks.
\newblock In \emph{ICML Workshop on Visualization for Deep Learning}, 2016.

\bibitem[Nushi et~al.(2018)Nushi, Kamar, and Horvitz]{pandora1809}
Nushi, B., Kamar, E., and Horvitz, E.
\newblock Towards accountable {AI:} hybrid human-machine analyses for
  characterizing system failure.
\newblock In Chen, Y. and Kazai, G. (eds.), \emph{Proceedings of the Sixth AAAI
  Conference on Human Computation and Crowdsourcing, HCOMP}, pp.\  126--135.
  {AAAI} Press, 2018.
\newblock URL
  \url{https://aaai.org/ocs/index.php/HCOMP/HCOMP18/paper/view/17930}.

\bibitem[Olah et~al.(2018)Olah, Satyanarayan, Johnson, Carter, Schubert, Ye,
  and Mordvintsev]{olah2018the}
Olah, C., Satyanarayan, A., Johnson, I., Carter, S., Schubert, L., Ye, K., and
  Mordvintsev, A.
\newblock The building blocks of interpretability.
\newblock \emph{Distill}, 2018.
\newblock \doi{10.23915/distill.00010}.
\newblock https://distill.pub/2018/building-blocks.

\bibitem[O'Shaughnessy et~al.(2019)O'Shaughnessy, Canal, Connor, Davenport, and
  Rozell]{oshaughnessy2020generative}
O'Shaughnessy, M., Canal, G., Connor, M., Davenport, M., and Rozell, C.
\newblock Generative causal explanations of black-box classifiers.
\newblock In \emph{NeurIPS}, 2019.

\bibitem[Peters et~al.(2016)Peters, Bühlmann, and
  Meinshausen]{noauthororeditor}
Peters, J., Bühlmann, P., and Meinshausen, N.
\newblock Causal inference by using invariant prediction: identification and
  confidence intervals.
\newblock 2016.
\newblock URL
  \url{https://rss.onlinelibrary.wiley.com/doi/full/10.1111/rssb.12167}.

\bibitem[Radford et~al.(2021)Radford, Kim, Hallacy, Ramesh, Goh, Agarwal,
  Sastry, Askell, Mishkin, Clark, et~al.]{clip}
Radford, A., Kim, J.~W., Hallacy, C., Ramesh, A., Goh, G., Agarwal, S., Sastry,
  G., Askell, A., Mishkin, P., Clark, J., et~al.
\newblock Learning transferable visual models from natural language
  supervision.
\newblock \emph{arXiv preprint arXiv:2103.00020}, 2021.

\bibitem[Rahimian \& Mehrotra(2019)Rahimian and
  Mehrotra]{Rahimian2019DistributionallyRO}
Rahimian, H. and Mehrotra, S.
\newblock Distributionally robust optimization: A review.
\newblock \emph{ArXiv}, abs/1908.05659, 2019.

\bibitem[Ribeiro et~al.(2016)Ribeiro, Singh, and Guestrin]{ribeiro2016}
Ribeiro, M.~T., Singh, S., and Guestrin, C.
\newblock "why should i trust you?": Explaining the predictions of any
  classifier.
\newblock In \emph{Proceedings of the 22nd ACM SIGKDD International Conference
  on Knowledge Discovery and Data Mining}, KDD '16, pp.\  1135–1144, New
  York, NY, USA, 2016. Association for Computing Machinery.
\newblock ISBN 9781450342322.
\newblock \doi{10.1145/2939672.2939778}.
\newblock URL \url{https://doi.org/10.1145/2939672.2939778}.

\bibitem[Rosenfeld et~al.(2021)Rosenfeld, Ravikumar, and
  Risteski]{rosenfeld2021the}
Rosenfeld, E., Ravikumar, P.~K., and Risteski, A.
\newblock The risks of invariant risk minimization.
\newblock In \emph{International Conference on Learning Representations}, 2021.
\newblock URL \url{https://openreview.net/forum?id=BbNIbVPJ-42}.

\bibitem[Rosenfeld et~al.(2022)Rosenfeld, Ravikumar, and
  Risteski]{Rosenfeld2021AnOL}
Rosenfeld, E., Ravikumar, P., and Risteski, A.
\newblock An online learning approach to interpolation and extrapolation in
  domain generalization.
\newblock 2022.

\bibitem[Sagawa* et~al.(2020)Sagawa*, Koh*, Hashimoto, and
  Liang]{Sagawa*2020Distributionally}
Sagawa*, S., Koh*, P.~W., Hashimoto, T.~B., and Liang, P.
\newblock Distributionally robust neural networks.
\newblock In \emph{International Conference on Learning Representations}, 2020.
\newblock URL \url{https://openreview.net/forum?id=ryxGuJrFvS}.

\bibitem[Salman et~al.(2020)Salman, Ilyas, Engstrom, Kapoor, and
  Madry]{robust_resnets}
Salman, H., Ilyas, A., Engstrom, L., Kapoor, A., and Madry, A.
\newblock Do adversarially robust imagenet models transfer better?
\newblock In \emph{ArXiv preprint arXiv:2007.08489}, 2020.

\bibitem[Sandler et~al.(2018)Sandler, Howard, Zhu, Zhmoginov, and
  Chen]{mobilenetv2}
Sandler, M., Howard, A., Zhu, M., Zhmoginov, A., and Chen, L.-C.
\newblock Mobilenetv2: Inverted residuals and linear bottlenecks.
\newblock In \emph{Proceedings of the IEEE conference on computer vision and
  pattern recognition}, pp.\  4510--4520, 2018.

\bibitem[Selvaraju et~al.(2019)Selvaraju, Das, Vedantam, Cogswell, Parikh, and
  Batra]{SelvarajuDVCPB16}
Selvaraju, R.~R., Das, A., Vedantam, R., Cogswell, M., Parikh, D., and Batra,
  D.
\newblock Grad-cam: Visual explanations from deep networks via gradient-based
  localization.
\newblock \emph{International Journal of Computer Vision}, 128:\penalty0
  336--359, 2019.

\bibitem[Simonyan \& Zisserman(2014)Simonyan and Zisserman]{vgg}
Simonyan, K. and Zisserman, A.
\newblock Very deep convolutional networks for large-scale image recognition.
\newblock \emph{arXiv preprint arXiv:1409.1556}, 2014.

\bibitem[Simonyan et~al.(2014)Simonyan, Vedaldi, and
  Zisserman]{Simonyan2013DeepIC}
Simonyan, K., Vedaldi, A., and Zisserman, A.
\newblock Deep inside convolutional networks: Visualising image classification
  models and saliency maps.
\newblock In \emph{Workshop at International Conference on Learning
  Representations}, 2014.

\bibitem[Simpson et~al.(2019)Simpson, Dutil, Bengio, and
  Cohen]{simpson2019gradmask}
Simpson, B., Dutil, F., Bengio, Y., and Cohen, J.~P.
\newblock Gradmask: Reduce overfitting by regularizing saliency, 2019.

\bibitem[Singla \& Feizi(2022)Singla and Feizi]{salientimagenet2021}
Singla, S. and Feizi, S.
\newblock Salient imagenet: How to discover spurious features in deep learning?
\newblock In \emph{International Conference on Learning Representations}, 2022.
\newblock URL \url{https://openreview.net/forum?id=XVPqLyNxSyh}.

\bibitem[Singla et~al.(2019)Singla, Wallace, Feng, and
  Feizi]{Singla2019UnderstandingIO}
Singla, S., Wallace, E., Feng, S., and Feizi, S.
\newblock Understanding impacts of high-order loss approximations and features
  in deep learning interpretation.
\newblock In \emph{ICML}, 2019.

\bibitem[Singla et~al.(2021)Singla, Nushi, Shah, Kamar, and
  Horvitz]{singlaCVPR2021}
Singla, S., Nushi, B., Shah, S., Kamar, E., and Horvitz, E.
\newblock Understanding failures of deep networks via robust feature
  extraction.
\newblock In \emph{The IEEE Conference on Computer Vision and Pattern
  Recognition (CVPR)}, 2021.

\bibitem[Smilkov et~al.(2017)Smilkov, Thorat, Kim, Vi{\'{e}}gas, and
  Wattenberg]{Smilkov2017SmoothGradRN}
Smilkov, D., Thorat, N., Kim, B., Vi{\'{e}}gas, F.~B., and Wattenberg, M.
\newblock Smoothgrad: removing noise by adding noise.
\newblock In \emph{ICML Workshop on Visualization for Deep Learning}, 2017.

\bibitem[Sturmfels et~al.(2020)Sturmfels, Lundberg, and
  Lee]{sturmfels2020visualizing}
Sturmfels, P., Lundberg, S., and Lee, S.-I.
\newblock Visualizing the impact of feature attribution baselines.
\newblock \emph{Distill}, 2020.
\newblock \doi{10.23915/distill.00022}.
\newblock https://distill.pub/2020/attribution-baselines.

\bibitem[Sundararajan et~al.(2017)Sundararajan, Taly, and
  Yan]{Sundararajan2017AxiomaticAF}
Sundararajan, M., Taly, A., and Yan, Q.
\newblock Axiomatic attribution for deep networks.
\newblock In \emph{ICML}, 2017.

\bibitem[Szegedy et~al.(2017)Szegedy, Ioffe, Vanhoucke, and Alemi]{inceptionv2}
Szegedy, C., Ioffe, S., Vanhoucke, V., and Alemi, A.~A.
\newblock Inception-v4, inception-resnet and the impact of residual connections
  on learning.
\newblock In \emph{Thirty-first AAAI conference on artificial intelligence},
  2017.

\bibitem[Tan \& Le(2019)Tan and Le]{efficientnet}
Tan, M. and Le, Q.
\newblock Efficientnet: Rethinking model scaling for convolutional neural
  networks.
\newblock In \emph{International Conference on Machine Learning}, pp.\
  6105--6114. PMLR, 2019.

\bibitem[Tan et~al.(2019)Tan, Chen, Pang, Vasudevan, Sandler, Howard, and
  Le]{mnasnet}
Tan, M., Chen, B., Pang, R., Vasudevan, V., Sandler, M., Howard, A., and Le,
  Q.~V.
\newblock Mnasnet: Platform-aware neural architecture search for mobile.
\newblock In \emph{Proceedings of the IEEE/CVF Conference on Computer Vision
  and Pattern Recognition}, pp.\  2820--2828, 2019.

\bibitem[Tian \& Pearl(2001)Tian and Pearl]{10.5555/2074022.2074085}
Tian, J. and Pearl, J.
\newblock Causal discovery from changes.
\newblock In \emph{Proceedings of the Seventeenth Conference on Uncertainty in
  Artificial Intelligence}, UAI'01, pp.\  512–521, San Francisco, CA, USA,
  2001. Morgan Kaufmann Publishers Inc.
\newblock ISBN 1558608001.

\bibitem[Touvron et~al.(2021)Touvron, Cord, Douze, Massa, Sablayrolles, and
  J{\'e}gou]{deit}
Touvron, H., Cord, M., Douze, M., Massa, F., Sablayrolles, A., and J{\'e}gou,
  H.
\newblock Training data-efficient image transformers \& distillation through
  attention.
\newblock In \emph{International Conference on Machine Learning}, pp.\
  10347--10357. PMLR, 2021.

\bibitem[Tsipras et~al.(2018)Tsipras, Santurkar, Engstrom, Turner, and
  Madry]{tsipras2019robustness}
Tsipras, D., Santurkar, S., Engstrom, L., Turner, A., and Madry, A.
\newblock Robustness may be at odds with accuracy.
\newblock In \emph{ICLR}, 2018.

\bibitem[Tsipras et~al.(2020)Tsipras, Santurkar, Engstrom, Ilyas, and
  Madry]{tsipras2020imagenet}
Tsipras, D., Santurkar, S., Engstrom, L., Ilyas, A., and Madry, A.
\newblock From imagenet to image classification: Contextualizing progress on
  benchmarks.
\newblock In \emph{ICML}, 2020.

\bibitem[Verma et~al.(2020)Verma, Dickerson, and
  Hines]{verma2020counterfactual}
Verma, S., Dickerson, J., and Hines, K.
\newblock Counterfactual explanations for machine learning: A review, 2020.

\bibitem[Widmer \& Kub{\'a}t(2004)Widmer and Kub{\'a}t]{Widmer2004LearningIT}
Widmer, G. and Kub{\'a}t, M.
\newblock Learning in the presence of concept drift and hidden contexts.
\newblock \emph{Machine Learning}, 23:\penalty0 69--101, 2004.

\bibitem[Wightman(2019)]{timm}
Wightman, R.
\newblock Pytorch image models.
\newblock \url{https://github.com/rwightman/pytorch-image-models}, 2019.

\bibitem[Wong et~al.(2020)Wong, Rice, and Kolter]{wong_fast}
Wong, E., Rice, L., and Kolter, J.~Z.
\newblock Fast is better than free: Revisiting adversarial training.
\newblock \emph{CoRR}, abs/2001.03994, 2020.
\newblock URL \url{https://arxiv.org/abs/2001.03994}.

\bibitem[Wong et~al.(2021)Wong, Santurkar, and Madry]{sparsewong21b}
Wong, E., Santurkar, S., and Madry, A.
\newblock Leveraging sparse linear layers for debuggable deep networks.
\newblock In Meila, M. and Zhang, T. (eds.), \emph{Proceedings of the 38th
  International Conference on Machine Learning}, volume 139 of
  \emph{Proceedings of Machine Learning Research}, pp.\  11205--11216. PMLR,
  18--24 Jul 2021.
\newblock URL \url{https://proceedings.mlr.press/v139/wong21b.html}.

\bibitem[Xiao et~al.(2021)Xiao, Engstrom, Ilyas, and Madry]{xiao2021noise}
Xiao, K.~Y., Engstrom, L., Ilyas, A., and Madry, A.
\newblock Noise or signal: The role of image backgrounds in object recognition.
\newblock In \emph{International Conference on Learning Representations}, 2021.
\newblock URL \url{https://openreview.net/forum?id=gl3D-xY7wLq}.

\bibitem[Xie et~al.(2020)Xie, Chen, Liu, and Li]{Xie2020RiskVP}
Xie, C., Chen, F., Liu, Y., and Li, Z.
\newblock Risk variance penalization: From distributional robustness to
  causality.
\newblock \emph{ArXiv}, abs/2006.07544, 2020.

\bibitem[Xie et~al.(2017)Xie, Girshick, Doll{\'a}r, Tu, and He]{resnext}
Xie, S., Girshick, R., Doll{\'a}r, P., Tu, Z., and He, K.
\newblock Aggregated residual transformations for deep neural networks.
\newblock In \emph{Proceedings of the IEEE conference on computer vision and
  pattern recognition}, pp.\  1492--1500, 2017.

\bibitem[Yeh et~al.(2019)Yeh, Hsieh, Suggala, Inouye, and
  Ravikumar]{Yeh2019OnT}
Yeh, C.-K., Hsieh, C.-Y., Suggala, A.~S., Inouye, D.~I., and Ravikumar, P.~D.
\newblock On the (in)fidelity and sensitivity of explanations.
\newblock In \emph{NeurIPS}, 2019.

\bibitem[Yosinski et~al.(2016)Yosinski, Clune, Nguyen, Fuchs, and
  Lipson]{YosinskiCNFL15}
Yosinski, J., Clune, J., Nguyen, A.~M., Fuchs, T.~J., and Lipson, H.
\newblock Understanding neural networks through deep visualization.
\newblock In \emph{ICML Deep Learning Workshop}, 2016.

\bibitem[Zagoruyko \& Komodakis(2016)Zagoruyko and Komodakis]{wide_resnet}
Zagoruyko, S. and Komodakis, N.
\newblock Wide residual networks.
\newblock In \emph{BMVC}, 2016.

\bibitem[Zech et~al.(2018)Zech, Badgeley, Liu, Costa, Titano, and
  Oermann]{Zech2018VariableGP}
Zech, J.~R., Badgeley, M.~A., Liu, M., Costa, A.~B., Titano, J.~J., and
  Oermann, E.~K.
\newblock Variable generalization performance of a deep learning model to
  detect pneumonia in chest radiographs: A cross-sectional study.
\newblock \emph{PLoS Medicine}, 15, 2018.

\bibitem[Zeiler \& Fergus(2014)Zeiler and Fergus]{ZeilerF13}
Zeiler, M.~D. and Fergus, R.
\newblock Visualizing and understanding convolutional networks.
\newblock In \emph{ECCV}, 2014.

\bibitem[Zhang et~al.(2018)Zhang, Wang, Molino, Li, and
  Ebert]{zhang2018manifold}
Zhang, J., Wang, Y., Molino, P., Li, L., and Ebert, D.~S.
\newblock Manifold: A model-agnostic framework for interpretation and diagnosis
  of machine learning models.
\newblock \emph{IEEE transactions on visualization and computer graphics},
  25\penalty0 (1):\penalty0 364--373, 2018.

\bibitem[Zhou et~al.(2016)Zhou, Khosla, A., Oliva, and Torralba]{ZhouKLOT15}
Zhou, B., Khosla, A., A., L., Oliva, A., and Torralba, A.
\newblock {Learning Deep Features for Discriminative Localization.}
\newblock In \emph{The IEEE Conference on Computer Vision and Pattern
  Recognition (CVPR)}, 2016.

\bibitem[Zhou et~al.(2018)Zhou, Bau, Oliva, and Torralba]{zhou2018interpreting}
Zhou, B., Bau, D., Oliva, A., and Torralba, A.
\newblock Interpreting deep visual representations via network dissection.
\newblock \emph{IEEE transactions on pattern analysis and machine
  intelligence}, 41\penalty0 (9):\penalty0 2131--2145, 2018.

\end{thebibliography}
\bibliographystyle{icml2022}

%%%%%%%%%%%%%%%%%%%%%%%%%%%%%%%%%%%%%%%%%%%%%%%%%%%%%%%%%%%%%%%%%%%%%%%%%%%%%%%
%%%%%%%%%%%%%%%%%%%%%%%%%%%%%%%%%%%%%%%%%%%%%%%%%%%%%%%%%%%%%%%%%%%%%%%%%%%%%%%
% APPENDIX
%%%%%%%%%%%%%%%%%%%%%%%%%%%%%%%%%%%%%%%%%%%%%%%%%%%%%%%%%%%%%%%%%%%%%%%%%%%%%%%
%%%%%%%%%%%%%%%%%%%%%%%%%%%%%%%%%%%%%%%%%%%%%%%%%%%%%%%%%%%%%%%%%%%%%%%%%%%%%%%
\newpage
\appendix
\onecolumn

\noindent {\bf \LARGE Appendix}

\section{Details on Pretrained Models}\label{sec:appendix_pretrained_models}

\begin{table}[ht!]
    \centering
    \begin{tabular}{|l|c|cc|cc|c|}\toprule
Model & Clean Accuracy & Core Accuracy & (Drop) & Spurious Accuracy &(Drop) & RCS \\ \midrule
ConViT-T/16 & 87.23\% & 84.30 \% & (-2.92\%)& 76.95\%  & (-10.27\%) & 18.97 \\
ConViT-S/16 & 92.85\% & 91.62 \% & (-1.23\%)& 88.16\%  & (-4.69\%) & 17.11 \\
ConViT-B/16 & 94.83\% & 93.83 \% & (-1.00\%)& 91.07\%  & (-3.76\%) & 18.29 \\
DeiT-T/16 & 86.81\% & 83.45 \% & (-3.36\%)& 75.26\%  & (-11.55\%) & 19.83 \\
DeiT-S/16 & 93.06\% & 91.44 \% & (-1.61\%)& 87.06\%  & (-5.99\%) & 20.38 \\
DeiT-B/16 & 95.76\% & 94.62 \% & (-1.13\%)& 91.61\%  & (-4.14\%) & 21.85 \\
Swin-B/4 Window7 & 92.93\% & 91.58 \% & (-1.35\%)& 88.19\%  & (-4.73\%) & 16.73 \\
Swin-S/4 Window7 & 93.58\% & 93.12 \% & (-0.46\%)& 89.22\%  & (-4.36\%) & 22.07 \\
Swin-T/4 Window7 & 92.19\% & 91.45 \% & (-0.74\%)& 86.33\%  & (-5.86\%) & 23.05 \\
ViT-T/16 & 61.06\% & 36.81 \% & (-24.25\%)& 11.73\%  & (-49.33\%) & 51.66 \\
ViT-S/16 & 87.59\% & 79.24 \% & (-8.35\%)& 60.43\%  & (-27.16\%) & 31.18 \\
ViT-S/32 & 80.65\% & 70.46 \% & (-10.19\%)& 56.01\%  & (-24.64\%) & 19.65 \\
ViT-B/16 & 88.68\% & 83.51 \% & (-5.17\%)& 71.45\%  & (-17.23\%) & 26.76 \\
ViT-B/32 & 87.62\% & 82.55 \% & (-5.07\%)& 74.67\%  & (-12.95\%) & 18.42 \\
DINO ViTs8 & 92.05\% & 86.94 \% & (-5.12\%)& 74.02\%  & (-18.04\%) & 33.09 \\
DINO ViTs16 & 91.16\% & 85.33 \% & (-5.83\%)& 69.95\%  & (-21.21\%) & 34.39 \\
DINO ResNet50 & 87.03\% & 80.81 \% & (-6.22\%)& 60.79\%  & (-26.24\%) & 34.28 \\
MoCo-v3 ViT-S/ & 84.38\% & 73.67 \% & (-10.71\%)& 46.63\%  & (-37.75\%) & 33.93 \\
MoCo-v3 ViT-B/ & 87.38\% & 79.73 \% & (-7.65\%)& 63.34\%  & (-24.03\%) & 28.78 \\
MoCo-v3 ResNet50 & 85.37\% & 78.95 \% & (-6.42\%)& 62.45\%  & (-22.92\%) & 28.16 \\
CLIP ViT-B/16 & 76.62\% & 69.32 \% & (-7.31\%)& 58.92\%  & (-17.70\%) & 14.48 \\
CLIP ViT-B/32 & 73.08\% & 65.41 \% & (-7.67\%)& 55.83\%  & (-17.25\%) & 12.16 \\
CLIP ResNet50 & 68.71\% & 55.85 \% & (-12.85\%)& 31.18\%  & (-37.52\%) & 28.34 \\
CLIP ResNet101 & 70.93\% & 59.73 \% & (-11.20\%)& 38.00\%  & (-32.93\%) & 22.23 \\
EfficientNet B0 & 89.52\% & 88.23 \% & (-1.29\%)& 67.60\%  & (-21.92\%) & 46.70 \\
EfficientNet B4 & 87.18\% & 91.16 \% & (3.98\%)& 82.32\%  & (-4.85\%) & 33.31 \\
Inception V4 & 87.55\% & 81.93 \% & (-5.62\%)& 67.70\%  & (-19.85\%) & 28.25 \\
MnasNet A1 & 86.28\% & 82.87 \% & (-3.41\%)& 52.30\%  & (-33.98\%) & 47.15 \\
MobileNetv2 100 & 83.82\% & 81.15 \% & (-2.67\%)& 36.88\%  & (-46.94\%) & 54.02 \\
ResNet18 & 85.42\% & 78.69 \% & (-6.73\%)& 48.21\%  & (-37.21\%) & 41.69 \\
ResNet34 & 84.69\% & 85.12 \% & (0.43\%)& 45.00\%  & (-39.69\%) & 57.40 \\
ResNet50 & 92.22\% & 84.47 \% & (-7.75\%)& 57.59\%  & (-34.64\%) & 46.39 \\
Resnext50 32X4D & 88.07\% & 91.19 \% & (3.12\%)& 71.78\%  & (-16.28\%) & 52.41 \\
Robust ResNet18 Eps1 & 86.44\% & 75.37 \% & (-11.08\%)& 39.06\%  & (-47.39\%) & 42.43 \\
Robust ResNet18 Eps3 & 80.81\% & 72.81 \% & (-8.00\%)& 48.04\%  & (-32.77\%) & 31.30 \\
Robust ResNet50 Eps1 & 91.66\% & 84.56 \% & (-7.10\%)& 56.12\%  & (-35.53\%) & 47.93 \\
Robust ResNet50 Eps3 & 81.98\% & 76.90 \% & (-5.08\%)& 56.26\%  & (-25.72\%) & 30.88 \\
SimCLR ResNet50X1 & 87.13\% & 75.06 \% & (-12.07\%)& 53.46\%  & (-33.66\%) & 30.21 \\
Vgg19 & 91.03\% & 79.66 \% & (-11.37\%)& 42.17\%  & (-48.85\%) & 47.95 \\
Vgg19 Bn & 90.29\% & 82.44 \% & (-7.85\%)& 54.83\%  & (-35.46\%) & 44.02 \\
Wide ResNet50 & 93.09\% & 93.36 \% & (0.27\%)& 81.40\%  & (-11.68\%) & 47.36 \\
Wide ResNet101 & 91.53\% & 90.45 \% & (-1.08\%)& 71.18\%  & (-20.35\%) & 50.21 \\
 \bottomrule
    \end{tabular}
    \caption{Complete results for evaluation of several pretrained Imagenet models. We present core accuracy, spurious accuracy, and $\RCS$, as described in section \ref{sec:benchmarking}. All metrics are computed over the test set of Salient-Imagenet-1M.}
    \label{tab:pretrained_all}
\end{table}

In Section \ref{sec:benchmarking}, we present a framework for evaluating the reliance on spurious features of {\it any} model pretrained on ImageNet, and present results for a breadth of models. We now share greater detail on the models studied and the the results obtained in Table \ref{tab:pretrained_all}. For consistency, we obtain nearly all pretrained weights from the timm framework \cite{timm}, with the exception of self-supervised and CLIP model weights, which were obtained directly from the original sources, and adversarially trained networks, obtained from \cite{robust_resnets}. 

%\Mazda{Edit this mazda}
We study a large set of pretrained ImageNet models ($42$ total), spanning various architectures and training paradigms:

\textbf{Convolution-based models}: ResNets \cite{resnet}, Wide ResNets \cite{wide_resnet}, ResNext \cite{resnext}, Inceptionv2 \cite{inceptionv2}, VGG \cite{vgg}, Efficientnet \cite{efficientnet}, MNasNet \cite{mnasnet}, and MobileNetv2 \cite{mobilenetv2}. We refer to this group as ConvNets. 

\textbf{Vision Transformer-based models}: ViT \cite{vit}, DeiT \cite{deit}, ConViT \cite{convit}, and Swin Transformers \cite{swin}. We refer to this group as ViTs. 

\textbf{Robust ResNets}: Adversarially trained Resnets with $\ell_2$ projected gradient descent \cite{robust_resnets}. We refer to this group as Robust ResNets. 

\textbf{Self-supervised models}: MoCo-v3 \cite{MoCov3}, DINO \cite{dino}, and SimCLR \cite{simclr}, on both transformer and ResNet backbones, yielding the groups SS ViTs and SS ResNets. SS Models are evaluated with a linear layer trained atop fixed features with full supervision (as is standard practice). 

%For these models, we use encoders fixed from self-supervised pretraining, along with linear layers trained with supervision atop the fixed features to perform ImageNet classification, as is standard in evaluating self-supervised learners. 

\textbf{CLIP models}: Zero-shot classification models based on CLIP \cite{clip}, again both transformers and ResNets. We follow the zero-shot evaluation procedure of comparing the dot product of an image encoding to the average encoded vector of eight template text captions per class. 

\subsection{Additional Observations}

Model size has small and inconsistent effects on $\RCS$, which allows for comparing averages across categories with varying model sizes. The primary trend of transformers having lower $\RCS$ than ConvNets is validated in the smaller cohort of CLIP models, where the average $\RCS$ decreases from $0.25$ for CLIP ResNets to $0.13$ for CLIP ViTs. However, for self-supervised models MoCo-v3 and DINO, transformers and ResNets yield similar sensitivities. Interestingly, for ResNet backbones, self-supervised training decreases $\RCS$ (-$0.15$), while the reverse is true for transformers ($+0.10$). 

Surprisingly, we observe that {\it adversarial training in ResNets decreases $\RCS$} from $0.44$ to $0.38$, despite their objective of making models more reliable through increasing adversarial robustness. Further, the attack budget $\epsilon$ used during training, with higher attack budget leading to more robustness, validates our observed trend, as the gap in $\RCS$ between Robust Resnets and Resnets is larger for robust models with $\epsilon=3$ than for those with $\epsilon=1$.

Lastly, we find that increasing patch size in some ViTs leads to large drops in $\RCS$. Specifically, changing patchsize from 16 to 32 reduces $\RCS$ from $0.14$ to $0.12$ for CLIP ViTs, and from $0.29$ to $0.19$ for \cite{vit}'s original ViT.

\subsection{Conjectures}

We stress that detailed experiments are necessary to rigorously make further claims regarding our observations. However, we are intrigued by the low $\RCS$ in transformers. Further, the conflicting effect of self-supervision across architectures suggests there may be more factors at play. Vision transformers are emerging rapidly, and most transformer-based models (including DeiT, ConViT, and Swin) follow the training procedure of \citet{deit}, where data-efficiency was achieved via {\it heavy augmentation}. Augmentation is also used in self-supervised learning, to create multiple views of the same image. Seeing as many modern augmentations potentially corrupt core regions, we ponder if the increased gain in test accuracy may come at the cost of raising sensitivity to spurious regions. Further experimentation in this direction may be of interest. 

\clearpage

\section{Role of Spurious Features in Robust Resnet-50 Predictions}\label{sec:motivation_appendix}
%\Mazda{update text and images. Add math notation to clarify analysis}
\begin{figure}
    \centering
    \begin{minipage}{0.5\linewidth}
    \vspace{5mm}
    \centering
    \includegraphics[width=0.8\linewidth]{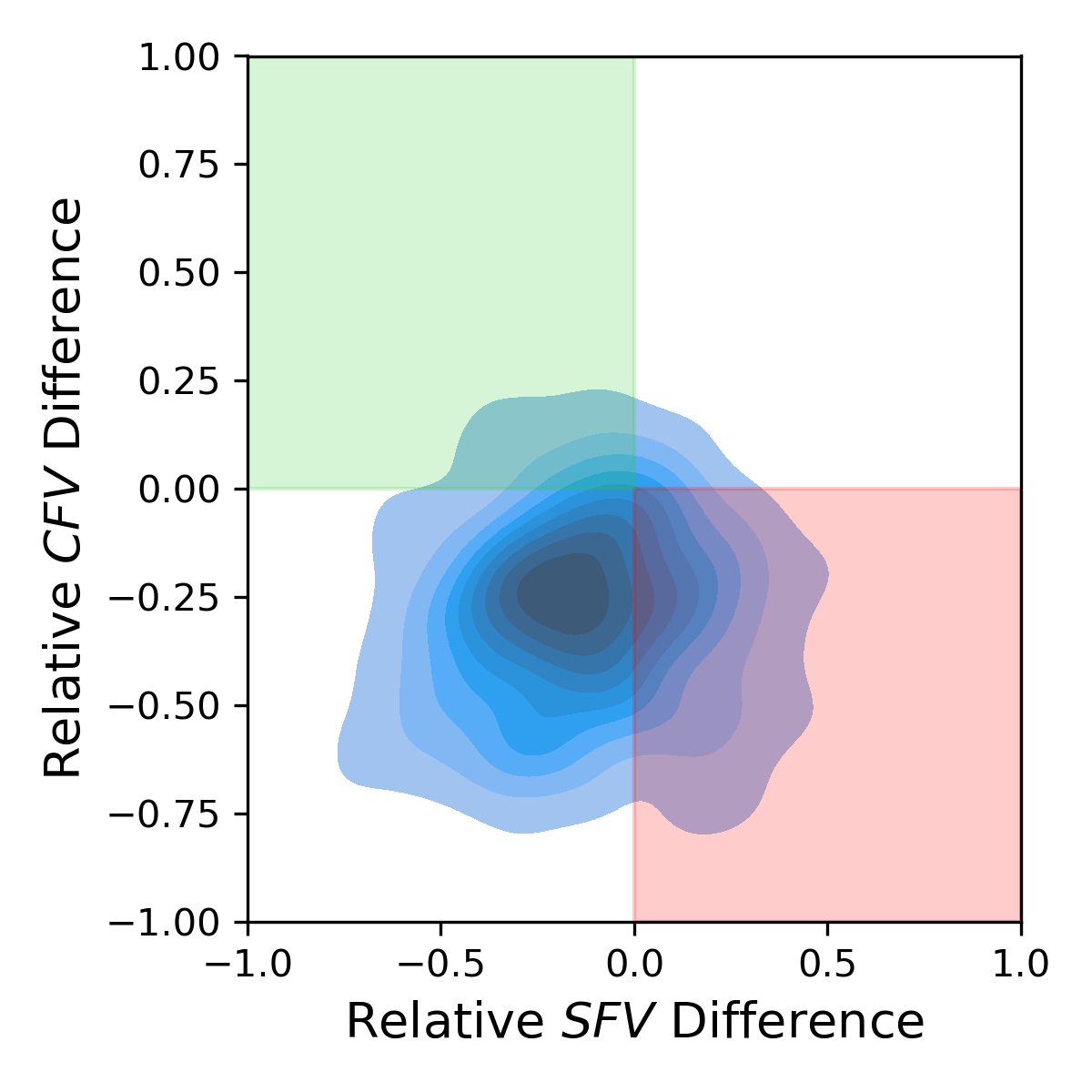}
    \end{minipage}
    \begin{minipage}{0.21\linewidth}
    \centering
    \begin{subfigure}{\linewidth}
    \centering
    \includegraphics[width=1.51\linewidth]{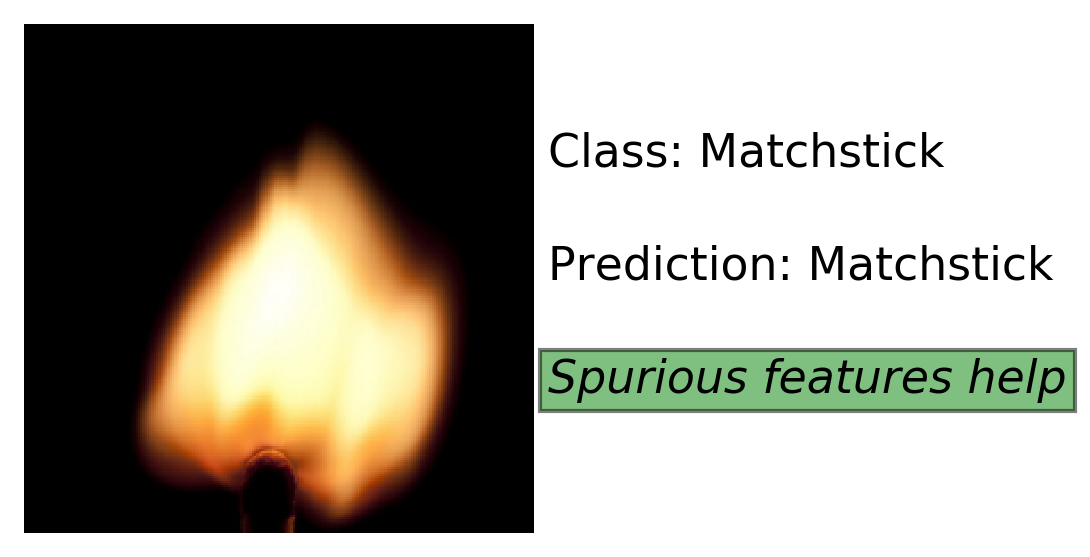}    
    \end{subfigure}
    \begin{subfigure}{\linewidth}
    \centering
    \includegraphics[width=1.6\linewidth]{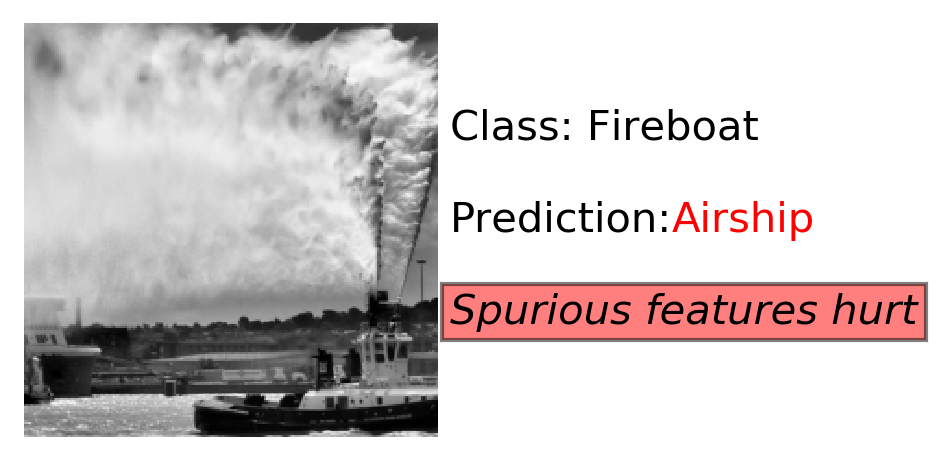}
    \end{subfigure}
    \end{minipage}
    \caption{Spurious correlations help and hurt classifiers. ({\bf Left}) Distribution per class of the relative difference in average core and spurious feature values between misclassified (MC) and correctly classified (CC) samples to the class. ({\bf Right}) Examples of classes where spurious features help correct classification (top) and where they lead to misclassification (bottom, original class: fireboat).}
    \label{fig:err_analysis}
\end{figure}

% \subsection{On the learning of Spurious Features}

% It is no secret that ERM provides no provisions to avoid the learning of spurious features. Many works have observed the reliance of state of the art classifiers on spurious features, finding that altering removal of foreground content 

% The issue of models learning to depend on spurious features has long been suspe

% \subsection{Error Analysis}
In this section, we motivate the study of spurious correlations through an error analysis. Importantly, this analysis makes no use of additive noise to gauge sensitivity, offering a distinct and complementary perspective to our other findings. Specifically, we use the annotations of neural nodes in the Robust Resnet-50 used to generate NAMs, directly inspecting feature activations. We first outline the intuition and key takeaways, before delving into detailed discussion of methods and results.

\subsection{The Matchstick Example: Spurious features can help or hurt, depending on class}

In Figure \ref{fig:err_analysis}, we display two samples where the spurious feature value of the predicted class is much higher, but core feature values are significantly lower, when compared to the average for correctly classified images of the class. Thus, the prediction is made due to the activations of the class' spurious features. For the matchstick, recognizing the spurious feature of the flame compensated for the fact that most of the matchstick lies outside of the image region, leading to a correct classification. However, the fireboat is misclassified as an airship, due to the spurious feature of what seems like a cloudy sky. 

The analysis of this section inspects core and spurious feature values for correct and incorrect classifications {\it on a classwise basis}. We find that spurious features hurt (i.e. activate higher for misclassified samples, even when core features correctly activate lower) much more often than they help (i.e. activate higher for true instances of the class, even when core features incorrectly activate lower). Specifically, for classes where the difference in spurious and core feature values between misclasssified and correctly classified samples have opposite sign ($101$ classes), the spurious features hurt $85\%$ of the time. 

This offers quantitative evidence to how spurious features lead to misclassification in a model trained with single-label supervision. However, we also demonstrate the how the role of spurious correlations varies, even from class to class. In summary, spurious features can be harmful, but true understanding of their roles requires careful analysis. We hope Salient Imagenet-1M opens the door to inquiries of this kind.

\subsection{Methodology for Error Analysis}

For an image $\image$ from class $\class$ with core features $\causal(\class)$ and spuriuos features $\spurious(\class)$, denote the representation vector (i.e. activations of neurons in penultimate layer of Robust Resnet-50) for $\image$ as $\br(\image)$. For a set $\mathcal{X}$, we measure the average activation of core features as follows: 

\begin{definition} {\bf (Core Feature Value)}
We define {\it core feature value} over a set of inputs $\mathcal{X}$, for both a class-feature pair $(\class, \feature)$ and a single class $\class$ (denoted $\CFV_{\class, \feature}$ and $\CFV_{\class}$ respectively) as follows:
\[
    \CFV_\class (\mathcal{X}) := \frac{1}{|\causal(\class)|} \sum_{\feature \in \causal(\class)} \CFV_{\class, \feature} (\mathcal{X}), \text{ with } \CFV_{\class, \feature}(\mathcal{X}) := \frac{1}{|\mathcal{X}|}\sum_{\image \in \mathcal{X}} \br_\feature (\image)
\] \label{def:core_ftr_val}
\end{definition}

We can analogously define {\it spurious feature value} for a class $\class$ and set $\mathcal{X}$ by replacing $\causal(\class)$ with $\spurious(\class)$ in \ref{def:core_ftr_val}. 

We now define groups $\mathcal{CC}_\class, \mathcal{MC}_\class$, corresponding to correctly and incorrectly classified samples {\it to} class $\class$ (i.e. by prediction). That is, with $h$ denoting the Robust Resnet-50, 
$$\mathcal{CC}_\class = \{\image \in \class | h(\image) = \class\}, \mathcal{MC}_\class = \{\image \not \in \class | h(\image) = \class\}$$

Using the above notation, we obtain the metrics presented in the figure \ref{fig:err_analysis}. Specifically, {\it Relative $\CFV$ Difference} refers to:

\begin{definition}{\bf (Relative $\CFV$ Difference)}
\[
\text{Relative } \CFV\text{ Difference} (\class) = \frac{\CFV_\class (\mathcal{MC}_\class) - \CFV_\class (\mathcal{CC}_\class)}{\max(\CFV_\class (\mathcal{MC}_\class), \CFV_\class (\mathcal{CC}_\class))}
\]
\end{definition}

This metric lies between $-1$ and $1$. When the relative $\CFV$ difference of a class $\class$ is positive, that entails that misclassified samples activate the core features for the class more than correctly classified samples. {\it Relative $\SFV$ Difference} is defined analogously by replacing $\CFV$ with $\SFV$.

For classes with at least one spurious and one core feature (324 total), we evaluate relative $\CFV$ difference and relative $\SFV$ difference, using images in the test set of Salient Imagenet-1M (though the analysis does not require core and spurious masks). We display the kernel destiny estimate of the distribution of (relative $\SFV$ difference, relative $\CFV$ difference) pairs in figure \ref{fig:err_analysis}.

\subsection{Findings}

For $84$ classes, relative $\SFV$ difference is positive while the relative $\CFV$ difference is negative (red quandrant). This suggests that the incorrect prediction of misclassified samples is due to high activation of spurious features, and not core features. Conversely, for $17$ classes, the reverse is true (green quadrant in fig. \ref{fig:err_analysis}), suggesting that core features erroneously activate higher for misclassifications. However, the spurious feature activations are higher for instances of the class, correcting the mistakes of the core features. Thus, {\bf spurious features can both help and hurt classifiers, and their role varies significantly based on class}. We highlight this result, as it reflects the complicated nature of spurious features, and their roles in deep models. We believe the complete removal of spurious features may have unintended consequences, and suggest careful analysis, with respect to class, when addressing spurious features. 

% In other words, the model's use of spurious features leads to misclassifications to that class based on the spurious features. For example, the fireboat is misclassified as an airship, likely due to the presence of gray, cloud-like water in the sky.

\section{Selecting the neural features highly predictive of the class}\label{sec:appendix_select_neurons}

The neural feature vector (i.e the vector of penultimate layer neurons) can have a large size ($2048$ for the Robust Resnet-50 used in this work and the prior work of \citet{salientimagenet2021}) and visualizing all of these features for a particular class (say class $\class$) to determine whether they are core or spurious for $\class$ can be difficult. Thus, we select a small subset of these features that are highly predictive of class $\class$ and annotate them as core or spurious for $i$ using the same method as used in the prior work of \citet{salientimagenet2021}.

We first select a subset of images (from the training set) on which the robust model predicts the class $\class$. We compute the mean of neural feature vectors across all images in this subset denoted by $\overline{\rep}(\class)$. From the weight matrix $\weight$ of the last linear layer of the robust model, we extract the $\class^{th}$ row $\weight_{i,:}$ that maps the neural feature vector to the logit for the class $\class$. Next, we compute the hadamard product $\overline{\rep}(\class) \odot \weight_{\class,:}$. Intuitively, the $\feature^{th}$ element of this vector $(\overline{\rep}(\class) \odot \weight_{\class,:})_{\feature}$ captures the mean contribution of neural feature $\feature$ for predicting the class $\class$. This procedure leads to the following definition:

\begin{definition}
\label{def:feature_importance}
{The Neural Feature Importance of feature $\feature$ for class $\class$ is defined as: $\importance_{i,j} = \left(\overline{\rep}(i) \odot  \weight_{i,:}\right)_{j}$. For class $\class$, the neural feature with $k^{th}$ highest $IV$ is said to have the feature rank $k$.}
%\begin{align*}
%\importance_{i,j} = \left(\overline{\rep}(i) \odot  \weight_{i,:}\right)_{j}. 
%\end{align*}
\end{definition}
We then select the neural features with the highest-$5$ importance values (defined above) per class.

\section{Visualizing the neural features of a robust model}\label{sec:appendix_viz_neurons}

\subsection{Heatmap}\label{sec:appendix_heatmap}
Heatmap is generated by first converting the \fm\ (which is grayscale) to an RGB image (using the jet colormap). This is followed by overlaying the jet colormap on top of the original image using the following lines of code:

\begin{lstlisting}[language=Python]
import cv2

def compute_heatmap(img, fam):
    hm = cv2.applyColorMap(np.uint8(255 * nam), 
        cv2.COLORMAP_JET)
    hm = np.float32(hm) / 255
    hm = hm + img
    hm = hm / np.max(hm)
    return hm
\end{lstlisting}

\begin{figure}[h!]
\centering
\includegraphics[trim=0cm 0cm 1.8cm 0.8cm, clip, width=0.8\linewidth]{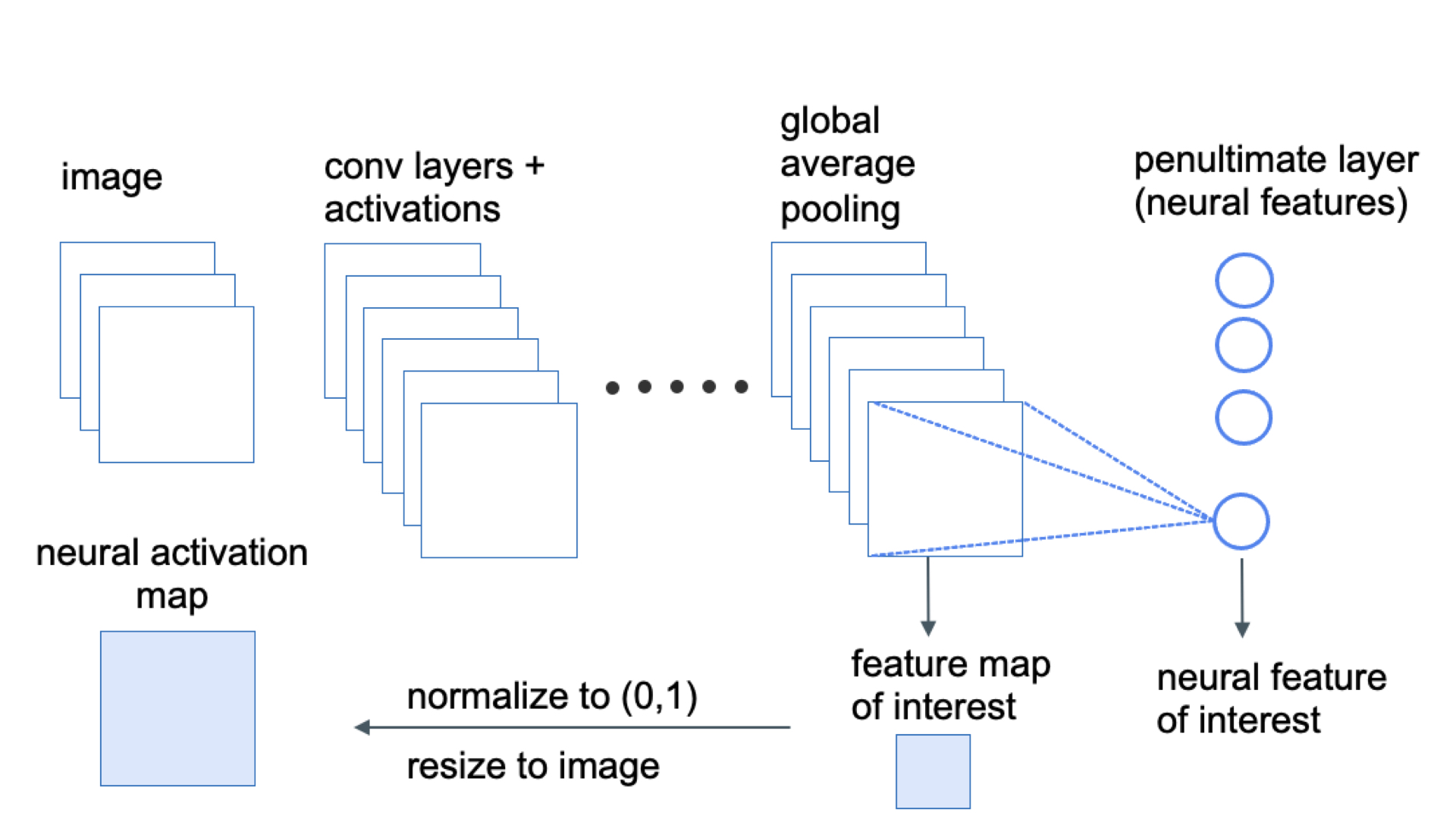}
\caption{Figure describing the Neural Activation Map generation procedure. To obtain the neural activation map for feature $\feature$, we select the feature map from the output of the tensor of the previous layer (i.e the layer before the global average pooling operation). Next, we simply normalize the feature map between 0 and 1 and resize the feature map to match the image size, giving the neural activation map. This figure is from \citet{salientimagenet2021} and included here for completeness.}
\label{fig:nam_procedure}
\end{figure}

\begin{figure}[h!]
\centering
\includegraphics[trim=0cm 1cm 2cm 0.8cm, clip, width=0.8\linewidth]{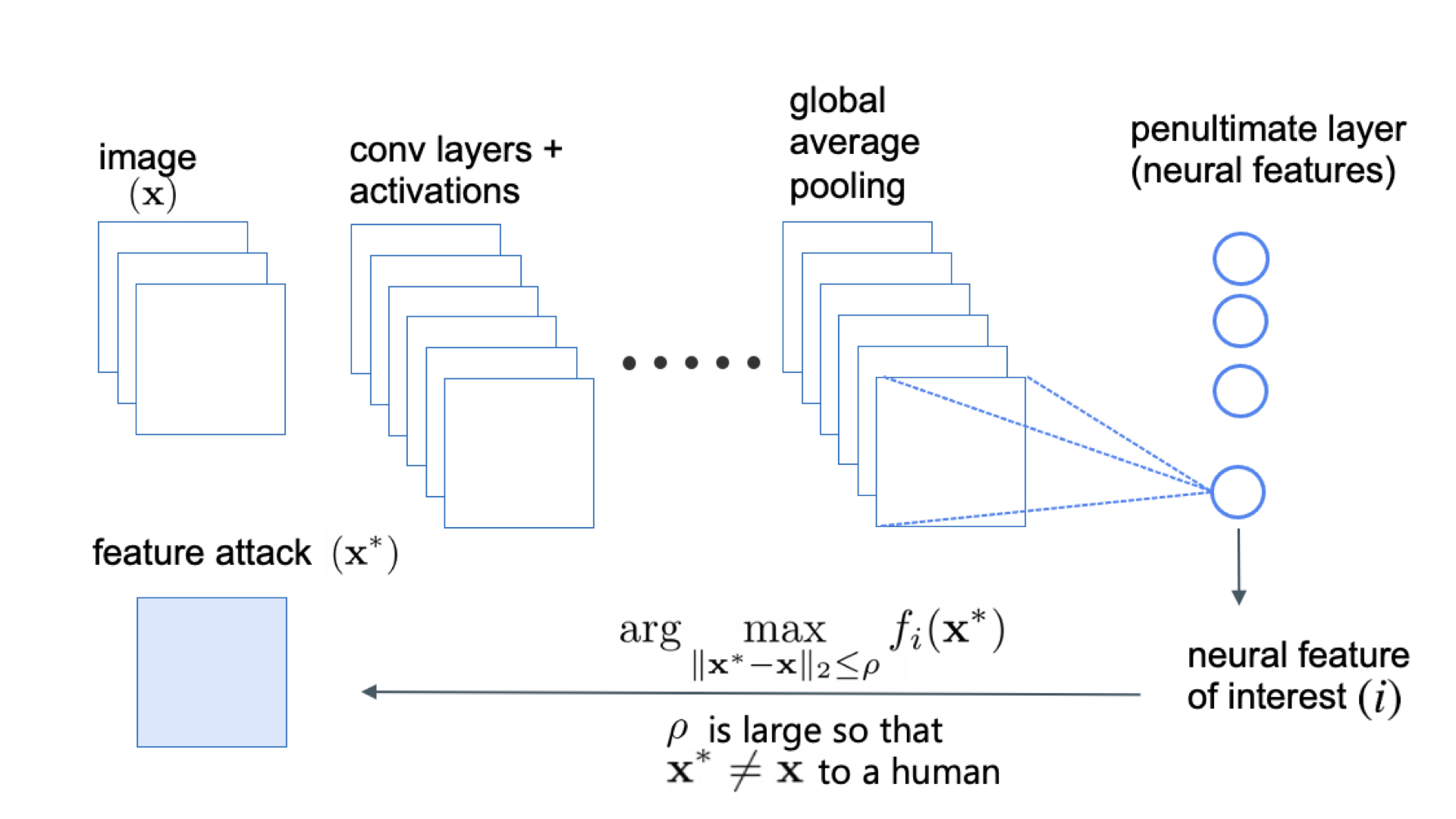}
\caption{Figure illustrating the feature attack procedure. We select the feature we are interested in and simply optimize the image to maximize its value to generate the visualization. $\rho$ is a hyperparameter used to control the amount of change allowed in the image. For optimization, we use gradient ascent with step size = $40$, number of iterations = $25$ and $\rho$ = $500$. This figure is from \citet{salientimagenet2021} and included here for completeness.}
\label{fig:feature_attack_procedure}
\end{figure}
\clearpage

\section{Mechanical Turk study for discovering spurious features}\label{sec:mturk_discovery_appendix}

\begin{figure}[h!]
\centering
\begin{subfigure}{0.4\linewidth}
\centering
\includegraphics[trim=0cm 0cm 0cm 0cm,  width=\linewidth]{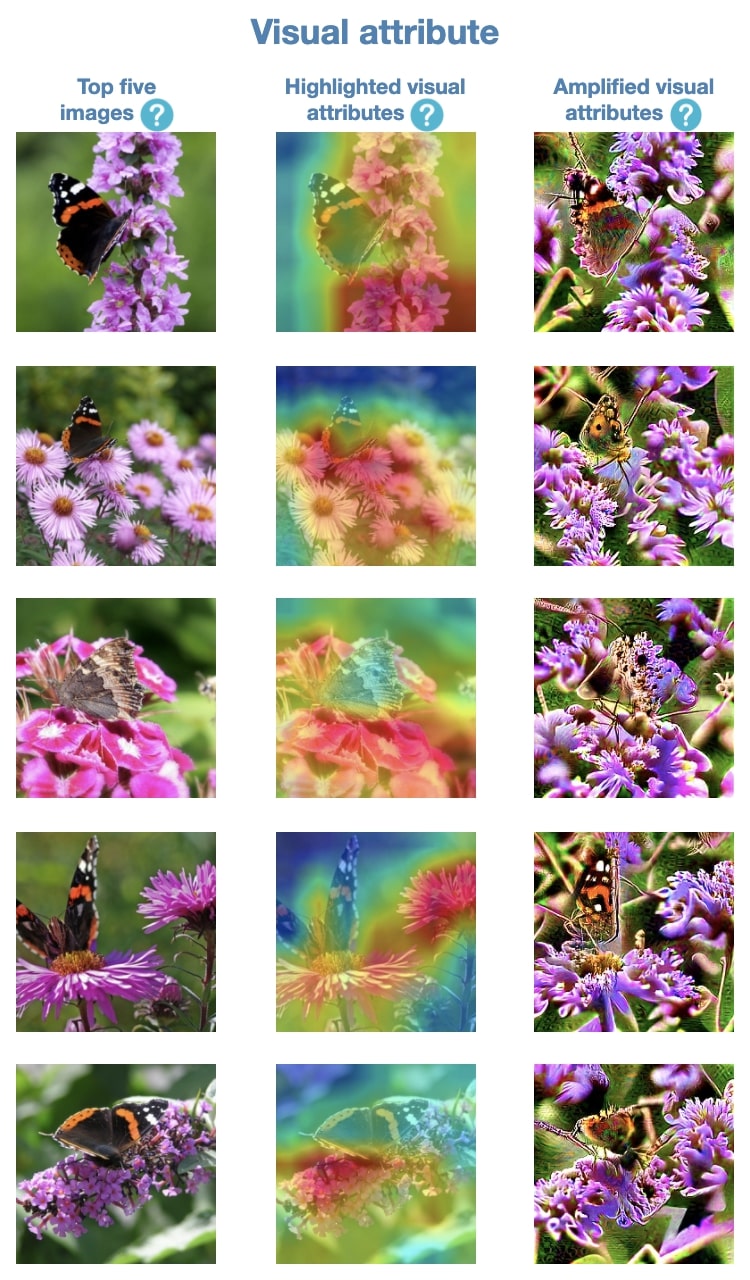}
\caption{Visual attribute}
\label{subfig:discover_spurious_leftpanel}
\end{subfigure}\ \ \ \ 
\begin{subfigure}{0.4\linewidth}
\centering
\includegraphics[trim=0cm 0cm 0cm 0cm,  width=\linewidth]{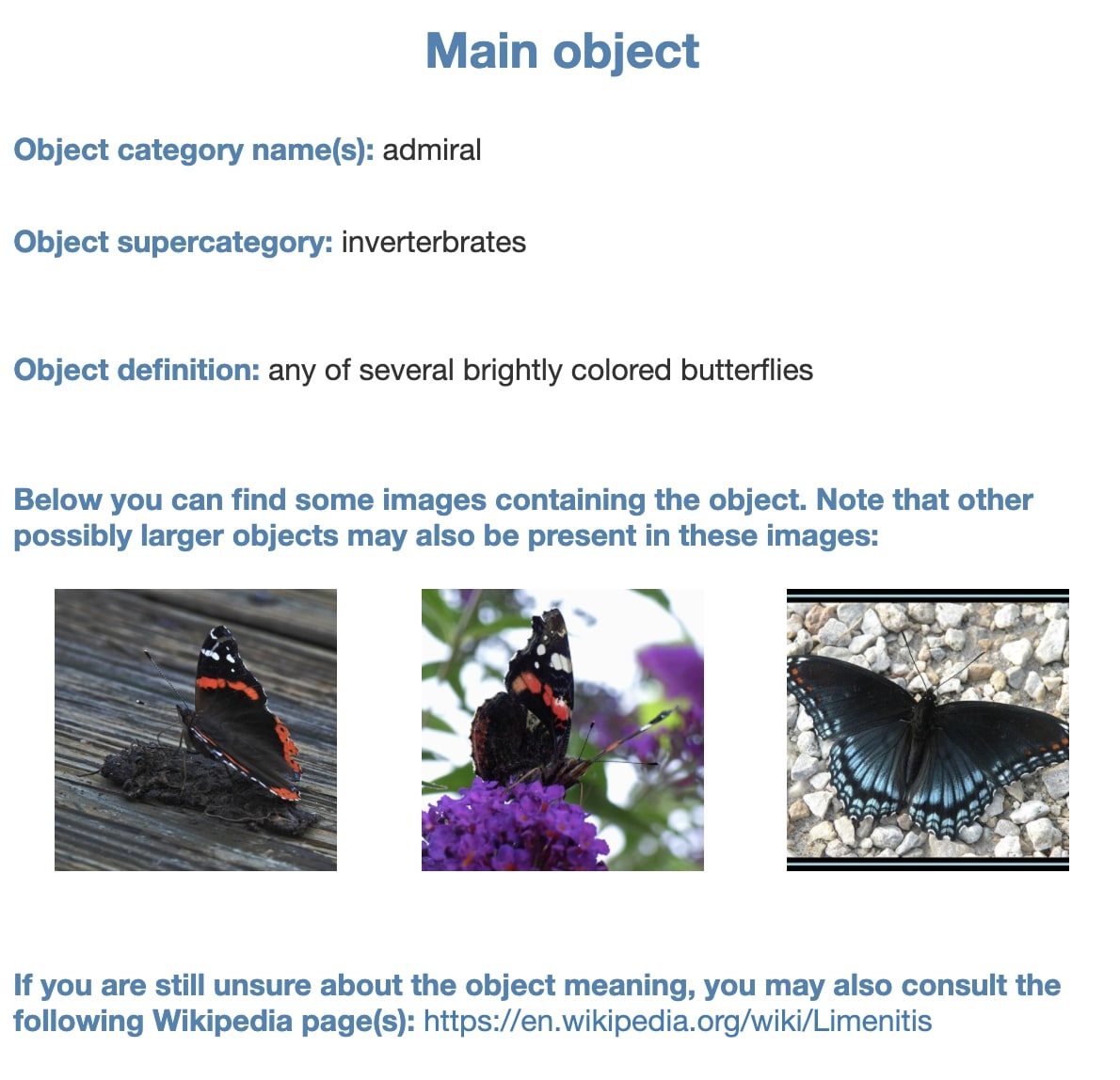}
\caption{Main object}
\label{subfig:discover_spurious_rightpanel}
\end{subfigure} \\~\\
\begin{subfigure}{0.8\linewidth}
\centering
\includegraphics[trim=0cm 0cm 0cm 0cm,  width=\linewidth]{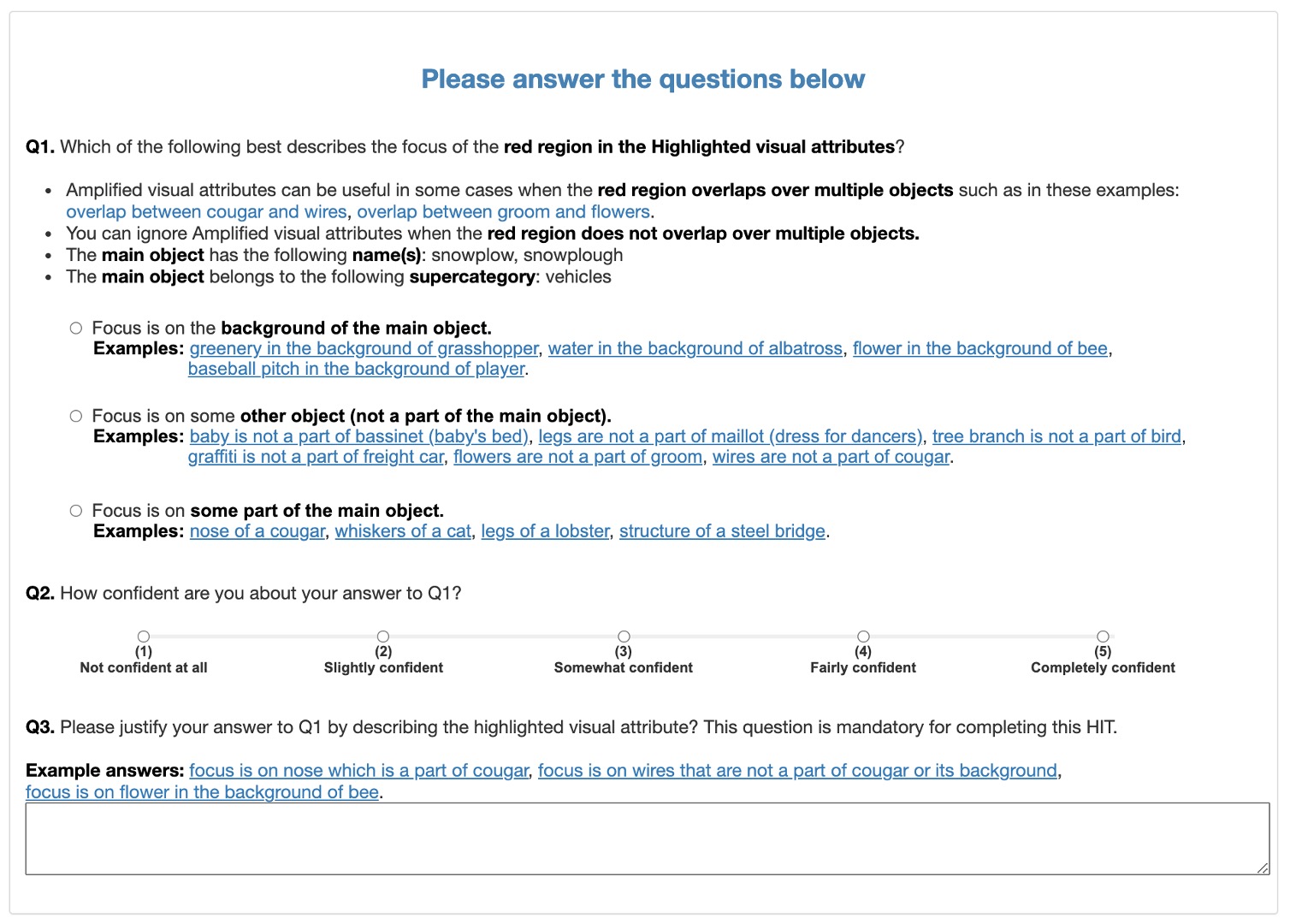}
\caption{Questionnaire}
\label{subfig:discover_spurious_questions}
\end{subfigure}
\caption{Mechanical Turk study for discovering spurious features. This figure is from \citet{salientimagenet2021} and included here for completeness.}
\label{fig:discover_spurious_study}
\end{figure}

The design for the Mechanical Turk study is shown in Figure \ref{fig:discover_spurious_study}. The left panel visualizing the neuron is shown in Figure \ref{subfig:discover_spurious_leftpanel}. The right panel describing the object class is shown in Figure \ref{subfig:discover_spurious_rightpanel}. The questionnaire is shown in Figure \ref{subfig:discover_spurious_questions}. We ask the workers to determine whether they think the visual attribute (given on the left) is a part of the main object (given on the right), some separate object or the background of the main object. We also ask the workers to provide reasons for their answers and rate their confidence on a likert scale from $1$ to $5$. The visualizations for which majority of workers selected either background or separate object as the answer were deemed to be spurious.  Workers were paid \$$0.1$ per HIT, with an average salary of $\$8$ per hour. In total, we had $137$ unique workers, each completing $140.15$ tasks (on average).

\section{Mechanical Turk study for validating heatmaps}\label{sec:mturk_validate_appendix}

\begin{figure}[h!]
\centering
\begin{subfigure}{0.9\linewidth}
\centering
\includegraphics[trim=0cm 0cm 0cm 0cm,  width=\linewidth]{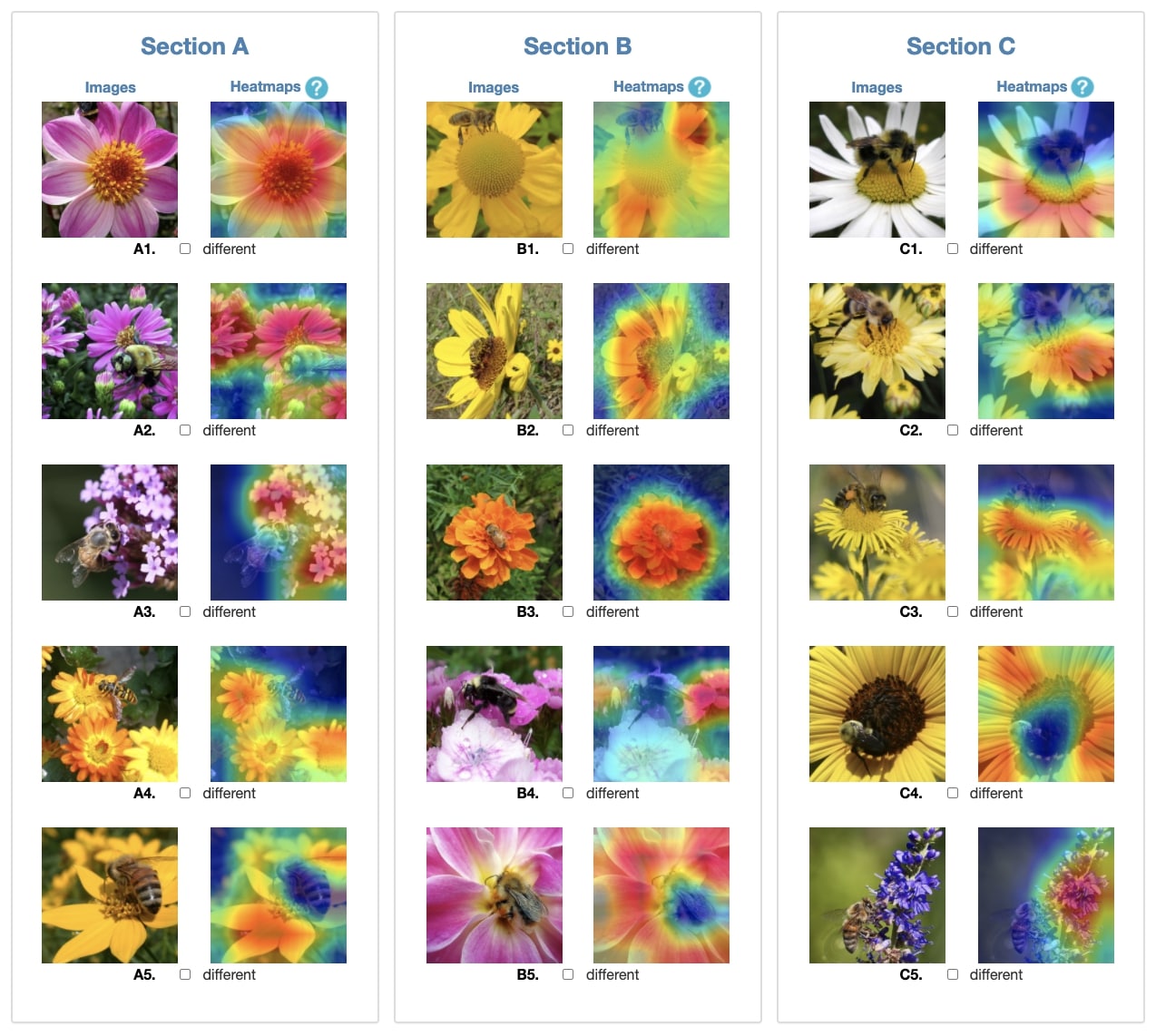}
\caption{Heatmaps highlighting the visual attributes}
\label{subfig:validate_all_panels}
\end{subfigure} \\~\\
\begin{subfigure}{0.9\linewidth}
\centering
\includegraphics[trim=0cm 0cm 0cm 0cm,  width=\linewidth]{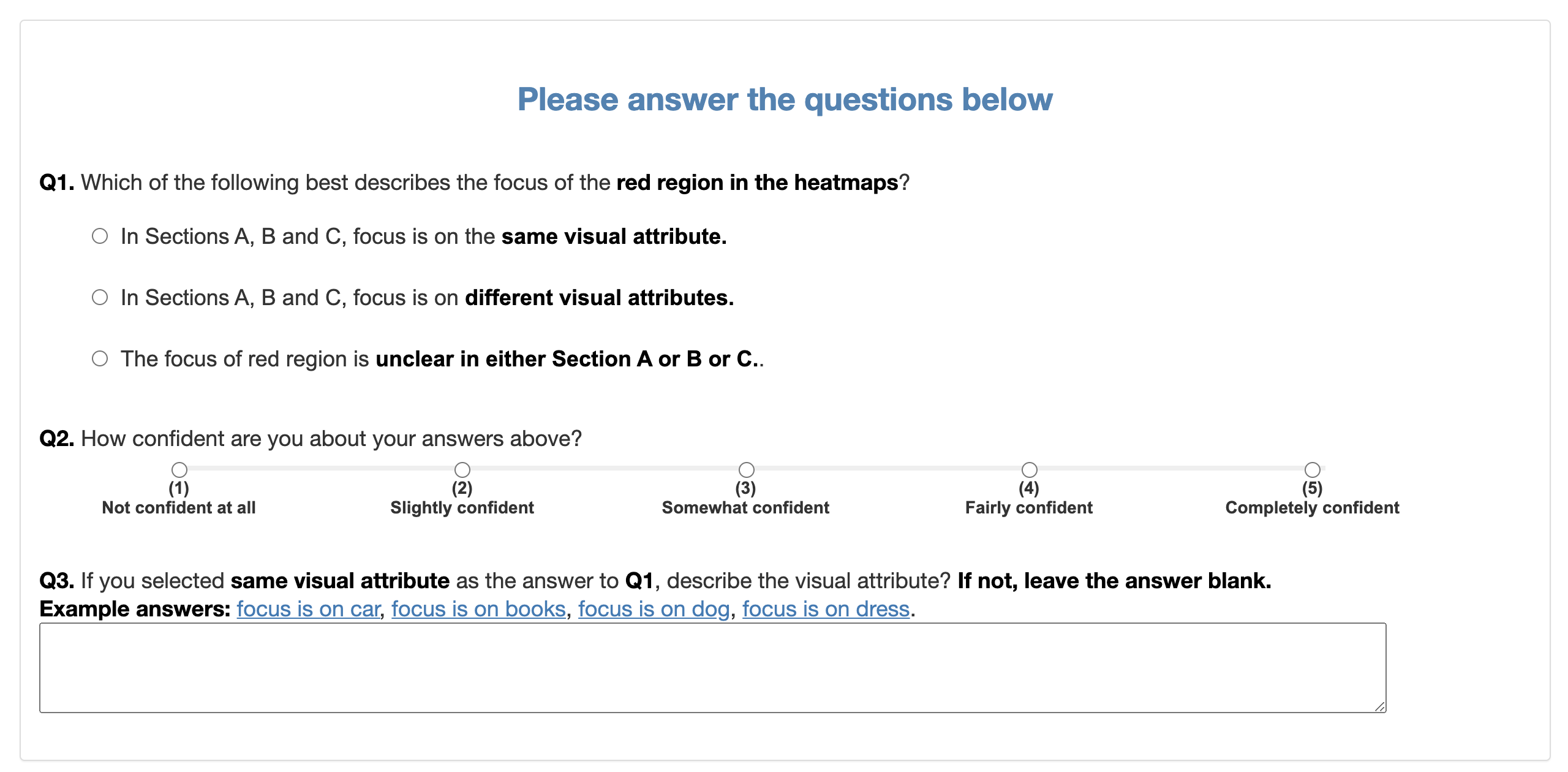}
\caption{Questionnaire}
\label{subfig:mturk_validate_questions}
\end{subfigure}
\caption{Mechanical Turk study for validating heatmaps}
\label{fig:mturk_validation_study}
\end{figure}

The design for the Mechanical Turk study is shown in Figure \ref{fig:mturk_validation_study}. The three panels showing heatmaps for different images from a class are shown in Figure \ref{subfig:validate_all_panels}. The questionnaire is shown in Figure \ref{subfig:mturk_validate_questions}. For each heatmap, workers were asked to determine if the highlighted attribute looked different from at least $3$ other heatmaps in the same panel. We also ask the workers to determine whether they think the focus of the heatmap is on the same object (in the three panels), different objects or whether they think the visualization in any of the panels is unclear. Same as in the previous study (Section \ref{sec:mturk_discovery_appendix}), we ask the workers to provide reasons for their answers and rate their confidence on a likert scale from $1$ to $5$. The visualizations for which at least $4$ workers selected same as the answer and for which at least $4$ workers did not select "different" as the answer for all $15$ heatmaps were deemed to be validated i.e for this subset of $260$ images, we assume that the \fm s focus on the same visual attribute.  

\clearpage

\section{Summary of the dataset validated using Mechanical Turk study in Section \ref{sec:mturk_studies}}\label{sec:appendix_validated_dataset}

Using the Mechanical Turk study in Section \ref{sec:mturk_studies}, for $4216$ (out of $4370$) core pairs i.e. (class=$\class$, feature=$\feature$)  pairs where $\feature$ is core for $\class$, we validate that NAMs for the top-$260$ images indeed focus on the desired visual attribute. This directly results in $4216 \times 260 = 1,096,160$ \textit{validated core masks}. 

Similarly, by taking the union of images in the sets $\dataset(\class, \feature)$, again where $\feature$ is core for $\class$ and (class=$\class$, feature=$\feature$) is among the $4216$ validated pairs, we obtain $565,950$ unique images.

% core, spurious = $629439$ \\
% core = $565950$ \\

% \input{background}
% \input{non_background}
% \input{motivation_err_analysis}
\section{Details on Relative Core Sensitivity}\label{sec:appendix_rcs}

In this work, we use a novel metric to facilitate comparisons of core and spurious accuracies across a diverse set of models. Specifically, {\it relative core sensitivity} ($\RCS$) is designed to address the potential lurking variable of {\it general noise robustness}. For example, a model that is generally very robust to noise will see small degradation due to noise anywhere. Thus, the absolute difference between core and spurious accuracy will be small, regardless of the relative model sensitivity to either region. To normalize against this limitation, we scale the absolute gap by the total possible gap (made precise below).

We note that the metric Relative Core Sensitivity ($\RCS$) adapted here is same as the metric called Relative Foreground Sensitivity ($\mathrm{RFS}$) in \citet{mazda}. We include two detailed derivations (one inspired from the original geometric derivation in \citet{mazda}, and a new algebraic derivation) here for brevity.
%, and a discussion in the novel contributions of our work compared to \citet{mazda}.

\subsection{Algebraic Derivation}

Consider a model with core and spurious accuracies $\causalaccabv, \spuriousaccabv$ respectively. We define $\bar{a} = \frac{(\causalaccabv +\spuriousaccabv)}{2}$, and use $\bar{a}$ as a proxy for general noise robustness. The gap between $\causalaccabv$ and $\spuriousaccabv$ reflects sensitivity to noise in the non-core regions relative to core regions. 

We seek to compute the maximum gap between $\causalaccabv$ and $\spuriousaccabv$ for a {\it fixed} general noise robustness of $\bar{a}$. The arguments maximizing the linear objective $\causalaccabv-\spuriousaccabv$ will occur at the boundary of the feasible region. There are two non-trivial cases to compare (the gap is obviously not maximized if $\causalaccabv\leq\spuriousaccabv$ so we ignore these cases):
\begin{itemize}
    \item The maximum gap occurs when $\causalaccabv=1$. Thus, $\spuriousaccabv=2\bar{a}-1$, yielding a gap of $\causalaccabv-\spuriousaccabv= 2(1-\bar{a})$. 
    \item The maximum gap occurs when $\spuriousaccabv=0$. Thus, $\causalaccabv=2\bar{a}$, yielding a gap of $\causalaccabv-\spuriousaccabv=2\bar{a}$.
\end{itemize}

However, notice that the feasability of the above cases is contingent on $\bar{a}$. Specifically, $\causalaccabv$ can only be $1$ if $\bar{a}\geq 0.5$, and $\spuriousaccabv$ can only be $0$ if $\bar{a} \leq 0.5$. Thus, as a piecewise function with respect to $\bar{a}$, the maximum gap is $2\bar{a}$ for $\bar{a} \leq 0.5$ and $2(1-\bar{a})$ for $\bar{a}\geq 0.5$. Now, observe that this piecewise definition can be consolidated as $2\min(\bar{a}, 1-\bar{a})$.

Hence, defining $\RCS$ to be the ratio of the absolute gap between core and spurious accuracy, and the total possible gap for any model with general noise robustness of $\bar{a}$, yields the original formula $\RCS = \frac{\causalaccabv-\spuriousaccabv}{2\min(\bar{a}, 1-\bar{a})}$.

\subsection{Geometric Derivation}

$\RCS$ can also be viewed geometrically as the ratio of the distance of point $(\spuriousaccabv, \causalaccabv)$ above the diagonal over the maximum distance from the diagonal for models with fixed general noise robustness $\bar{a}$. 

First, observe the distance of $(\spuriousaccabv, \causalaccabv)$ from the diagonal is given by the distance between the point and $(\bar{a}, \bar{a})$, yielding:

$$\text{Distance to Diagonal} = \sqrt{2}(\bar{a} - \spuriousaccabv) = \frac{\sqrt{2}(\causalaccabv-\spuriousaccabv)}{2}$$

assuming that $\causalaccabv > \spuriousaccabv$ (though otherwise the sign would simply be flipped). 

The maximum distance from the diagonal is constrained by the fact that $0 \leq \causalaccabv, \spuriousaccabv \leq 1$. Because $\bar{a}$ is fixed, we necessarily lie on the line $\causalaccabv = 2\bar{a} - \spuriousaccabv$. Notice that when $\bar{a} \leq 0.5$, we intersect the boundary on the $y$-axis, at point $(0,2\bar{a})$. When $\bar{a} \geq 0.5$, we intersect the boundary defined by $y=1$, at the point $(1-2\bar{a}, 1)$. The corresponding maximum distances are then:
$$\text{Max Distance to Diagonal} = \sqrt{2}\bar{a} \;\text{ for }\; \bar{a} \leq 0.5, \text{ and } \sqrt{2}(1-\bar{a}) \;\text{ for }\; \bar{a} \geq 0.5$$

As in the algebraic derivation, the piecewise formula can be resolved as $\sqrt{2}\min(\bar{a}, 1-\bar{a})$. Therefore, in the final ratio of distance of $(\spuriousaccabv, \causalaccabv)$ to diagonal over maximum distance to diagonal for fixed $\bar{a}$, the $\sqrt{2}$ terms cancel, yielding the formula for $\RCS$. For a pictographic geometric derivation, we refer readers to \citet{mazda}.

\section{Novel Contributions compared to Previous Work} \label{sec:appendix_comparing_to_rival10}

Certain aspects of our analysis are similar to previous work. Namely, the $\RCS$ metric is adapted from the {\it relative foreground sensitivity} metric of \citet{mazda}, and both \citet{mazda} and \citet{salientimagenet2021} conduct noise-based analyses to discern model sensitivity to image regions. Further, some of our observations on pretrained models were also noted in \citet{mazda} (i.e. lower sensitivity to core/foreground regions in transformers and adversarial trained Resnets, relative to Resnets). We acknowledge the inspiration taken from these efforts, and highlight two key distinguishing aspects of our work that we believe significantly add to the prior findings. 

The first is scale, in both data and models: our evaluation includes $226k$ images from $985$ classes, compared to $5k$ from $20$ classes (organized into ten subsets called RIVAL$10$) in \citet{mazda}, and $52k$ from $232$ classes in \citet{salientimagenet2021}; we evaluate $42$ models, compared to $17$ in \citet{mazda} and $4$ in \citet{salientimagenet2021}. 

The second is that evaluating $\RCS$ on the test set of Salient Imagenet-1M can be done without making {\it any} changes to pretrained models, where as \citet{mazda} require models to be finetuned on the ten class subset of images they consider. While finetuning is a standard procedure, it changes the weights of neural features used to perform classification, which may introduce biases. Moreover, certain core features may be discarded when the classification task is simplified to the much coarser labels of RIVAL$10$. The evaluation in \citet{salientimagenet2021} does not attempt to control for varying noise robustness, as models are not compared to one another directly.

Thus, we believe the findings of our experiments, due to the scale and lack of modifying pretrained models, may be empirically stronger than those of \citet{mazda}. Nonetheless, we find it encouraging that we corroborate the findings of \citet{mazda} on a separate, larger dataset. 

\section{Using data augmentation with Salient Imagenet-1M}\label{sec:appendix_corm_training}

\subsection{Training Details for CoRM Models}
We follow the fast training procedure for the baseline in \citet{wong_fast}. Cyclic learning rates are applied from $0.1$ to $0.004$ for an SGD optimizer over 15 epochs. The only augmentation is to resize and center crop images to $224\times 224$. We discuss this choice below. 

\subsection{On Augmentation for Salient Imagenet-1M}

Random cropping, a common augmentation technique, was not directly possible for the Salient ImageNet-1M version used in this work, as the masks were obtained for images after undergoing the standard ImageNet test transformations of resizing and taking a square center crop. However, there are multiple approaches for incorporating augmentation going forward. 

%First, one can apply any augmentation, so long as it follows the aforementioned standard test transformation. The limitation is that image content outside the original square crop is not available.

First, one can generate the core masks by generating NAMs on the fly using the same robust model (used in this work for generating the NAMs for Salient Imagenet-1M) during training. That is, after performing any augmentation on the {\it original} image, one can compute NAMs for the relevant features on the augmented image, and use these directly as done in this work.

Second, one can precompute NAMs for all original images. This would require computing NAMs using training images that have been resized such that the shorter side is $224$. Then during training, the same random cropping transformation can be applied to the image and the mask to obtain the masks for the relevant core/non-core regions.

%it can be passed through the Robust Resnet-50 used in NAM generation. Afterwards, the NAMs will undergo the reverse rescaling transformation so that they correspond to the original image dimensions. 

Thus, while augmentation was not used in this work, it is certainly feasible for Salient Imagenet-1M and will be explored in future works.

\section{Examples of spurious features}\label{sec:spurious_appendix}
For each (class=$\class$, feature=$\feature$) pair where $\feature$ is known to be spurious for the class $\class$, we analyze the sensitivity of various standard (non-robust) trained models: Resnet-50, Efficientnet-B7, CLIP VIT-B32, VIT-B32 to different spurious features. 

We first compute the clean accuracy for the set $\dataset(\class, \feature)$ for each model (called \textit{initial} in the the figure captions below). Next, for each image and spurious mask in $\dataset(\class, \feature)$, i.e. $\bx,\bs \in \dataset(\class, \feature)$, we compute $\bx^{*} = \bx + \sigma \left(\bs \odot \bz \right)$ where $\bz \sim \mathcal{N}\left(\zeros, \bI\right)$. Next, we compute the accuracy for each model using these noisy images and the drop in model accuracy (called \textit{accuracy drop} in the figure captions below). We use $\sigma = 0.25$.

%we compute the accuracy by adding Gaussian noise (using $\sigma=0.25$) to the spurious mask $\bs$ for each image $\bx$ in the set $\dataset(\class, \feature)$. i.e. $\bx \in \dataset(\class, \feature)$ in this set. 

\subsection{Background spurious features}\label{sec:background_spurious_appendix}

\begin{figure}[h!]
\centering
\begin{subfigure}{\linewidth}
\centering
\includegraphics[trim=0cm 0cm 0cm 0cm, clip, width=0.9\linewidth]{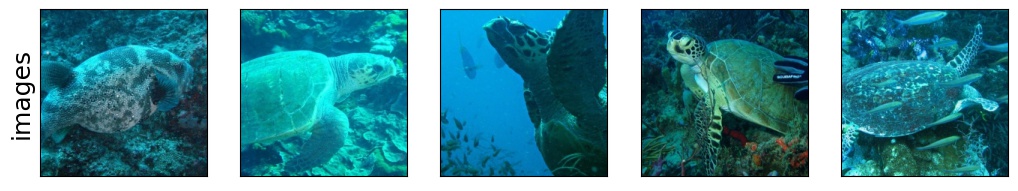}
\end{subfigure}\
\begin{subfigure}{\linewidth}
\centering
\includegraphics[trim=0cm 0cm 0cm 0cm, clip, width=0.9\linewidth]{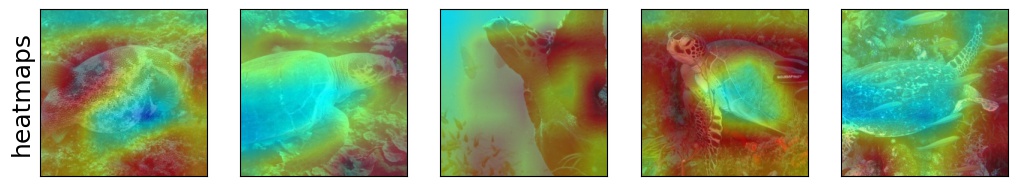}
\end{subfigure}\
\begin{subfigure}{\linewidth}
\centering
\includegraphics[trim=0cm 0cm 0cm 0cm, clip, width=0.9\linewidth]{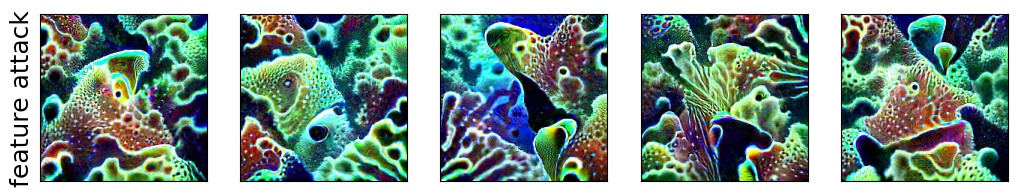}
\end{subfigure}
\caption{Visualization of feature \textbf{981} for class \textbf{loggerhead} (class index: \textbf{33}).\\ For \textbf{Resnet-50}, accuracy drop: \textcolor{red}{\textbf{-73.846\%}} (initial: \textbf{96.923\%}). For \textbf{Efficientnet-B7}, accuracy drop: \textcolor{red}{\textbf{-35.385}}\% (initial: \textbf{89.231\%}).\\ For \textbf{CLIP VIT-B32}, accuracy drop: \textcolor{red}{\textbf{-53.846\%}} (initial: \textbf{84.615\%}). For \textbf{VIT-B32}, accuracy drop: \textcolor{red}{\textbf{-30.769\%}} (initial: \textbf{76.923\%}).}
\label{fig:appendix_33_981}
\end{figure}

% \clearpage

\begin{figure}[h!]
\centering
\begin{subfigure}{\linewidth}
\centering
\includegraphics[trim=0cm 0cm 0cm 0cm, clip, width=0.9\linewidth]{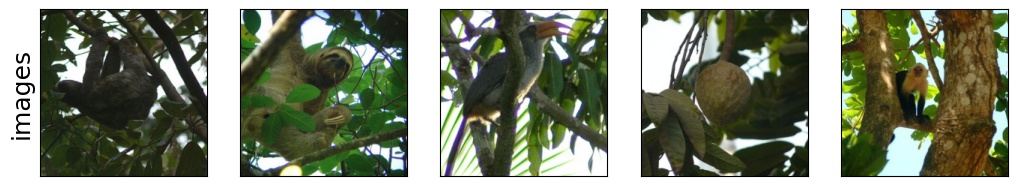}
\end{subfigure}\
\begin{subfigure}{\linewidth}
\centering
\includegraphics[trim=0cm 0cm 0cm 0cm, clip, width=0.9\linewidth]{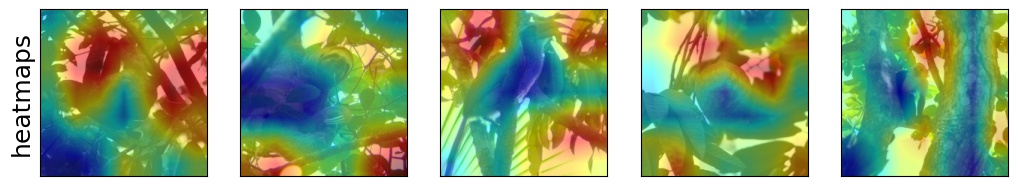}
\end{subfigure}\
\begin{subfigure}{\linewidth}
\centering
\includegraphics[trim=0cm 0cm 0cm 0cm, clip, width=0.9\linewidth]{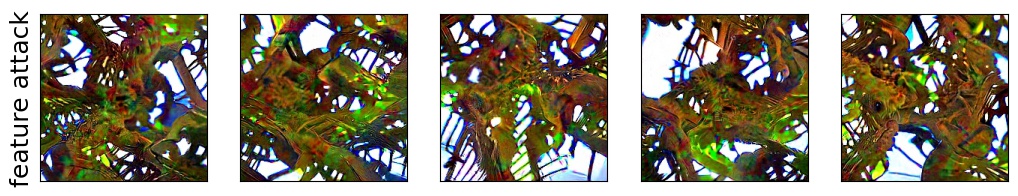}
\end{subfigure}
\caption{Visualization of feature \textbf{1199} for class \textbf{three toed sloth} (class index: \textbf{364}).\\ For \textbf{Resnet-50}, accuracy drop: \textcolor{red}{\textbf{-64.615\%}} (initial: \textbf{96.923\%}). For \textbf{Efficientnet-B7}, accuracy drop: \textcolor{red}{\textbf{-24.615}}\% (initial: \textbf{93.846\%}).\\ For \textbf{CLIP VIT-B32}, accuracy drop: \textcolor{red}{\textbf{-67.692\%}} (initial: \textbf{67.692\%}). For \textbf{VIT-B32}, accuracy drop: \textcolor{red}{\textbf{-4.615\%}} (initial: \textbf{87.692\%}).}
\label{fig:appendix_364_1199}
\end{figure}

% \clearpage

% \begin{figure}[h!]
% \centering
% \begin{subfigure}{\linewidth}
% \centering
% \includegraphics[trim=0cm 0cm 0cm 0cm, clip, width=0.9\linewidth]{visualizations/379_912_images}
% \end{subfigure}\
% \begin{subfigure}{\linewidth}
% \centering
% \includegraphics[trim=0cm 0cm 0cm 0cm, clip, width=0.9\linewidth]{visualizations/379_912_heatmaps}
% \end{subfigure}\
% \begin{subfigure}{\linewidth}
% \centering
% \includegraphics[trim=0cm 0cm 0cm 0cm, clip, width=0.9\linewidth]{visualizations/379_912_attacks}
% \end{subfigure}
% \caption{Visualization of feature \textbf{912} for class \textbf{howler monkey} (class index: \textbf{379}).\\ For \textbf{Resnet-50}, accuracy drop: \textcolor{red}{\textbf{-23.077\%}} (initial: \textbf{92.308\%}). For \textbf{Efficientnet-B7}, accuracy drop: \textcolor{red}{\textbf{-32.308}}\% (initial: \textbf{93.846\%}).\\ For \textbf{CLIP VIT-B32}, accuracy drop: \textcolor{red}{\textbf{-46.154\%}} (initial: \textbf{46.154\%}). For \textbf{VIT-B32}, accuracy drop: \textcolor{blue}{\textbf{+1.539\%}} (initial: \textbf{76.923\%}).}
% \label{fig:appendix_379_912}
% \end{figure}

% \clearpage

\begin{figure}[h!]
\centering
\begin{subfigure}{\linewidth}
\centering
\includegraphics[trim=0cm 0cm 0cm 0cm, clip, width=0.9\linewidth]{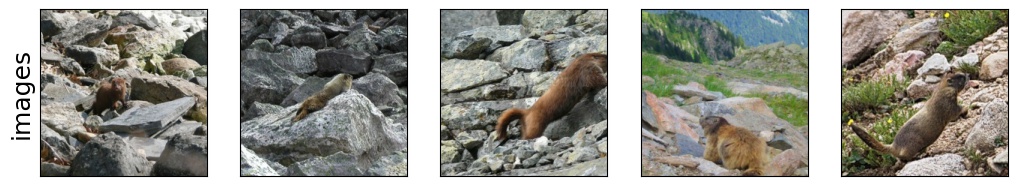}
\end{subfigure}\
\begin{subfigure}{\linewidth}
\centering
\includegraphics[trim=0cm 0cm 0cm 0cm, clip, width=0.9\linewidth]{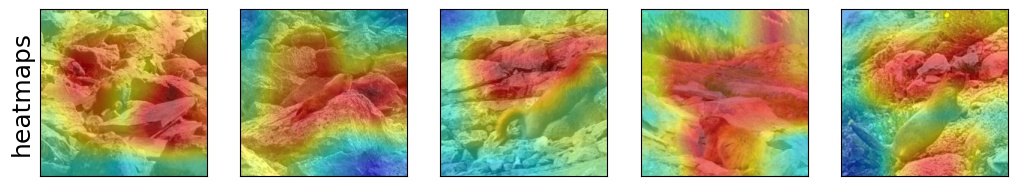}
\end{subfigure}\
\begin{subfigure}{\linewidth}
\centering
\includegraphics[trim=0cm 0cm 0cm 0cm, clip, width=0.9\linewidth]{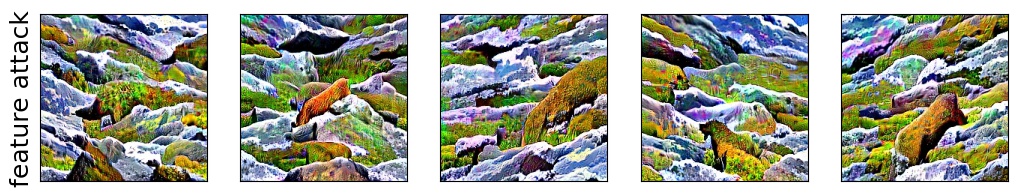}
\end{subfigure}
\caption{Visualization of feature \textbf{820} for class \textbf{marmot} (class index: \textbf{336}).\\ For \textbf{Resnet-50}, accuracy drop: \textcolor{red}{\textbf{-64.616\%}} (initial: \textbf{95.385\%}). For \textbf{Efficientnet-B7}, accuracy drop: \textcolor{red}{\textbf{-24.616}}\% (initial: \textbf{86.154\%}).\\ For \textbf{CLIP VIT-B32}, accuracy drop: \textcolor{red}{\textbf{-47.693\%}} (initial: \textbf{89.231\%}). For \textbf{VIT-B32}, accuracy drop: \textcolor{red}{\textbf{-40.0\%}} (initial: \textbf{86.154\%}).}
\label{fig:appendix_336_820}
\end{figure}

\clearpage

\begin{figure}[h!]
\centering
\begin{subfigure}{\linewidth}
\centering
\includegraphics[trim=0cm 0cm 0cm 0cm, clip, width=0.9\linewidth]{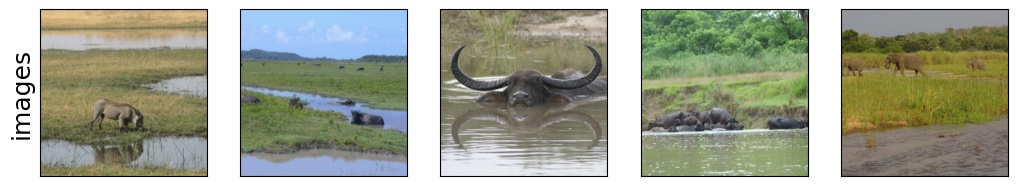}
\end{subfigure}\
\begin{subfigure}{\linewidth}
\centering
\includegraphics[trim=0cm 0cm 0cm 0cm, clip, width=0.9\linewidth]{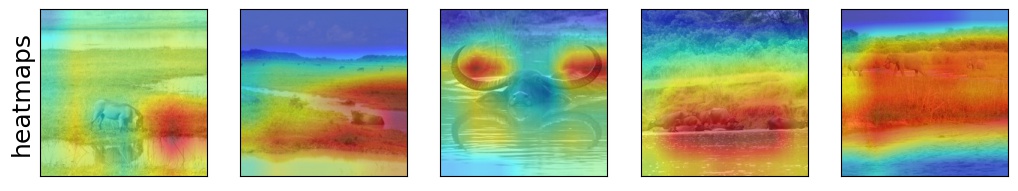}
\end{subfigure}\
\begin{subfigure}{\linewidth}
\centering
\includegraphics[trim=0cm 0cm 0cm 0cm, clip, width=0.9\linewidth]{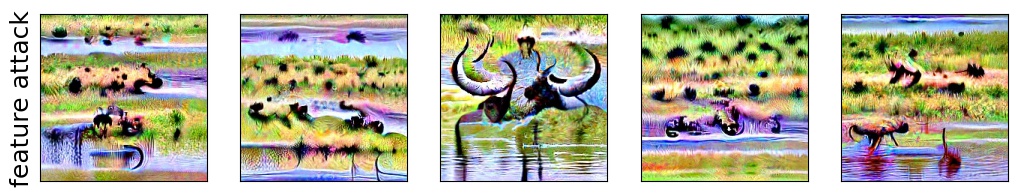}
\end{subfigure}
\caption{Visualization of feature \textbf{1127} for class \textbf{water buffalo} (class index: \textbf{346}).\\ For \textbf{Resnet-50}, accuracy drop: \textcolor{red}{\textbf{-58.461\%}} (initial: \textbf{84.615\%}). For \textbf{Efficientnet-B7}, accuracy drop: \textcolor{red}{\textbf{-44.615}}\% (initial: \textbf{87.692\%}).\\ For \textbf{CLIP VIT-B32}, accuracy drop: \textcolor{red}{\textbf{-43.076\%}} (initial: \textbf{81.538\%}). For \textbf{VIT-B32}, accuracy drop: \textcolor{red}{\textbf{-41.538\%}} (initial: \textbf{90.769\%}).}
\label{fig:appendix_346_1127}
\end{figure}

\begin{figure}[h!]
\centering
\begin{subfigure}{\linewidth}
\centering
\includegraphics[trim=0cm 0cm 0cm 0cm, clip, width=0.9\linewidth]{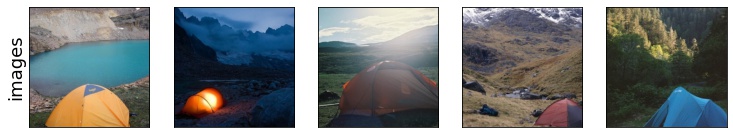}
\end{subfigure}\
\begin{subfigure}{\linewidth}
\centering
\includegraphics[trim=0cm 0cm 0cm 0cm, clip, width=0.9\linewidth]{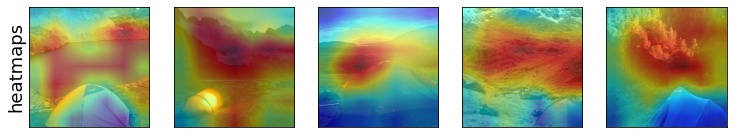}
\end{subfigure}\
\begin{subfigure}{\linewidth}
\centering
\includegraphics[trim=0cm 0cm 0cm 0cm, clip, width=0.9\linewidth]{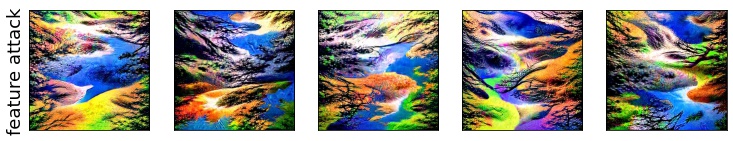}
\end{subfigure}
\caption{Visualization of feature \textbf{1239} for class \textbf{mountain tent} (class index: \textbf{672}).\\ For \textbf{Resnet-50}, accuracy drop: \textcolor{red}{\textbf{-47.692\%}} (initial: \textbf{96.923\%}). For \textbf{Efficientnet-B7}, accuracy drop: \textcolor{red}{\textbf{-10.77}}\% (initial: \textbf{95.385\%}).\\ For \textbf{CLIP VIT-B32}, accuracy drop: \textcolor{red}{\textbf{-75.384\%}} (initial: \textbf{87.692\%}). For \textbf{VIT-B32}, accuracy drop: \textcolor{red}{\textbf{-20.0\%}} (initial: \textbf{96.923\%}).}
\label{fig:appendix_672_1239}
\end{figure}

\clearpage

\begin{figure}[h!]
\centering
\begin{subfigure}{\linewidth}
\centering
\includegraphics[trim=0cm 0cm 0cm 0cm, clip, width=0.9\linewidth]{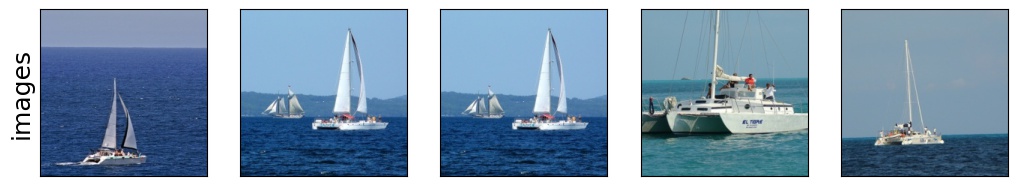}
\end{subfigure}\
\begin{subfigure}{\linewidth}
\centering
\includegraphics[trim=0cm 0cm 0cm 0cm, clip, width=0.9\linewidth]{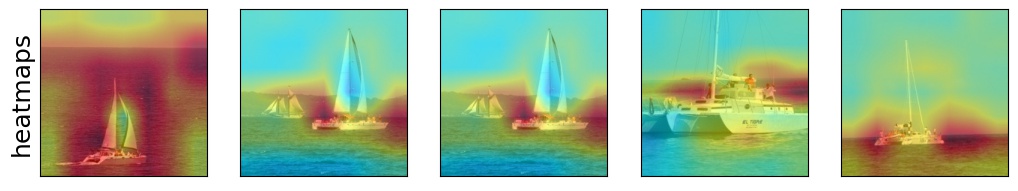}
\end{subfigure}\
\begin{subfigure}{\linewidth}
\centering
\includegraphics[trim=0cm 0cm 0cm 0cm, clip, width=0.9\linewidth]{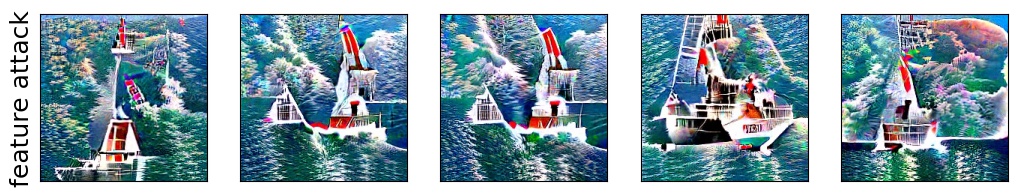}
\end{subfigure}
\caption{Visualization of feature \textbf{961} for class \textbf{catamaran} (class index: \textbf{484}).\\ For \textbf{Resnet-50}, accuracy drop: \textcolor{red}{\textbf{-50.769\%}} (initial: \textbf{93.846\%}). For \textbf{Efficientnet-B7}, accuracy drop: \textcolor{red}{\textbf{-20.0}}\% (initial: \textbf{95.385\%}).\\ For \textbf{CLIP VIT-B32}, accuracy drop: \textcolor{red}{\textbf{-46.154\%}} (initial: \textbf{89.231\%}). For \textbf{VIT-B32}, accuracy drop: \textcolor{red}{\textbf{-9.231\%}} (initial: \textbf{93.846\%}).}
\label{fig:appendix_484_961}
\end{figure}

% \clearpage

\begin{figure}[h!]
\centering
\begin{subfigure}{\linewidth}
\centering
\includegraphics[trim=0cm 0cm 0cm 0cm, clip, width=0.9\linewidth]{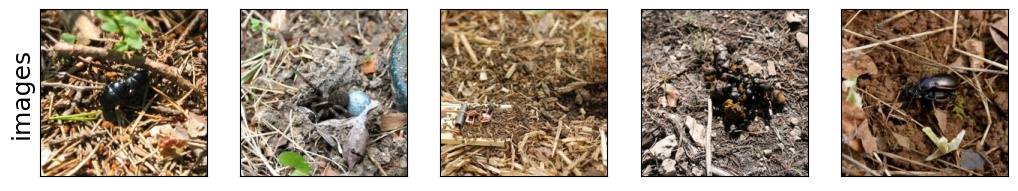}
\end{subfigure}\
\begin{subfigure}{\linewidth}
\centering
\includegraphics[trim=0cm 0cm 0cm 0cm, clip, width=0.9\linewidth]{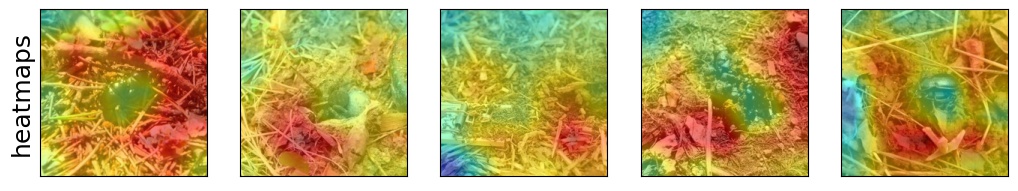}
\end{subfigure}\
\begin{subfigure}{\linewidth}
\centering
\includegraphics[trim=0cm 0cm 0cm 0cm, clip, width=0.9\linewidth]{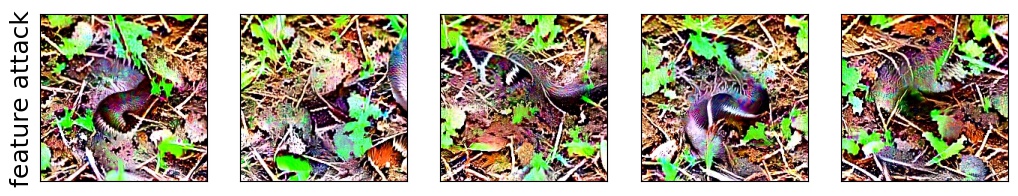}
\end{subfigure}
\caption{Visualization of feature \textbf{880} for class \textbf{dung beetle} (class index: \textbf{305}).\\ For \textbf{Resnet-50}, accuracy drop: \textcolor{red}{\textbf{-50.77\%}} (initial: \textbf{98.462\%}). For \textbf{Efficientnet-B7}, accuracy drop: \textcolor{red}{\textbf{-6.154}}\% (initial: \textbf{98.462\%}).\\ For \textbf{CLIP VIT-B32}, accuracy drop: \textcolor{red}{\textbf{-52.308\%}} (initial: \textbf{80.0\%}). For \textbf{VIT-B32}, accuracy drop: \textcolor{red}{\textbf{-13.846\%}} (initial: \textbf{90.769\%}).}
\label{fig:appendix_305_880}
\end{figure}

\clearpage

\begin{figure}[h!]
\centering
\begin{subfigure}{\linewidth}
\centering
\includegraphics[trim=0cm 0cm 0cm 0cm, clip, width=0.9\linewidth]{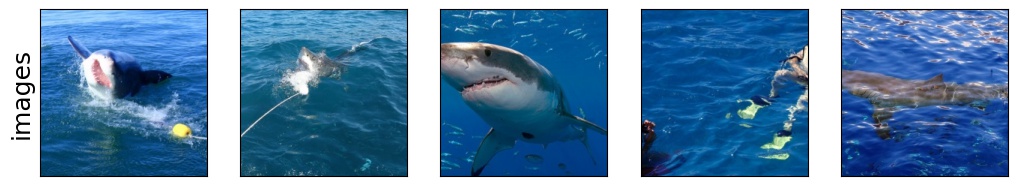}
\end{subfigure}\
\begin{subfigure}{\linewidth}
\centering
\includegraphics[trim=0cm 0cm 0cm 0cm, clip, width=0.9\linewidth]{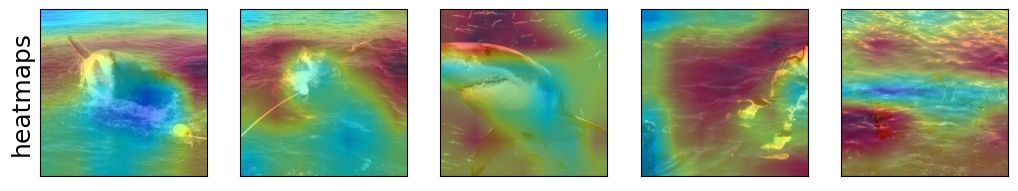}
\end{subfigure}\
\begin{subfigure}{\linewidth}
\centering
\includegraphics[trim=0cm 0cm 0cm 0cm, clip, width=0.9\linewidth]{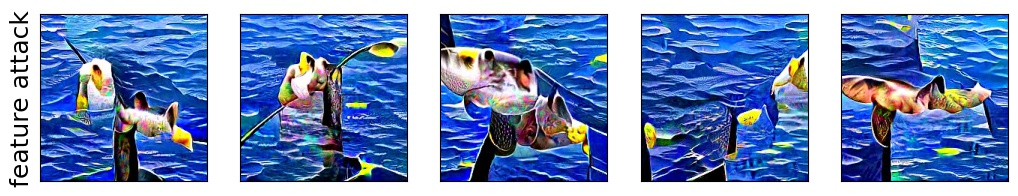}
\end{subfigure}
\caption{Visualization of feature \textbf{1697} for class \textbf{great white shark} (class index: \textbf{2}).\\ For \textbf{Resnet-50}, accuracy drop: \textcolor{red}{\textbf{-44.615\%}} (initial: \textbf{90.769\%}). For \textbf{Efficientnet-B7}, accuracy drop: \textcolor{red}{\textbf{-38.461}}\% (initial: \textbf{87.692\%}).\\ For \textbf{CLIP VIT-B32}, accuracy drop: \textcolor{red}{\textbf{-26.154\%}} (initial: \textbf{49.231\%}). For \textbf{VIT-B32}, accuracy drop: \textcolor{red}{\textbf{-23.077\%}} (initial: \textbf{84.615\%}).}
\label{fig:appendix_2_1697}
\end{figure}

% \clearpage

\begin{figure}[h!]
\centering
\begin{subfigure}{\linewidth}
\centering
\includegraphics[trim=0cm 0cm 0cm 0cm, clip, width=0.9\linewidth]{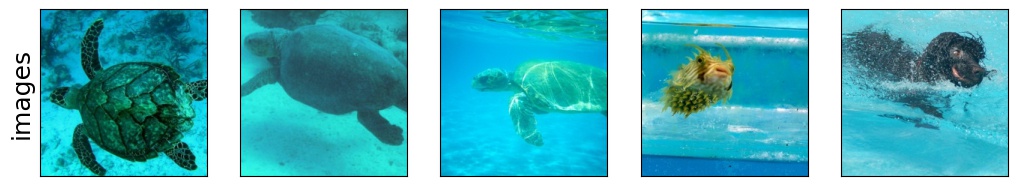}
\end{subfigure}\
\begin{subfigure}{\linewidth}
\centering
\includegraphics[trim=0cm 0cm 0cm 0cm, clip, width=0.9\linewidth]{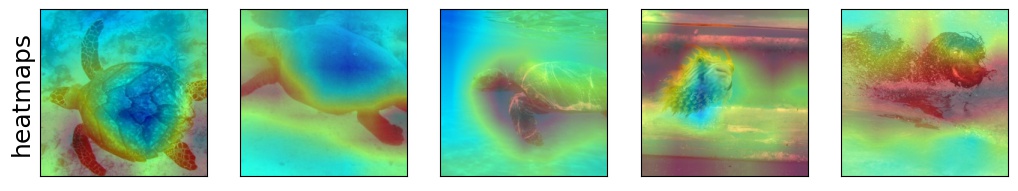}
\end{subfigure}\
\begin{subfigure}{\linewidth}
\centering
\includegraphics[trim=0cm 0cm 0cm 0cm, clip, width=0.9\linewidth]{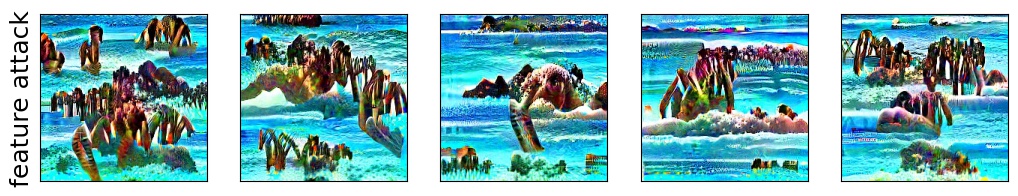}
\end{subfigure}
\caption{Visualization of feature \textbf{491} for class \textbf{loggerhead} (class index: \textbf{33}).\\ For \textbf{Resnet-50}, accuracy drop: \textcolor{red}{\textbf{-78.462\%}} (initial: \textbf{95.385\%}). For \textbf{Efficientnet-B7}, accuracy drop: \textcolor{red}{\textbf{-29.231}}\% (initial: \textbf{89.231\%}).\\ For \textbf{CLIP VIT-B32}, accuracy drop: \textcolor{red}{\textbf{-35.384\%}} (initial: \textbf{87.692\%}). For \textbf{VIT-B32}, accuracy drop: \textcolor{red}{\textbf{-21.539\%}} (initial: \textbf{83.077\%}).}
\label{fig:appendix_33_491}
\end{figure}

\clearpage

\begin{figure}[h!]
\centering
\begin{subfigure}{\linewidth}
\centering
\includegraphics[trim=0cm 0cm 0cm 0cm, clip, width=0.9\linewidth]{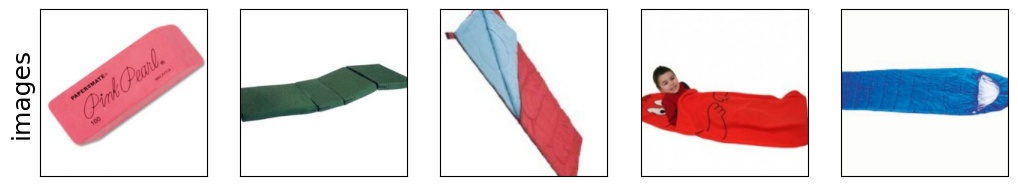}
\end{subfigure}\
\begin{subfigure}{\linewidth}
\centering
\includegraphics[trim=0cm 0cm 0cm 0cm, clip, width=0.9\linewidth]{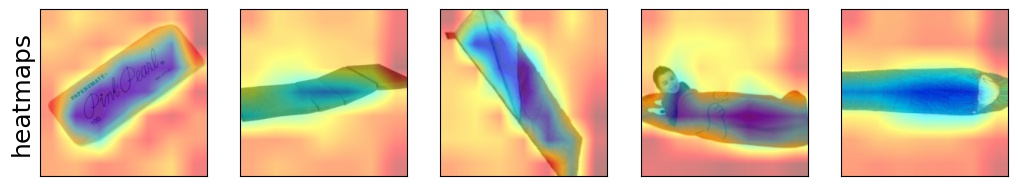}
\end{subfigure}\
\begin{subfigure}{\linewidth}
\centering
\includegraphics[trim=0cm 0cm 0cm 0cm, clip, width=0.9\linewidth]{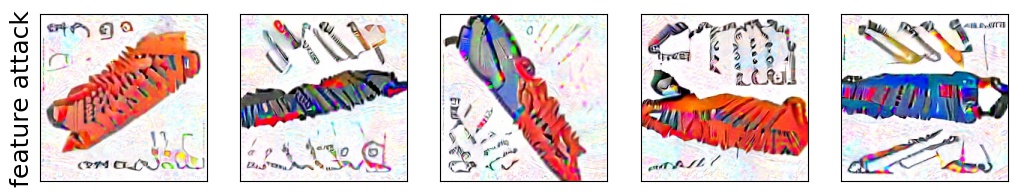}
\end{subfigure}
\caption{Visualization of feature \textbf{118} for class \textbf{sleeping bag} (class index: \textbf{797}).\\ For \textbf{Resnet-50}, accuracy drop: \textcolor{red}{\textbf{-41.539\%}} (initial: \textbf{98.462\%}). For \textbf{Efficientnet-B7}, accuracy drop: \textcolor{red}{\textbf{-20.0}}\% (initial: \textbf{95.385\%}).\\ For \textbf{CLIP VIT-B32}, accuracy drop: \textcolor{red}{\textbf{-63.077\%}} (initial: \textbf{78.462\%}). For \textbf{VIT-B32}, accuracy drop: \textcolor{red}{\textbf{-10.769\%}} (initial: \textbf{84.615\%}).}
\label{fig:appendix_797_118}
\end{figure}

\begin{figure}[h!]
\centering
\begin{subfigure}{\linewidth}
\centering
\includegraphics[trim=0cm 0cm 0cm 0cm, clip, width=0.9\linewidth]{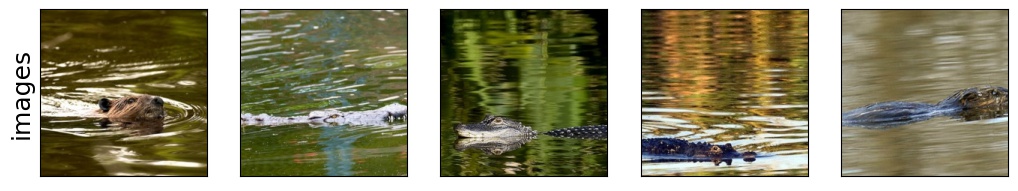}
\end{subfigure}\
\begin{subfigure}{\linewidth}
\centering
\includegraphics[trim=0cm 0cm 0cm 0cm, clip, width=0.9\linewidth]{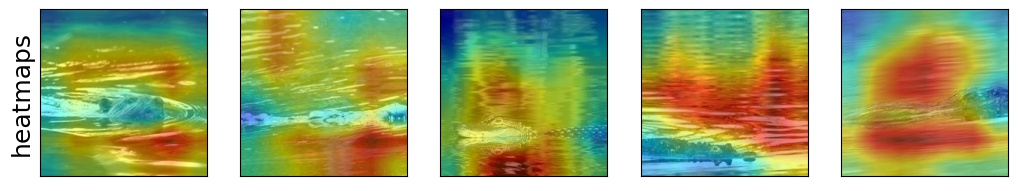}
\end{subfigure}\
\begin{subfigure}{\linewidth}
\centering
\includegraphics[trim=0cm 0cm 0cm 0cm, clip, width=0.9\linewidth]{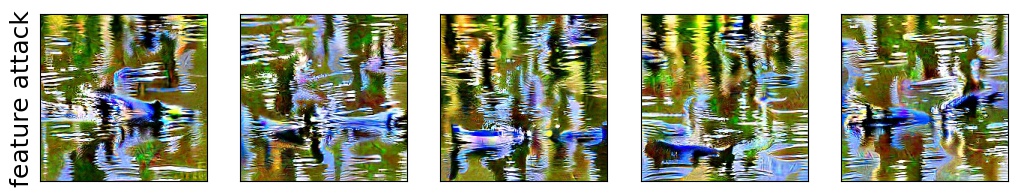}
\end{subfigure}
\caption{Visualization of feature \textbf{341} for class \textbf{beaver} (class index: \textbf{337}).\\ For \textbf{Resnet-50}, accuracy drop: \textcolor{red}{\textbf{-43.077\%}} (initial: \textbf{96.923\%}). For \textbf{Efficientnet-B7}, accuracy drop: \textcolor{red}{\textbf{-21.538}}\% (initial: \textbf{96.923\%}).\\ For \textbf{CLIP VIT-B32}, accuracy drop: \textcolor{red}{\textbf{-35.385\%}} (initial: \textbf{58.462\%}). For \textbf{VIT-B32}, accuracy drop: \textcolor{red}{\textbf{-12.308\%}} (initial: \textbf{89.231\%}).}
\label{fig:appendix_337_341}
\end{figure}

% \begin{figure}[h!]
% \centering
% \begin{subfigure}{\linewidth}
% \centering
% \includegraphics[trim=0cm 0cm 0cm 0cm, clip, width=0.9\linewidth]{visualizations/379_912_images}
% \end{subfigure}\
% \begin{subfigure}{\linewidth}
% \centering
% \includegraphics[trim=0cm 0cm 0cm 0cm, clip, width=0.9\linewidth]{visualizations/379_912_heatmaps}
% \end{subfigure}\
% \begin{subfigure}{\linewidth}
% \centering
% \includegraphics[trim=0cm 0cm 0cm 0cm, clip, width=0.9\linewidth]{visualizations/379_912_attacks}
% \end{subfigure}
% \caption{Visualization of feature \textbf{912} for class \textbf{howler monkey} (class index: \textbf{379}).\\ For \textbf{Resnet-50}, accuracy drop: \textcolor{red}{\textbf{-23.077\%}} (initial: \textbf{92.308\%}). For \textbf{Efficientnet-B7}, accuracy drop: \textcolor{red}{\textbf{-32.308}}\% (initial: \textbf{93.846\%}).\\ For \textbf{CLIP VIT-B32}, accuracy drop: \textcolor{red}{\textbf{-46.154\%}} (initial: \textbf{46.154\%}). For \textbf{VIT-B32}, accuracy drop: \textcolor{blue}{\textbf{+1.539\%}} (initial: \textbf{76.923\%}).}
% \label{fig:appendix_379_912}
% \end{figure}

% \clearpage
\clearpage

\begin{figure}[h!]
\centering
\begin{subfigure}{\linewidth}
\centering
\includegraphics[trim=0cm 0cm 0cm 0cm, clip, width=0.9\linewidth]{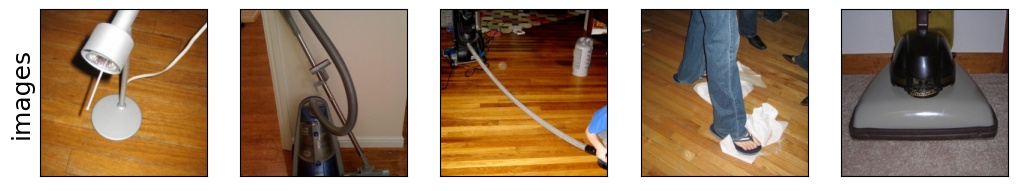}
\end{subfigure}\
\begin{subfigure}{\linewidth}
\centering
\includegraphics[trim=0cm 0cm 0cm 0cm, clip, width=0.9\linewidth]{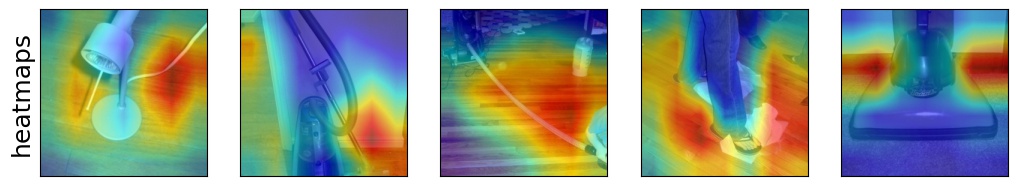}
\end{subfigure}\
\begin{subfigure}{\linewidth}
\centering
\includegraphics[trim=0cm 0cm 0cm 0cm, clip, width=0.9\linewidth]{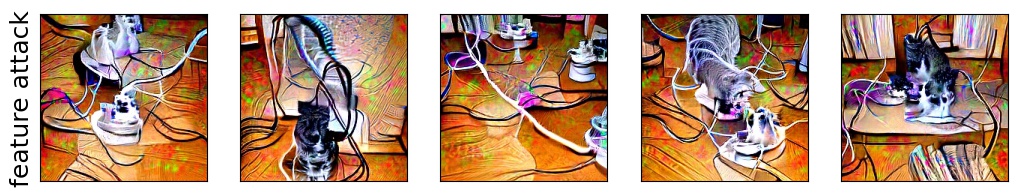}
\end{subfigure}
\caption{Visualization of feature \textbf{1291} for class \textbf{vacuum} (class index: \textbf{882}).\\ For \textbf{Resnet-50}, accuracy drop: \textcolor{red}{\textbf{-29.231\%}} (initial: \textbf{93.846\%}). For \textbf{Efficientnet-B7}, accuracy drop: \textcolor{red}{\textbf{-18.462}}\% (initial: \textbf{86.154\%}).\\ For \textbf{CLIP VIT-B32}, accuracy drop: \textcolor{red}{\textbf{-44.615\%}} (initial: \textbf{70.769\%}). For \textbf{VIT-B32}, accuracy drop: \textcolor{red}{\textbf{-26.154\%}} (initial: \textbf{92.308\%}).}
\label{fig:appendix_882_1291}
\end{figure}

\begin{figure}[h!]
\centering
\begin{subfigure}{\linewidth}
\centering
\includegraphics[trim=0cm 0cm 0cm 0cm, clip, width=0.9\linewidth]{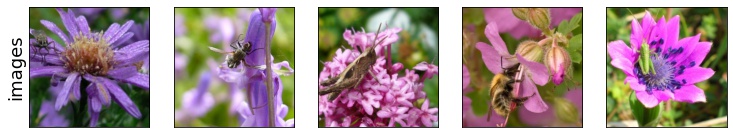}
\end{subfigure}\
\begin{subfigure}{\linewidth}
\centering
\includegraphics[trim=0cm 0cm 0cm 0cm, clip, width=0.9\linewidth]{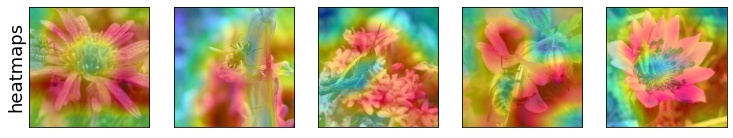}
\end{subfigure}\
\begin{subfigure}{\linewidth}
\centering
\includegraphics[trim=0cm 0cm 0cm 0cm, clip, width=0.9\linewidth]{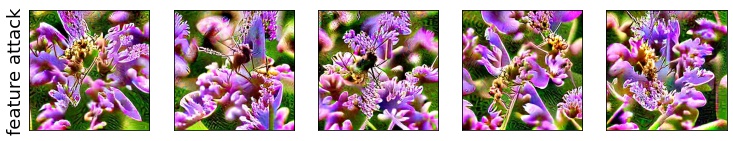}
\end{subfigure}
\caption{Visualization of feature \textbf{1797} for class \textbf{bee} (class index: \textbf{309}).\\ For \textbf{Resnet-50}, accuracy drop: \textcolor{red}{\textbf{-40.0\%}} (initial: \textbf{90.769\%}). For \textbf{Efficientnet-B7}, accuracy drop: \textcolor{red}{\textbf{-15.385}}\% (initial: \textbf{89.231\%}).\\ For \textbf{CLIP VIT-B32}, accuracy drop: \textcolor{red}{\textbf{-36.923\%}} (initial: \textbf{95.385\%}). For \textbf{VIT-B32}, accuracy drop: \textcolor{red}{\textbf{-3.077\%}} (initial: \textbf{89.231\%}).}
\label{fig:appendix_309_1797}
\end{figure}

\begin{figure}[h!]
\centering
\begin{subfigure}{\linewidth}
\centering
\includegraphics[trim=0cm 0cm 0cm 0cm, clip, width=0.9\linewidth]{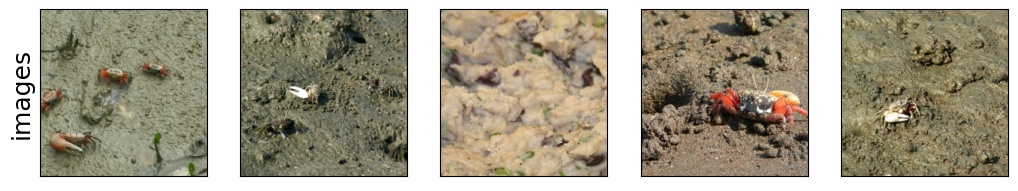}
\end{subfigure}\
\begin{subfigure}{\linewidth}
\centering
\includegraphics[trim=0cm 0cm 0cm 0cm, clip, width=0.9\linewidth]{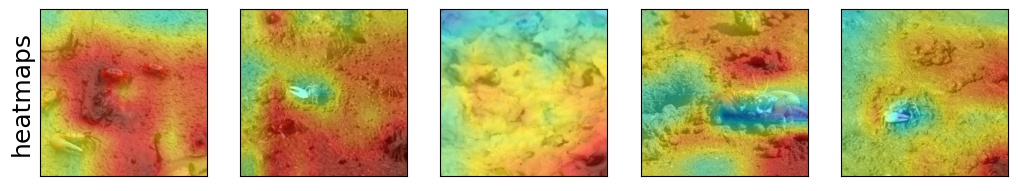}
\end{subfigure}\
\begin{subfigure}{\linewidth}
\centering
\includegraphics[trim=0cm 0cm 0cm 0cm, clip, width=0.9\linewidth]{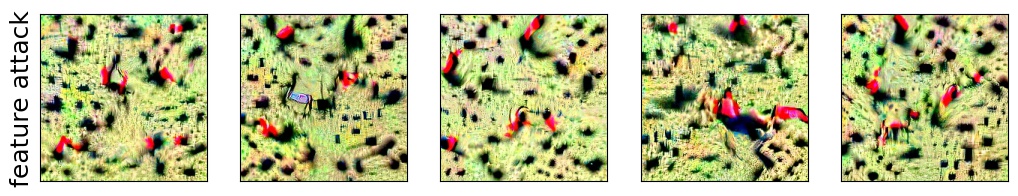}
\end{subfigure}
\caption{Visualization of feature \textbf{870} for class \textbf{fiddler crab} (class index: \textbf{120}).\\ For \textbf{Resnet-50}, accuracy drop: \textcolor{red}{\textbf{-24.615\%}} (initial: \textbf{100.0\%}). For \textbf{Efficientnet-B7}, accuracy drop: \textcolor{red}{\textbf{-18.462}}\% (initial: \textbf{98.462\%}).\\ For \textbf{CLIP VIT-B32}, accuracy drop: \textcolor{red}{\textbf{-61.539\%}} (initial: \textbf{89.231\%}). For \textbf{VIT-B32}, accuracy drop: \textcolor{red}{\textbf{-6.154\%}} (initial: \textbf{95.385\%}).}
\label{fig:appendix_120_870}
\end{figure}

% \begin{figure}[h!]
% \centering
% \begin{subfigure}{\linewidth}
% \centering
% \includegraphics[trim=0cm 0cm 0cm 0cm, clip, width=0.9\linewidth]{visualizations/337_925_images}
% \end{subfigure}\
% \begin{subfigure}{\linewidth}
% \centering
% \includegraphics[trim=0cm 0cm 0cm 0cm, clip, width=0.9\linewidth]{visualizations/337_925_heatmaps}
% \end{subfigure}\
% \begin{subfigure}{\linewidth}
% \centering
% \includegraphics[trim=0cm 0cm 0cm 0cm, clip, width=0.9\linewidth]{visualizations/337_925_attacks}
% \end{subfigure}
% \caption{Visualization of feature \textbf{925} for class \textbf{beaver} (class index: \textbf{337}).\\ For \textbf{Resnet-50}, accuracy drop: \textcolor{red}{\textbf{-27.692\%}} (initial: \textbf{100.0\%}). For \textbf{Efficientnet-B7}, accuracy drop: \textcolor{red}{\textbf{-15.385}}\% (initial: \textbf{96.923\%}).\\ For \textbf{CLIP VIT-B32}, accuracy drop: \textcolor{red}{\textbf{-41.539\%}} (initial: \textbf{72.308\%}). For \textbf{VIT-B32}, accuracy drop: \textcolor{red}{\textbf{-6.154\%}} (initial: \textbf{100.0\%}).}
% \label{fig:appendix_337_925}
% \end{figure}

% \clearpage

\begin{figure}[h!]
\centering
\begin{subfigure}{\linewidth}
\centering
\includegraphics[trim=0cm 0cm 0cm 0cm, clip, width=0.9\linewidth]{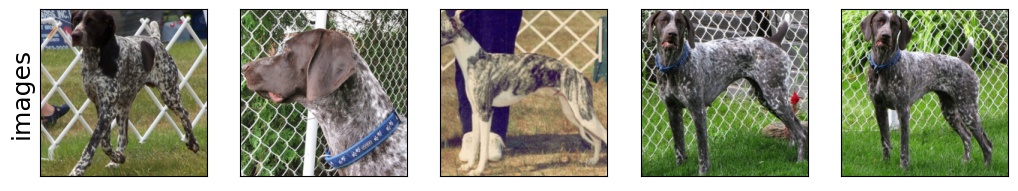}
\end{subfigure}\
\begin{subfigure}{\linewidth}
\centering
\includegraphics[trim=0cm 0cm 0cm 0cm, clip, width=0.9\linewidth]{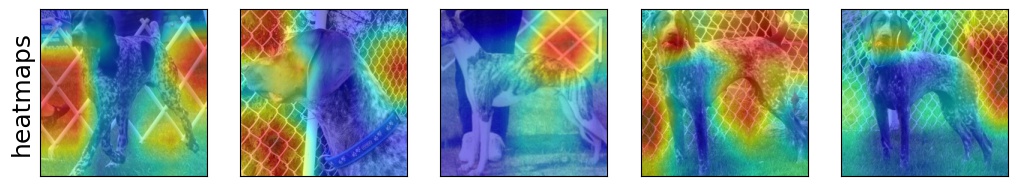}
\end{subfigure}\
\begin{subfigure}{\linewidth}
\centering
\includegraphics[trim=0cm 0cm 0cm 0cm, clip, width=0.9\linewidth]{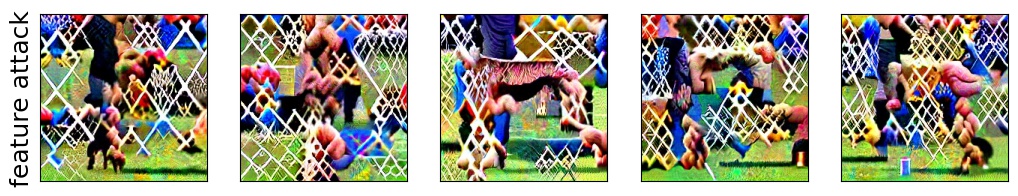}
\end{subfigure}
\caption{Visualization of feature \textbf{1856} for class \textbf{german short haired pointer} (class index: \textbf{210}).\\ For \textbf{Resnet-50}, accuracy drop: \textcolor{red}{\textbf{-27.693\%}} (initial: \textbf{95.385\%}). For \textbf{Efficientnet-B7}, accuracy drop: \textcolor{red}{\textbf{-3.077}}\% (initial: \textbf{98.462\%}).\\ For \textbf{CLIP VIT-B32}, accuracy drop: \textcolor{red}{\textbf{-32.307\%}} (initial: \textbf{81.538\%}). For \textbf{VIT-B32}, accuracy drop: \textcolor{red}{\textbf{-13.846\%}} (initial: \textbf{92.308\%}).}
\label{fig:appendix_210_1856}
\end{figure}

\clearpage

\begin{figure}[h!]
\centering
\begin{subfigure}{\linewidth}
\centering
\includegraphics[trim=0cm 0cm 0cm 0cm, clip, width=0.9\linewidth]{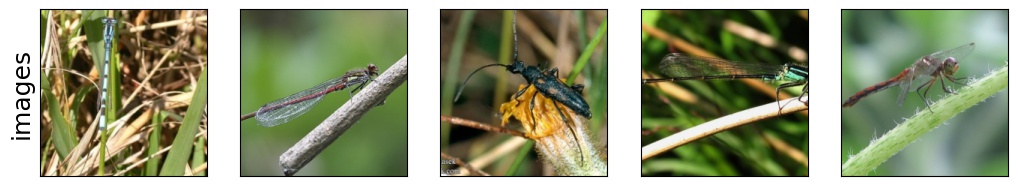}
\end{subfigure}\
\begin{subfigure}{\linewidth}
\centering
\includegraphics[trim=0cm 0cm 0cm 0cm, clip, width=0.9\linewidth]{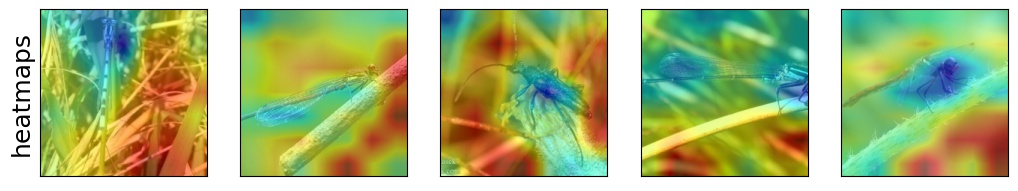}
\end{subfigure}\
\begin{subfigure}{\linewidth}
\centering
\includegraphics[trim=0cm 0cm 0cm 0cm, clip, width=0.9\linewidth]{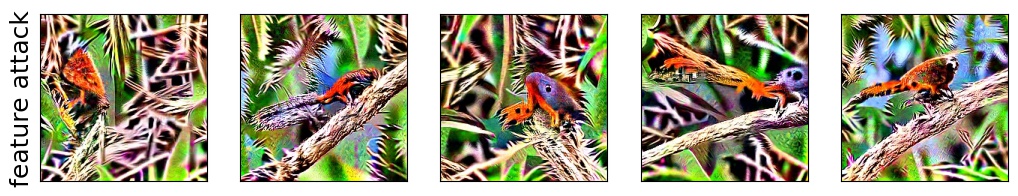}
\end{subfigure}
\caption{Visualization of feature \textbf{1556} for class \textbf{damselfly} (class index: \textbf{320}).\\ For \textbf{Resnet-50}, accuracy drop: \textcolor{red}{\textbf{-30.77\%}} (initial: \textbf{95.385\%}). For \textbf{Efficientnet-B7}, accuracy drop: \textcolor{red}{\textbf{-4.615}}\% (initial: \textbf{96.923\%}).\\ For \textbf{CLIP VIT-B32}, accuracy drop: \textcolor{red}{\textbf{-44.615\%}} (initial: \textbf{93.846\%}). For \textbf{VIT-B32}, accuracy drop: \textcolor{red}{\textbf{-10.769\%}} (initial: \textbf{89.231\%}).}
\label{fig:appendix_320_1556}
\end{figure}

% \clearpage

% \begin{figure}[h!]
% \centering
% \begin{subfigure}{\linewidth}
% \centering
% \includegraphics[trim=0cm 0cm 0cm 0cm, clip, width=0.9\linewidth]{visualizations/403_961_images}
% \end{subfigure}\
% \begin{subfigure}{\linewidth}
% \centering
% \includegraphics[trim=0cm 0cm 0cm 0cm, clip, width=0.9\linewidth]{visualizations/403_961_heatmaps}
% \end{subfigure}\
% \begin{subfigure}{\linewidth}
% \centering
% \includegraphics[trim=0cm 0cm 0cm 0cm, clip, width=0.9\linewidth]{visualizations/403_961_attacks}
% \end{subfigure}
% \caption{Visualization of feature \textbf{961} for class \textbf{aircraft carrier} (class index: \textbf{403}).\\ For \textbf{Resnet-50}, accuracy drop: \textcolor{red}{\textbf{-66.154\%}} (initial: \textbf{98.462\%}). For \textbf{Efficientnet-B7}, accuracy drop: \textcolor{red}{\textbf{-15.384}}\% (initial: \textbf{93.846\%}).\\ For \textbf{CLIP VIT-B32}, accuracy drop: \textcolor{red}{\textbf{-16.923\%}} (initial: \textbf{96.923\%}). For \textbf{VIT-B32}, accuracy drop: \textcolor{red}{\textbf{-23.077\%}} (initial: \textbf{95.385\%}).}
% \label{fig:appendix_403_961}
% \end{figure}

\begin{figure}[h!]
\centering
\begin{subfigure}{\linewidth}
\centering
\includegraphics[trim=0cm 0cm 0cm 0cm, clip, width=0.9\linewidth]{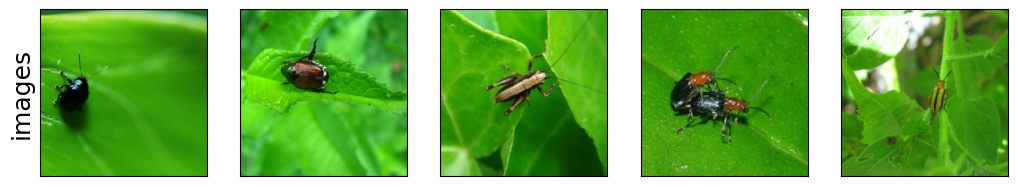}
\end{subfigure}\
\begin{subfigure}{\linewidth}
\centering
\includegraphics[trim=0cm 0cm 0cm 0cm, clip, width=0.9\linewidth]{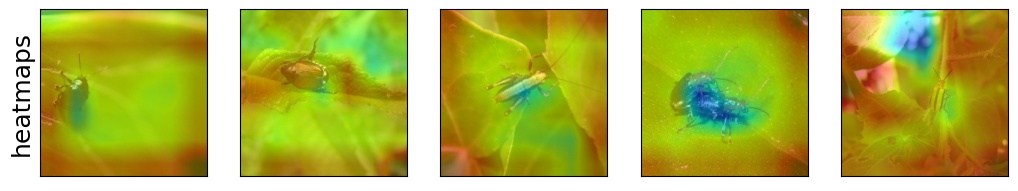}
\end{subfigure}\
\begin{subfigure}{\linewidth}
\centering
\includegraphics[trim=0cm 0cm 0cm 0cm, clip, width=0.9\linewidth]{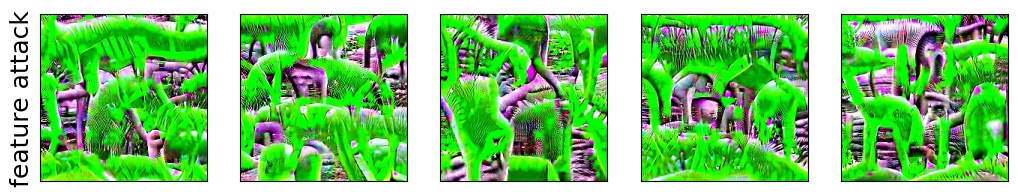}
\end{subfigure}
\caption{Visualization of feature \textbf{1994} for class \textbf{leaf beetle} (class index: \textbf{304}).\\ For \textbf{Resnet-50}, accuracy drop: \textcolor{red}{\textbf{-41.539\%}} (initial: \textbf{86.154\%}). For \textbf{Efficientnet-B7}, accuracy drop: \textcolor{red}{\textbf{-12.308}}\% (initial: \textbf{75.385\%}).\\ For \textbf{CLIP VIT-B32}, accuracy drop: \textcolor{red}{\textbf{-40.0\%}} (initial: \textbf{81.538\%}). For \textbf{VIT-B32}, accuracy drop: \textcolor{blue}{\textbf{+0.0\%}} (initial: \textbf{60.0\%}).}
\label{fig:appendix_304_1994}
\end{figure}

\clearpage
% \clearpage

% \begin{figure}[h!]
% \centering
% \begin{subfigure}{\linewidth}
% \centering
% \includegraphics[trim=0cm 0cm 0cm 0cm, clip, width=0.9\linewidth]{visualizations/756_1829_images}
% \end{subfigure}\
% \begin{subfigure}{\linewidth}
% \centering
% \includegraphics[trim=0cm 0cm 0cm 0cm, clip, width=0.9\linewidth]{visualizations/756_1829_heatmaps}
% \end{subfigure}\
% \begin{subfigure}{\linewidth}
% \centering
% \includegraphics[trim=0cm 0cm 0cm 0cm, clip, width=0.9\linewidth]{visualizations/756_1829_attacks}
% \end{subfigure}
% \caption{Visualization of feature \textbf{1829} for class \textbf{rain barrel} (class index: \textbf{756}).\\ For \textbf{Resnet-50}, accuracy drop: \textcolor{red}{\textbf{-55.385\%}} (initial: \textbf{96.923\%}). For \textbf{Efficientnet-B7}, accuracy drop: \textcolor{red}{\textbf{-15.385}}\% (initial: \textbf{98.462\%}).\\ For \textbf{CLIP VIT-B32}, accuracy drop: \textcolor{red}{\textbf{-18.462\%}} (initial: \textbf{86.154\%}). For \textbf{VIT-B32}, accuracy drop: \textcolor{red}{\textbf{-20.0\%}} (initial: \textbf{98.462\%}).}
% \label{fig:appendix_756_1829}
% \end{figure}

% \clearpage

\begin{figure}[h!]
\centering
\begin{subfigure}{\linewidth}
\centering
\includegraphics[trim=0cm 0cm 0cm 0cm, clip, width=0.9\linewidth]{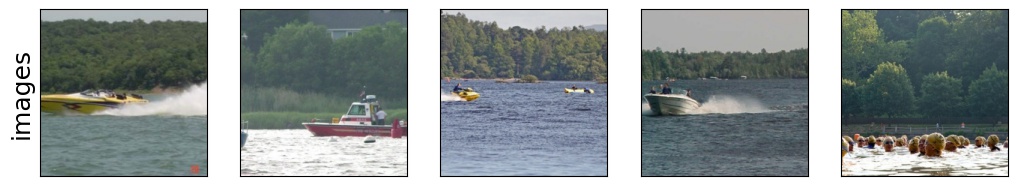}
\end{subfigure}\
\begin{subfigure}{\linewidth}
\centering
\includegraphics[trim=0cm 0cm 0cm 0cm, clip, width=0.9\linewidth]{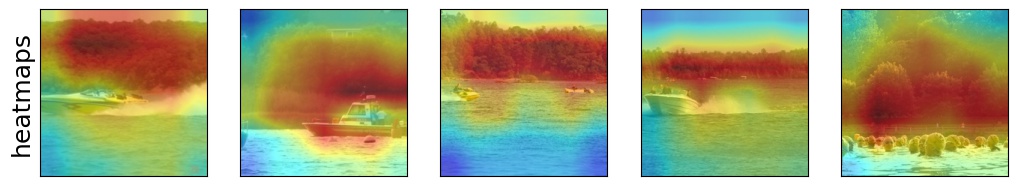}
\end{subfigure}\
\begin{subfigure}{\linewidth}
\centering
\includegraphics[trim=0cm 0cm 0cm 0cm, clip, width=0.9\linewidth]{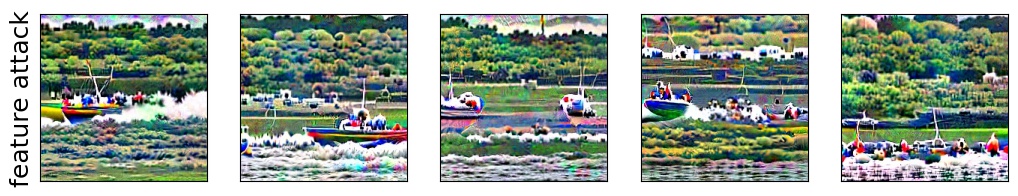}
\end{subfigure}
\caption{Visualization of feature \textbf{1901} for class \textbf{speedboat} (class index: \textbf{814}).\\ For \textbf{Resnet-50}, accuracy drop: \textcolor{red}{\textbf{-56.924\%}} (initial: \textbf{98.462\%}). For \textbf{Efficientnet-B7}, accuracy drop: \textcolor{red}{\textbf{-10.77}}\% (initial: \textbf{98.462\%}).\\ For \textbf{CLIP VIT-B32}, accuracy drop: \textcolor{red}{\textbf{-24.616\%}} (initial: \textbf{38.462\%}). For \textbf{VIT-B32}, accuracy drop: \textcolor{red}{\textbf{-13.846\%}} (initial: \textbf{100.0\%}).}
\label{fig:appendix_814_1901}
\end{figure}

\begin{figure}[h!]
\centering
\begin{subfigure}{\linewidth}
\centering
\includegraphics[trim=0cm 0cm 0cm 0cm, clip, width=0.9\linewidth]{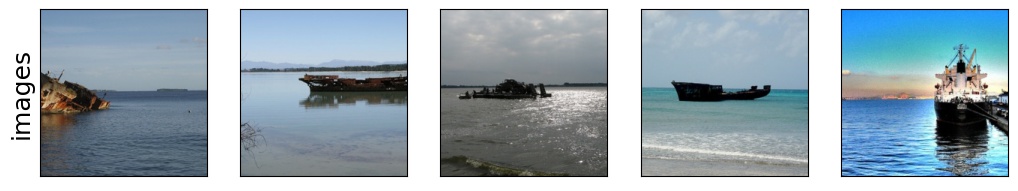}
\end{subfigure}\
\begin{subfigure}{\linewidth}
\centering
\includegraphics[trim=0cm 0cm 0cm 0cm, clip, width=0.9\linewidth]{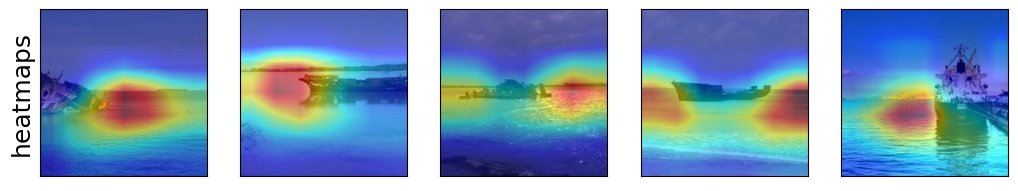}
\end{subfigure}\
\begin{subfigure}{\linewidth}
\centering
\includegraphics[trim=0cm 0cm 0cm 0cm, clip, width=0.9\linewidth]{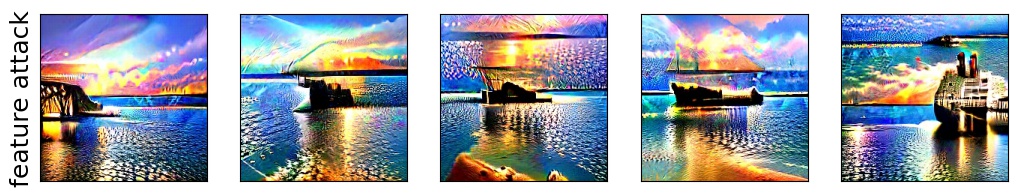}
\end{subfigure}
\caption{Visualization of feature \textbf{421} for class \textbf{wreck} (class index: \textbf{913}).\\ For \textbf{Resnet-50}, accuracy drop: \textcolor{red}{\textbf{-26.154\%}} (initial: \textbf{83.077\%}). For \textbf{Efficientnet-B7}, accuracy drop: \textcolor{red}{\textbf{-21.538}}\% (initial: \textbf{84.615\%}).\\ For \textbf{CLIP VIT-B32}, accuracy drop: \textcolor{red}{\textbf{-35.385\%}} (initial: \textbf{55.385\%}). For \textbf{VIT-B32}, accuracy drop: \textcolor{red}{\textbf{-18.461\%}} (initial: \textbf{67.692\%}).}
\label{fig:appendix_913_421}
\end{figure}

% \begin{figure}[h!]
% \centering
% \begin{subfigure}{\linewidth}
% \centering
% \includegraphics[trim=0cm 0cm 0cm 0cm, clip, width=0.9\linewidth]{visualizations/449_421_images}
% \end{subfigure}\
% \begin{subfigure}{\linewidth}
% \centering
% \includegraphics[trim=0cm 0cm 0cm 0cm, clip, width=0.9\linewidth]{visualizations/449_421_heatmaps}
% \end{subfigure}\
% \begin{subfigure}{\linewidth}
% \centering
% \includegraphics[trim=0cm 0cm 0cm 0cm, clip, width=0.9\linewidth]{visualizations/449_421_attacks}
% \end{subfigure}
% \caption{Visualization of feature \textbf{421} for class \textbf{boathouse} (class index: \textbf{449}).\\ For \textbf{Resnet-50}, accuracy drop: \textcolor{red}{\textbf{-16.923\%}} (initial: \textbf{92.308\%}). For \textbf{Efficientnet-B7}, accuracy drop: \textcolor{red}{\textbf{-12.308}}\% (initial: \textbf{73.846\%}).\\ For \textbf{CLIP VIT-B32}, accuracy drop: \textcolor{red}{\textbf{-29.231\%}} (initial: \textbf{75.385\%}). For \textbf{VIT-B32}, accuracy drop: \textcolor{red}{\textbf{-18.462\%}} (initial: \textbf{80.0\%}).}
% \label{fig:appendix_449_421}
% \end{figure}

\clearpage

% \begin{figure}[h!]
% \centering
% \begin{subfigure}{\linewidth}
% \centering
% \includegraphics[trim=0cm 0cm 0cm 0cm, clip, width=0.9\linewidth]{visualizations/149_1230_images}
% \end{subfigure}\
% \begin{subfigure}{\linewidth}
% \centering
% \includegraphics[trim=0cm 0cm 0cm 0cm, clip, width=0.9\linewidth]{visualizations/149_1230_heatmaps}
% \end{subfigure}\
% \begin{subfigure}{\linewidth}
% \centering
% \includegraphics[trim=0cm 0cm 0cm 0cm, clip, width=0.9\linewidth]{visualizations/149_1230_attacks}
% \end{subfigure}
% \caption{Visualization of feature \textbf{1230} for class \textbf{dugong} (class index: \textbf{149}).\\ For \textbf{Resnet-50}, accuracy drop: \textcolor{red}{\textbf{-24.615\%}} (initial: \textbf{96.923\%}). For \textbf{Efficientnet-B7}, accuracy drop: \textcolor{red}{\textbf{-9.231}}\% (initial: \textbf{96.923\%}).\\ For \textbf{CLIP VIT-B32}, accuracy drop: \textcolor{red}{\textbf{-29.231\%}} (initial: \textbf{98.462\%}). For \textbf{VIT-B32}, accuracy drop: \textcolor{red}{\textbf{-10.769\%}} (initial: \textbf{100.0\%}).}
% \label{fig:appendix_149_1230}
% \end{figure}

% \clearpage

\begin{figure}[h!]
\centering
\begin{subfigure}{\linewidth}
\centering
\includegraphics[trim=0cm 0cm 0cm 0cm, clip, width=0.9\linewidth]{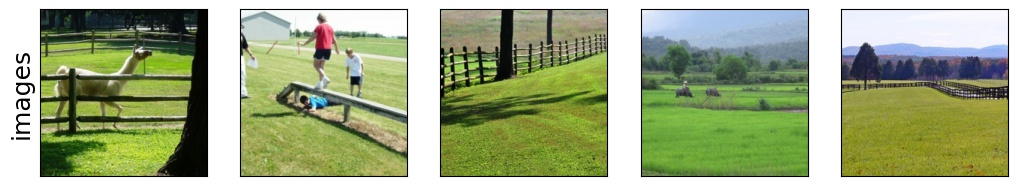}
\end{subfigure}\
\begin{subfigure}{\linewidth}
\centering
\includegraphics[trim=0cm 0cm 0cm 0cm, clip, width=0.9\linewidth]{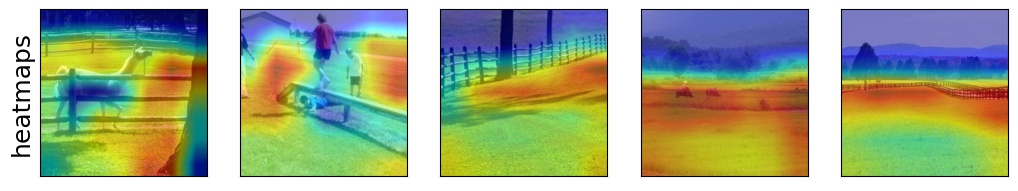}
\end{subfigure}\
\begin{subfigure}{\linewidth}
\centering
\includegraphics[trim=0cm 0cm 0cm 0cm, clip, width=0.9\linewidth]{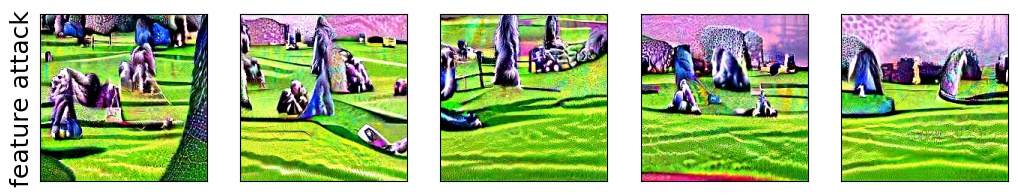}
\end{subfigure}
\caption{Visualization of feature \textbf{1050} for class \textbf{worm fence} (class index: \textbf{912}).\\ For \textbf{Resnet-50}, accuracy drop: \textcolor{red}{\textbf{-26.154\%}} (initial: \textbf{95.385\%}). For \textbf{Efficientnet-B7}, accuracy drop: \textcolor{red}{\textbf{-7.693}}\% (initial: \textbf{95.385\%}).\\ For \textbf{CLIP VIT-B32}, accuracy drop: \textcolor{red}{\textbf{-38.461\%}} (initial: \textbf{76.923\%}). For \textbf{VIT-B32}, accuracy drop: \textcolor{red}{\textbf{-6.154\%}} (initial: \textbf{96.923\%}).}
\label{fig:appendix_912_1050}
\end{figure}

\begin{figure}[h!]
\centering
\begin{subfigure}{\linewidth}
\centering
\includegraphics[trim=0cm 0cm 0cm 0cm, clip, width=0.9\linewidth]{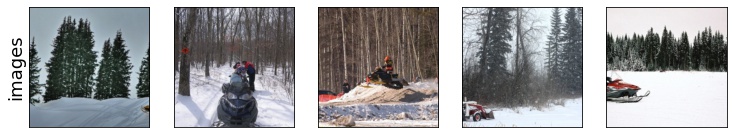}
\end{subfigure}\
\begin{subfigure}{\linewidth}
\centering
\includegraphics[trim=0cm 0cm 0cm 0cm, clip, width=0.9\linewidth]{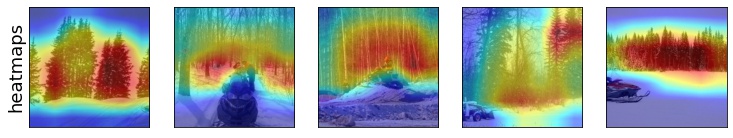}
\end{subfigure}\
\begin{subfigure}{\linewidth}
\centering
\includegraphics[trim=0cm 0cm 0cm 0cm, clip, width=0.9\linewidth]{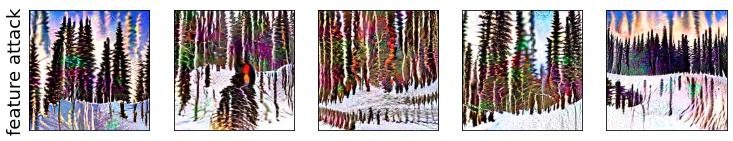}
\end{subfigure}
\caption{Visualization of feature \textbf{0} for class \textbf{snowmobile} (class index: \textbf{802}).\\ For \textbf{Resnet-50}, accuracy drop: \textcolor{red}{\textbf{-21.538\%}} (initial: \textbf{100.0\%}). For \textbf{Efficientnet-B7}, accuracy drop: \textcolor{red}{\textbf{-3.077}}\% (initial: \textbf{98.462\%}).\\ For \textbf{CLIP VIT-B32}, accuracy drop: \textcolor{red}{\textbf{-27.693\%}} (initial: \textbf{89.231\%}). For \textbf{VIT-B32}, accuracy drop: \textcolor{red}{\textbf{-10.769\%}} (initial: \textbf{93.846\%}).}
\label{fig:appendix_802_0}
\end{figure}

\clearpage

\begin{figure}[h!]
\centering
\begin{subfigure}{\linewidth}
\centering
\includegraphics[trim=0cm 0cm 0cm 0cm, clip, width=0.9\linewidth]{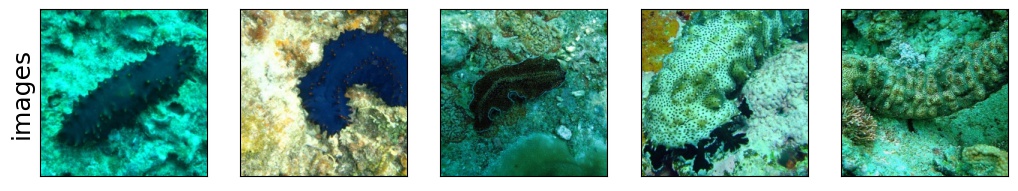}
\end{subfigure}\
\begin{subfigure}{\linewidth}
\centering
\includegraphics[trim=0cm 0cm 0cm 0cm, clip, width=0.9\linewidth]{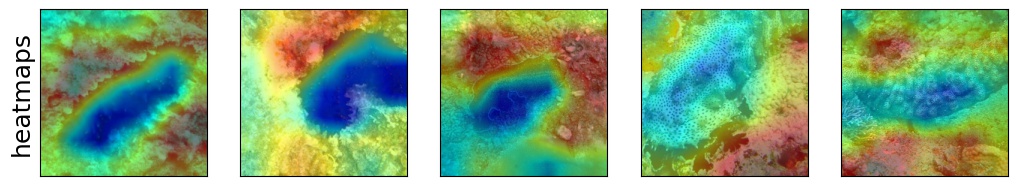}
\end{subfigure}\
\begin{subfigure}{\linewidth}
\centering
\includegraphics[trim=0cm 0cm 0cm 0cm, clip, width=0.9\linewidth]{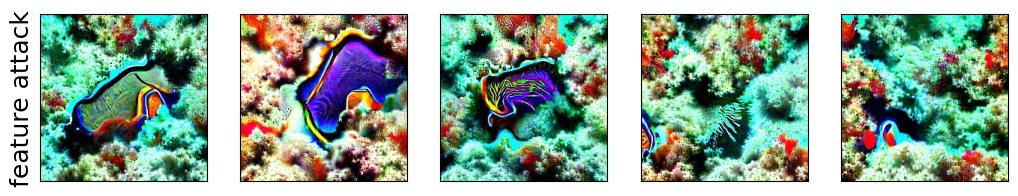}
\end{subfigure}
\caption{Visualization of feature \textbf{1753} for class \textbf{sea cucumber} (class index: \textbf{329}).\\ For \textbf{Resnet-50}, accuracy drop: \textcolor{red}{\textbf{-20.0\%}} (initial: \textbf{93.846\%}). For \textbf{Efficientnet-B7}, accuracy drop: \textcolor{red}{\textbf{-12.307}}\% (initial: \textbf{87.692\%}).\\ For \textbf{CLIP VIT-B32}, accuracy drop: \textcolor{red}{\textbf{-58.462\%}} (initial: \textbf{86.154\%}). For \textbf{VIT-B32}, accuracy drop: \textcolor{red}{\textbf{-1.539\%}} (initial: \textbf{66.154\%}).}
\label{fig:appendix_329_1753}
\end{figure}

% \clearpage

% \begin{figure}[h!]
% \centering
% \begin{subfigure}{\linewidth}
% \centering
% \includegraphics[trim=0cm 0cm 0cm 0cm, clip, width=0.9\linewidth]{visualizations/147_1697_images}
% \end{subfigure}\
% \begin{subfigure}{\linewidth}
% \centering
% \includegraphics[trim=0cm 0cm 0cm 0cm, clip, width=0.9\linewidth]{visualizations/147_1697_heatmaps}
% \end{subfigure}\
% \begin{subfigure}{\linewidth}
% \centering
% \includegraphics[trim=0cm 0cm 0cm 0cm, clip, width=0.9\linewidth]{visualizations/147_1697_attacks}
% \end{subfigure}
% \caption{Visualization of feature \textbf{1697} for class \textbf{grey whale} (class index: \textbf{147}).\\ For \textbf{Resnet-50}, accuracy drop: \textcolor{red}{\textbf{-35.384\%}} (initial: \textbf{87.692\%}). For \textbf{Efficientnet-B7}, accuracy drop: \textcolor{red}{\textbf{-4.615}}\% (initial: \textbf{87.692\%}).\\ For \textbf{CLIP VIT-B32}, accuracy drop: \textcolor{red}{\textbf{-20.0\%}} (initial: \textbf{81.538\%}). For \textbf{VIT-B32}, accuracy drop: \textcolor{red}{\textbf{-12.307\%}} (initial: \textbf{87.692\%}).}
% \label{fig:appendix_147_1697}
% \end{figure}

\begin{figure}[h!]
\centering
\begin{subfigure}{\linewidth}
\centering
\includegraphics[trim=0cm 0cm 0cm 0cm, clip, width=0.9\linewidth]{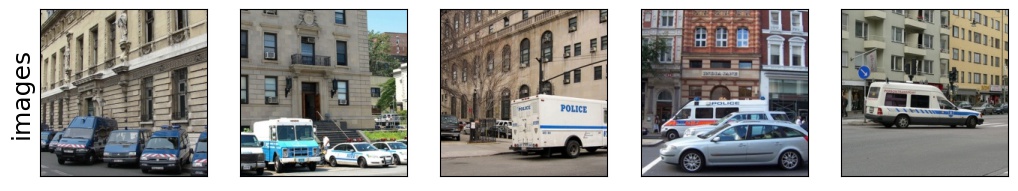}
\end{subfigure}\
\begin{subfigure}{\linewidth}
\centering
\includegraphics[trim=0cm 0cm 0cm 0cm, clip, width=0.9\linewidth]{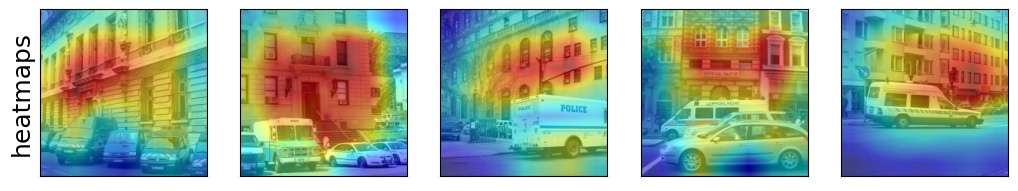}
\end{subfigure}\
\begin{subfigure}{\linewidth}
\centering
\includegraphics[trim=0cm 0cm 0cm 0cm, clip, width=0.9\linewidth]{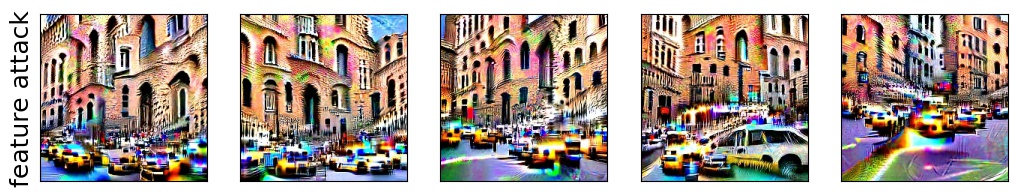}
\end{subfigure}
\caption{Visualization of feature \textbf{1026} for class \textbf{police van} (class index: \textbf{734}).\\ For \textbf{Resnet-50}, accuracy drop: \textcolor{red}{\textbf{-44.616\%}} (initial: \textbf{89.231\%}). For \textbf{Efficientnet-B7}, accuracy drop: \textcolor{red}{\textbf{-16.923}}\% (initial: \textbf{87.692\%}).\\ For \textbf{CLIP VIT-B32}, accuracy drop: \textcolor{red}{\textbf{-3.077\%}} (initial: \textbf{72.308\%}). For \textbf{VIT-B32}, accuracy drop: \textcolor{red}{\textbf{-13.847\%}} (initial: \textbf{78.462\%}).}
\label{fig:appendix_734_1026}
\end{figure}

\clearpage

\subsection{Foreground spurious features}\label{sec:foreground_spurious_appendix}

\begin{figure}[h!]
\centering
\begin{subfigure}{\linewidth}
\centering
\includegraphics[trim=0cm 0cm 0cm 0cm, clip, width=0.9\linewidth]{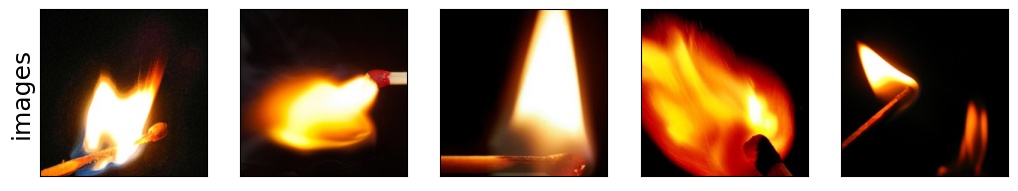}
\end{subfigure}\
\begin{subfigure}{\linewidth}
\centering
\includegraphics[trim=0cm 0cm 0cm 0cm, clip, width=0.9\linewidth]{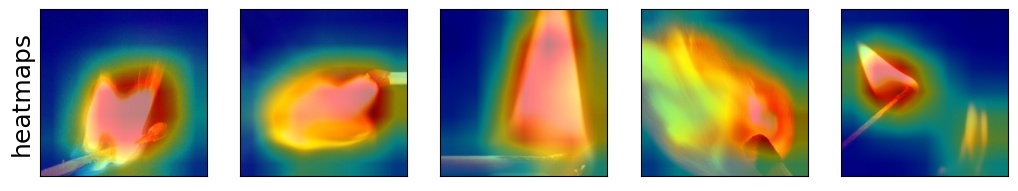}
\end{subfigure}\
\begin{subfigure}{\linewidth}
\centering
\includegraphics[trim=0cm 0cm 0cm 0cm, clip, width=0.9\linewidth]{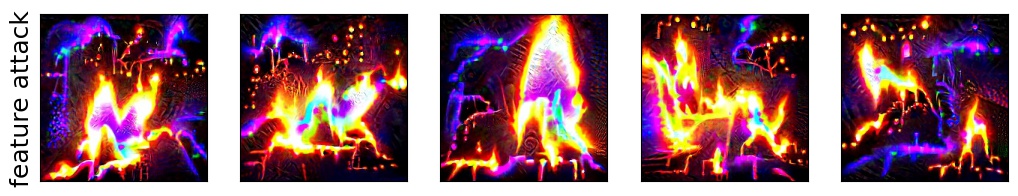}
\end{subfigure}
\caption{Visualization of feature \textbf{1986} for class \textbf{matchstick} (class index: \textbf{644}).\\ For \textbf{Resnet-50}, accuracy drop: \textcolor{red}{\textbf{-43.077\%}} (initial: \textbf{96.923\%}). For \textbf{Efficientnet-B7}, accuracy drop: \textcolor{red}{\textbf{-15.385}}\% (initial: \textbf{98.462\%}).\\ For \textbf{CLIP VIT-B32}, accuracy drop: \textcolor{red}{\textbf{-44.616\%}} (initial: \textbf{55.385\%}). For \textbf{VIT-B32}, accuracy drop: \textcolor{red}{\textbf{-4.616\%}} (initial: \textbf{95.385\%}).}
\label{fig:appendix_644_1986}
\end{figure}

% \clearpage

% \begin{figure}[h!]
% \centering
% \begin{subfigure}{\linewidth}
% \centering
% \includegraphics[trim=0cm 0cm 0cm 0cm, clip, width=0.9\linewidth]{visualizations/913_491_images}
% \end{subfigure}\
% \begin{subfigure}{\linewidth}
% \centering
% \includegraphics[trim=0cm 0cm 0cm 0cm, clip, width=0.9\linewidth]{visualizations/913_491_heatmaps}
% \end{subfigure}\
% \begin{subfigure}{\linewidth}
% \centering
% \includegraphics[trim=0cm 0cm 0cm 0cm, clip, width=0.9\linewidth]{visualizations/913_491_attacks}
% \end{subfigure}
% \caption{Visualization of feature \textbf{491} for class \textbf{wreck} (class index: \textbf{913}).\\ For \textbf{Resnet-50}, accuracy drop: \textcolor{red}{\textbf{-27.692\%}} (initial: \textbf{76.923\%}). For \textbf{Efficientnet-B7}, accuracy drop: \textcolor{red}{\textbf{-30.769}}\% (initial: \textbf{83.077\%}).\\ For \textbf{CLIP VIT-B32}, accuracy drop: \textcolor{red}{\textbf{-32.307\%}} (initial: \textbf{61.538\%}). For \textbf{VIT-B32}, accuracy drop: \textcolor{red}{\textbf{-21.538\%}} (initial: \textbf{67.692\%}).}
% \label{fig:appendix_913_491}
% \end{figure}

\begin{figure}[h!]
\centering
\begin{subfigure}{\linewidth}
\centering
\includegraphics[trim=0cm 0cm 0cm 0cm, clip, width=0.9\linewidth]{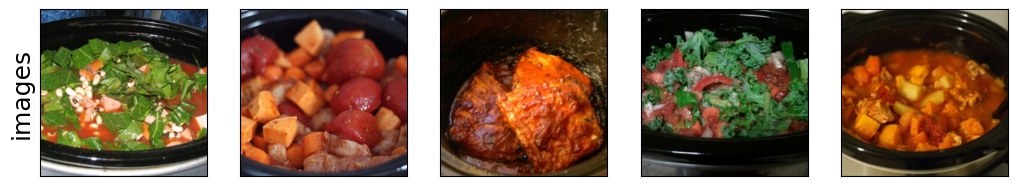}
\end{subfigure}\
\begin{subfigure}{\linewidth}
\centering
\includegraphics[trim=0cm 0cm 0cm 0cm, clip, width=0.9\linewidth]{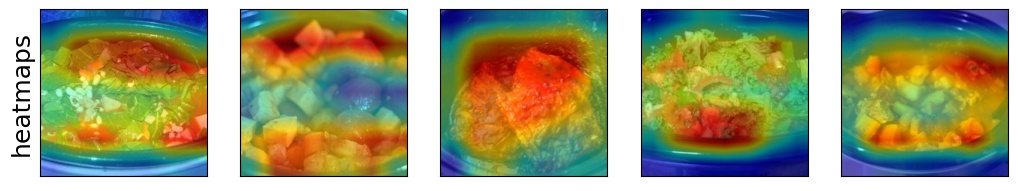}
\end{subfigure}\
\begin{subfigure}{\linewidth}
\centering
\includegraphics[trim=0cm 0cm 0cm 0cm, clip, width=0.9\linewidth]{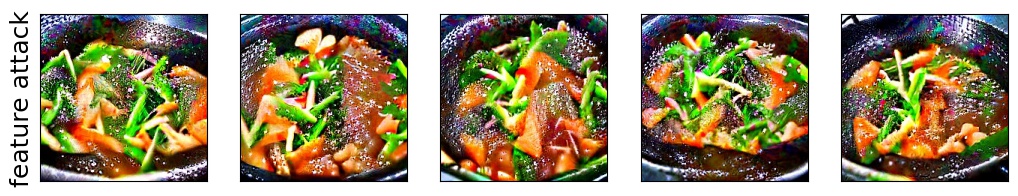}
\end{subfigure}
\caption{Visualization of feature \textbf{895} for class \textbf{crock pot} (class index: \textbf{521}).\\ For \textbf{Resnet-50}, accuracy drop: \textcolor{red}{\textbf{-38.461\%}} (initial: \textbf{67.692\%}). For \textbf{Efficientnet-B7}, accuracy drop: \textcolor{red}{\textbf{-13.846}}\% (initial: \textbf{64.615\%}).\\ For \textbf{CLIP VIT-B32}, accuracy drop: \textcolor{red}{\textbf{-61.538\%}} (initial: \textbf{67.692\%}). For \textbf{VIT-B32}, accuracy drop: \textcolor{red}{\textbf{-20.0\%}} (initial: \textbf{81.538\%}).}
\label{fig:appendix_521_895}
\end{figure}

\clearpage

\begin{figure}[h!]
\centering
\begin{subfigure}{\linewidth}
\centering
\includegraphics[trim=0cm 0cm 0cm 0cm, clip, width=0.9\linewidth]{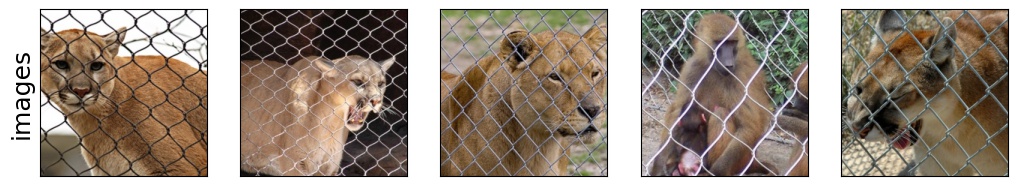}
\end{subfigure}\
\begin{subfigure}{\linewidth}
\centering
\includegraphics[trim=0cm 0cm 0cm 0cm, clip, width=0.9\linewidth]{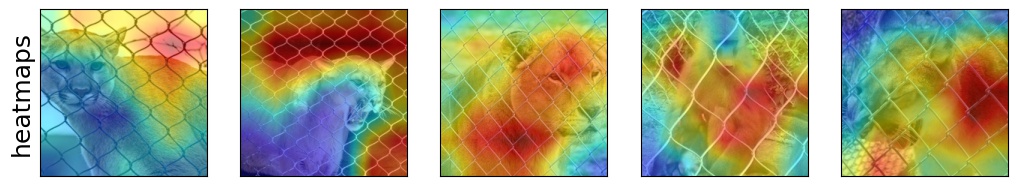}
\end{subfigure}\
\begin{subfigure}{\linewidth}
\centering
\includegraphics[trim=0cm 0cm 0cm 0cm, clip, width=0.9\linewidth]{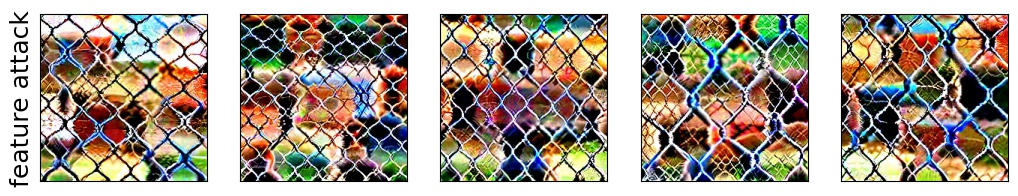}
\end{subfigure}
\caption{Visualization of feature \textbf{1401} for class \textbf{cougar} (class index: \textbf{286}).\\ For \textbf{Resnet-50}, accuracy drop: \textcolor{red}{\textbf{-43.077\%}} (initial: \textbf{96.923\%}). For \textbf{Efficientnet-B7}, accuracy drop: \textcolor{red}{\textbf{-7.692}}\% (initial: \textbf{96.923\%}).\\ For \textbf{CLIP VIT-B32}, accuracy drop: \textcolor{red}{\textbf{-72.308\%}} (initial: \textbf{83.077\%}). For \textbf{VIT-B32}, accuracy drop: \textcolor{red}{\textbf{-6.154\%}} (initial: \textbf{92.308\%}).}
\label{fig:appendix_286_1401}
\end{figure}

\begin{figure}[h!]
\centering
\begin{subfigure}{\linewidth}
\centering
\includegraphics[trim=0cm 0cm 0cm 0cm, clip, width=0.9\linewidth]{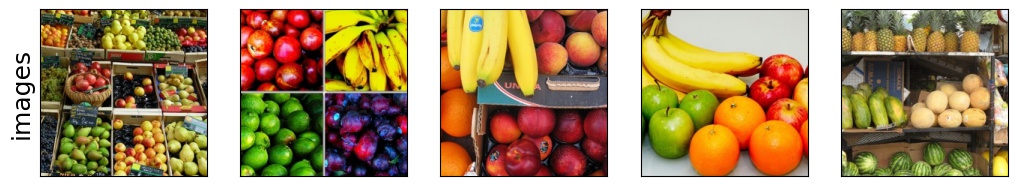}
\end{subfigure}\
\begin{subfigure}{\linewidth}
\centering
\includegraphics[trim=0cm 0cm 0cm 0cm, clip, width=0.9\linewidth]{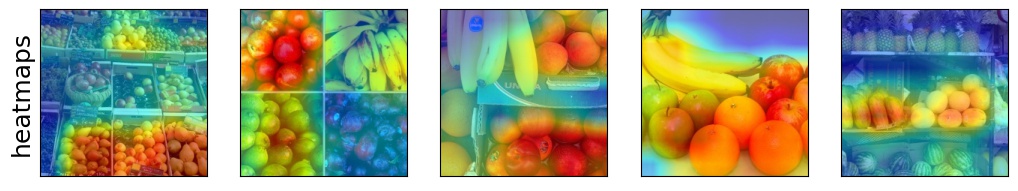}
\end{subfigure}\
\begin{subfigure}{\linewidth}
\centering
\includegraphics[trim=0cm 0cm 0cm 0cm, clip, width=0.9\linewidth]{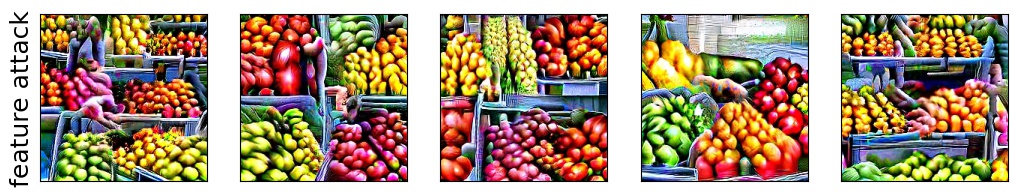}
\end{subfigure}
\caption{Visualization of feature \textbf{68} for class \textbf{banana} (class index: \textbf{954}).\\ For \textbf{Resnet-50}, accuracy drop: \textcolor{red}{\textbf{-46.154\%}} (initial: \textbf{89.231\%}). For \textbf{Efficientnet-B7}, accuracy drop: \textcolor{red}{\textbf{-18.462}}\% (initial: \textbf{92.308\%}).\\ For \textbf{CLIP VIT-B32}, accuracy drop: \textcolor{red}{\textbf{-27.693\%}} (initial: \textbf{52.308\%}). For \textbf{VIT-B32}, accuracy drop: \textcolor{red}{\textbf{-15.384\%}} (initial: \textbf{81.538\%}).}
\label{fig:appendix_954_68}
\end{figure}

\clearpage

% \begin{figure}[h!]
% \centering
% \begin{subfigure}{\linewidth}
% \centering
% \includegraphics[trim=0cm 0cm 0cm 0cm, clip, width=0.9\linewidth]{visualizations/996_1475_images}
% \end{subfigure}\
% \begin{subfigure}{\linewidth}
% \centering
% \includegraphics[trim=0cm 0cm 0cm 0cm, clip, width=0.9\linewidth]{visualizations/996_1475_heatmaps}
% \end{subfigure}\
% \begin{subfigure}{\linewidth}
% \centering
% \includegraphics[trim=0cm 0cm 0cm 0cm, clip, width=0.9\linewidth]{visualizations/996_1475_attacks}
% \end{subfigure}
% \caption{Visualization of feature \textbf{1475} for class \textbf{hen of the woods} (class index: \textbf{996}).\\ For \textbf{Resnet-50}, accuracy drop: \textcolor{red}{\textbf{-47.692\%}} (initial: \textbf{60.0\%}). For \textbf{Efficientnet-B7}, accuracy drop: \textcolor{red}{\textbf{-13.846}}\% (initial: \textbf{70.769\%}).\\ For \textbf{CLIP VIT-B32}, accuracy drop: \textcolor{red}{\textbf{-26.154\%}} (initial: \textbf{29.231\%}). For \textbf{VIT-B32}, accuracy drop: \textcolor{red}{\textbf{-16.924\%}} (initial: \textbf{38.462\%}).}
% \label{fig:appendix_996_1475}
% \end{figure}

\begin{figure}[h!]
\centering
\begin{subfigure}{\linewidth}
\centering
\includegraphics[trim=0cm 0cm 0cm 0cm, clip, width=0.9\linewidth]{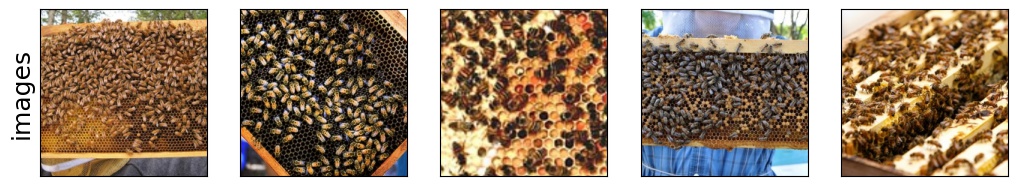}
\end{subfigure}\
\begin{subfigure}{\linewidth}
\centering
\includegraphics[trim=0cm 0cm 0cm 0cm, clip, width=0.9\linewidth]{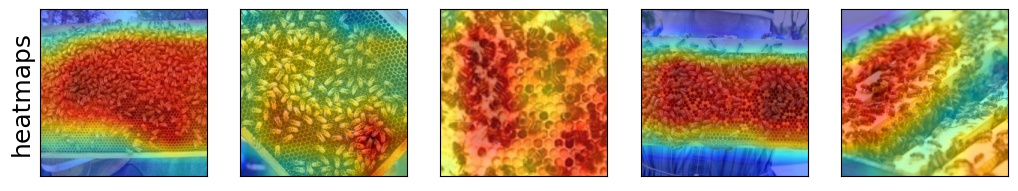}
\end{subfigure}\
\begin{subfigure}{\linewidth}
\centering
\includegraphics[trim=0cm 0cm 0cm 0cm, clip, width=0.9\linewidth]{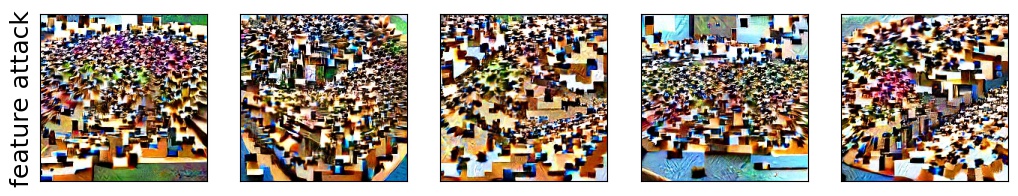}
\end{subfigure}
\caption{Visualization of feature \textbf{1849} for class \textbf{apiary} (class index: \textbf{410}).\\ For \textbf{Resnet-50}, accuracy drop: \textcolor{red}{\textbf{-53.846\%}} (initial: \textbf{87.692\%}). For \textbf{Efficientnet-B7}, accuracy drop: \textcolor{red}{\textbf{-26.153}}\% (initial: \textbf{81.538\%}).\\ For \textbf{CLIP VIT-B32}, accuracy drop: \textcolor{red}{\textbf{-16.923\%}} (initial: \textbf{61.538\%}). For \textbf{VIT-B32}, accuracy drop: \textcolor{red}{\textbf{-15.385\%}} (initial: \textbf{46.154\%}).}
\label{fig:appendix_410_1849}
\end{figure}

\begin{figure}[h!]
\centering
\begin{subfigure}{\linewidth}
\centering
\includegraphics[trim=0cm 0cm 0cm 0cm, clip, width=0.9\linewidth]{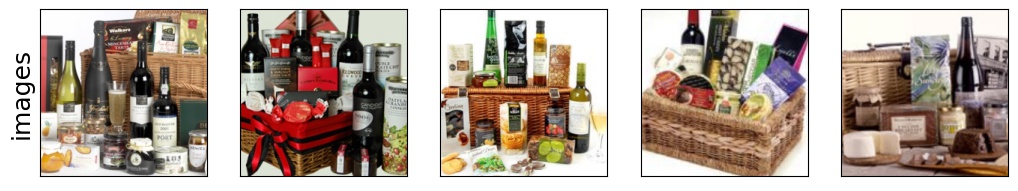}
\end{subfigure}\
\begin{subfigure}{\linewidth}
\centering
\includegraphics[trim=0cm 0cm 0cm 0cm, clip, width=0.9\linewidth]{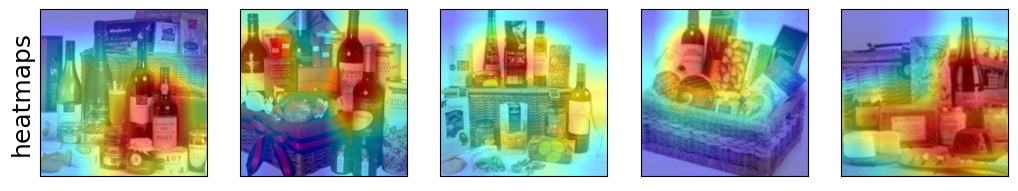}
\end{subfigure}\
\begin{subfigure}{\linewidth}
\centering
\includegraphics[trim=0cm 0cm 0cm 0cm, clip, width=0.9\linewidth]{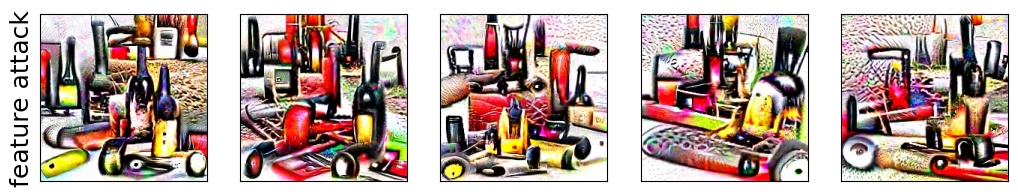}
\end{subfigure}
\caption{Visualization of feature \textbf{834} for class \textbf{hamper} (class index: \textbf{588}).\\ For \textbf{Resnet-50}, accuracy drop: \textcolor{red}{\textbf{-56.923\%}} (initial: \textbf{100.0\%}). For \textbf{Efficientnet-B7}, accuracy drop: \textcolor{red}{\textbf{-27.692}}\% (initial: \textbf{100.0\%}).\\ For \textbf{CLIP VIT-B32}, accuracy drop: \textcolor{red}{\textbf{-7.692\%}} (initial: \textbf{100.0\%}). For \textbf{VIT-B32}, accuracy drop: \textcolor{red}{\textbf{-10.769\%}} (initial: \textbf{93.846\%}).}
\label{fig:appendix_588_834}
\end{figure}

\clearpage

\begin{figure}[h!]
\centering
\begin{subfigure}{\linewidth}
\centering
\includegraphics[trim=0cm 0cm 0cm 0cm, clip, width=0.9\linewidth]{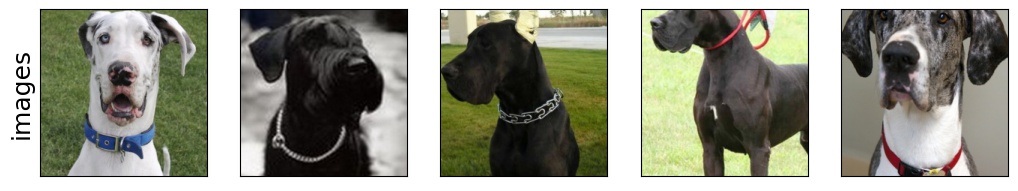}
\end{subfigure}\
\begin{subfigure}{\linewidth}
\centering
\includegraphics[trim=0cm 0cm 0cm 0cm, clip, width=0.9\linewidth]{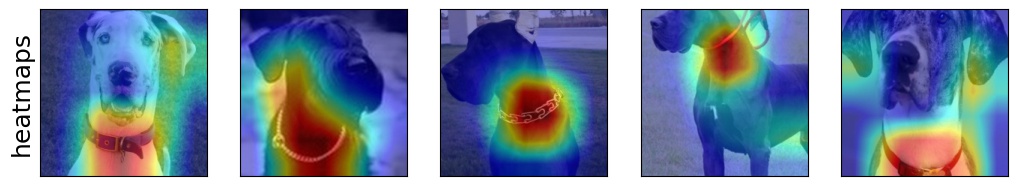}
\end{subfigure}\
\begin{subfigure}{\linewidth}
\centering
\includegraphics[trim=0cm 0cm 0cm 0cm, clip, width=0.9\linewidth]{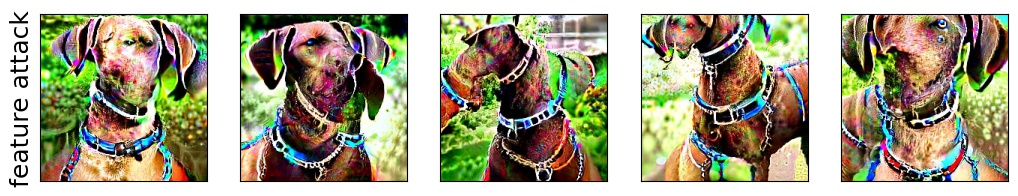}
\end{subfigure}
\caption{Visualization of feature \textbf{1634} for class \textbf{great dane} (class index: \textbf{246}).\\ For \textbf{Resnet-50}, accuracy drop: \textcolor{red}{\textbf{-23.077\%}} (initial: \textbf{95.385\%}). For \textbf{Efficientnet-B7}, accuracy drop: \textcolor{red}{\textbf{-13.846}}\% (initial: \textbf{100.0\%}).\\ For \textbf{CLIP VIT-B32}, accuracy drop: \textcolor{red}{\textbf{-38.461\%}} (initial: \textbf{41.538\%}). For \textbf{VIT-B32}, accuracy drop: \textcolor{red}{\textbf{-10.769\%}} (initial: \textbf{90.769\%}).}
\label{fig:appendix_246_1634}
\end{figure}

% \clearpage

\begin{figure}[h!]
\centering
\begin{subfigure}{\linewidth}
\centering
\includegraphics[trim=0cm 0cm 0cm 0cm, clip, width=0.9\linewidth]{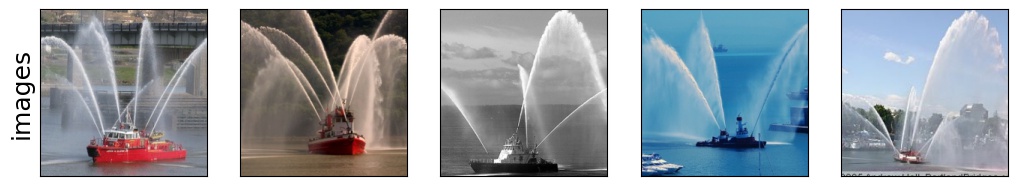}
\end{subfigure}\
\begin{subfigure}{\linewidth}
\centering
\includegraphics[trim=0cm 0cm 0cm 0cm, clip, width=0.9\linewidth]{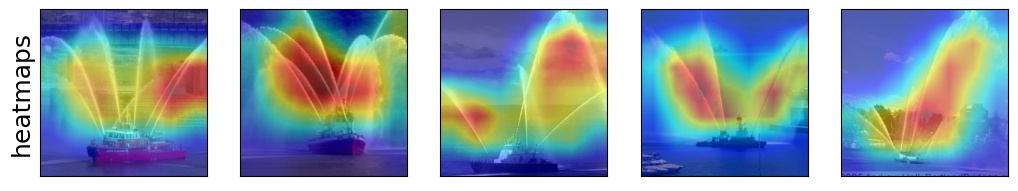}
\end{subfigure}\
\begin{subfigure}{\linewidth}
\centering
\includegraphics[trim=0cm 0cm 0cm 0cm, clip, width=0.9\linewidth]{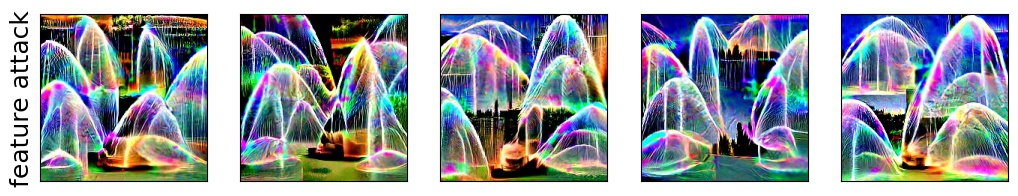}
\end{subfigure}
\caption{Visualization of feature \textbf{2010} for class \textbf{fireboat} (class index: \textbf{554}).\\ For \textbf{Resnet-50}, accuracy drop: \textcolor{red}{\textbf{-33.846\%}} (initial: \textbf{100.0\%}). For \textbf{Efficientnet-B7}, accuracy drop: \textcolor{blue}{\textbf{+0.0}}\% (initial: \textbf{100.0\%}).\\ For \textbf{CLIP VIT-B32}, accuracy drop: \textcolor{red}{\textbf{-36.923\%}} (initial: \textbf{96.923\%}). For \textbf{VIT-B32}, accuracy drop: \textcolor{red}{\textbf{-3.077\%}} (initial: \textbf{100.0\%}).}
\label{fig:appendix_554_2010}
\end{figure}

\clearpage

\begin{figure}[h!]
\centering
\begin{subfigure}{\linewidth}
\centering
\includegraphics[trim=0cm 0cm 0cm 0cm, clip, width=0.9\linewidth]{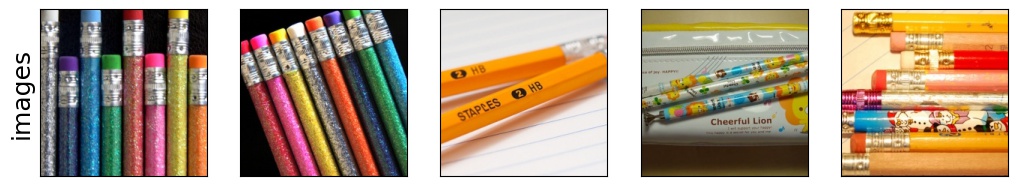}
\end{subfigure}\
\begin{subfigure}{\linewidth}
\centering
\includegraphics[trim=0cm 0cm 0cm 0cm, clip, width=0.9\linewidth]{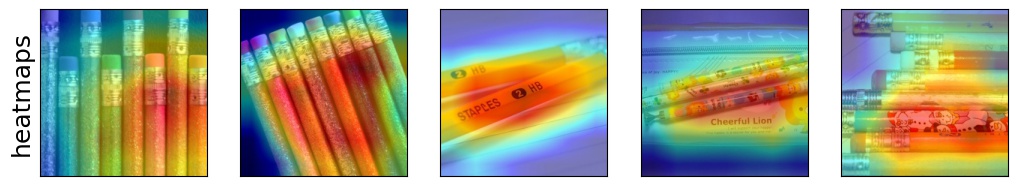}
\end{subfigure}\
\begin{subfigure}{\linewidth}
\centering
\includegraphics[trim=0cm 0cm 0cm 0cm, clip, width=0.9\linewidth]{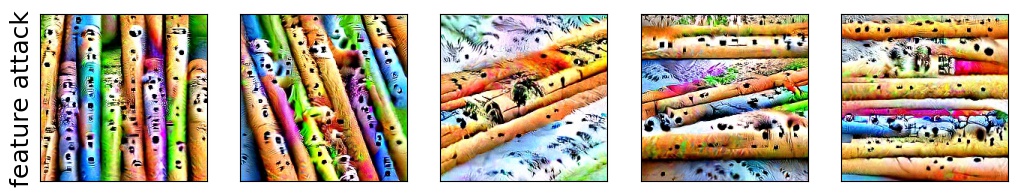}
\end{subfigure}
\caption{Visualization of feature \textbf{1052} for class \textbf{rubber eraser} (class index: \textbf{767}).\\ For \textbf{Resnet-50}, accuracy drop: \textcolor{red}{\textbf{-47.692\%}} (initial: \textbf{86.154\%}). For \textbf{Efficientnet-B7}, accuracy drop: \textcolor{red}{\textbf{-16.923}}\% (initial: \textbf{92.308\%}).\\ For \textbf{CLIP VIT-B32}, accuracy drop: \textcolor{red}{\textbf{-4.615\%}} (initial: \textbf{10.769\%}). For \textbf{VIT-B32}, accuracy drop: \textcolor{red}{\textbf{-18.462\%}} (initial: \textbf{80.0\%}).}
\label{fig:appendix_767_1052}
\end{figure}

\begin{figure}[h!]
\centering
\begin{subfigure}{\linewidth}
\centering
\includegraphics[trim=0cm 0cm 0cm 0cm, clip, width=0.9\linewidth]{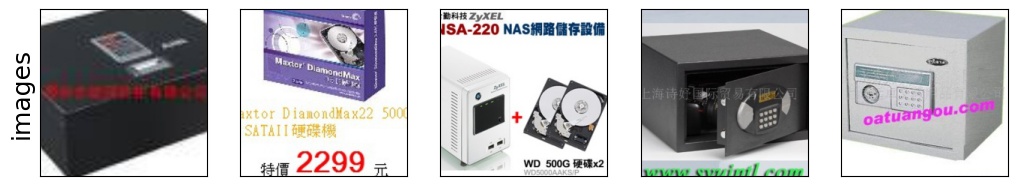}
\end{subfigure}\
\begin{subfigure}{\linewidth}
\centering
\includegraphics[trim=0cm 0cm 0cm 0cm, clip, width=0.9\linewidth]{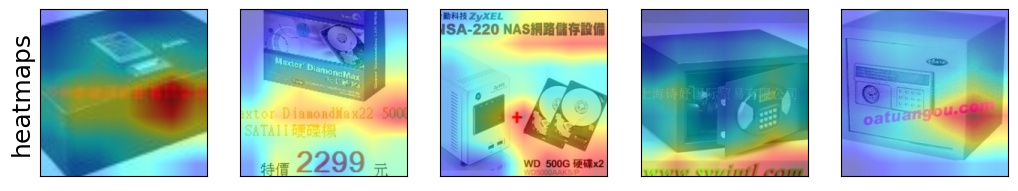}
\end{subfigure}\
\begin{subfigure}{\linewidth}
\centering
\includegraphics[trim=0cm 0cm 0cm 0cm, clip, width=0.9\linewidth]{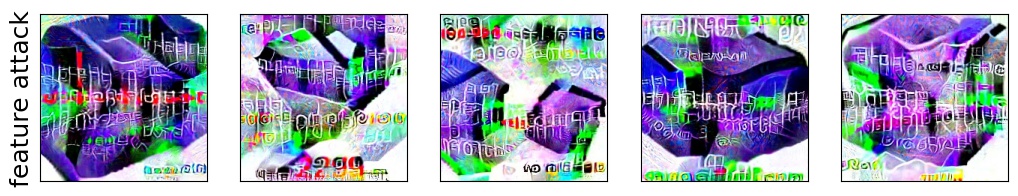}
\end{subfigure}
\caption{Visualization of feature \textbf{388} for class \textbf{safe} (class index: \textbf{771}).\\ For \textbf{Resnet-50}, accuracy drop: \textcolor{red}{\textbf{-26.154\%}} (initial: \textbf{89.231\%}). For \textbf{Efficientnet-B7}, accuracy drop: \textcolor{red}{\textbf{-6.154}}\% (initial: \textbf{86.154\%}).\\ For \textbf{CLIP VIT-B32}, accuracy drop: \textcolor{red}{\textbf{-30.77\%}} (initial: \textbf{38.462\%}). For \textbf{VIT-B32}, accuracy drop: \textcolor{red}{\textbf{-1.539\%}} (initial: \textbf{83.077\%}).}
\label{fig:appendix_771_388}
\end{figure}

\clearpage

\begin{figure}[h!]
\centering
\begin{subfigure}{\linewidth}
\centering
\includegraphics[trim=0cm 0cm 0cm 0cm, clip, width=0.9\linewidth]{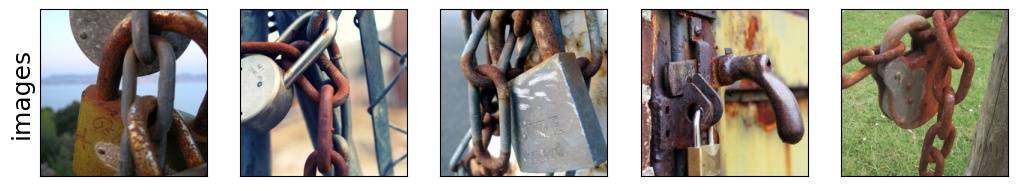}
\end{subfigure}\
\begin{subfigure}{\linewidth}
\centering
\includegraphics[trim=0cm 0cm 0cm 0cm, clip, width=0.9\linewidth]{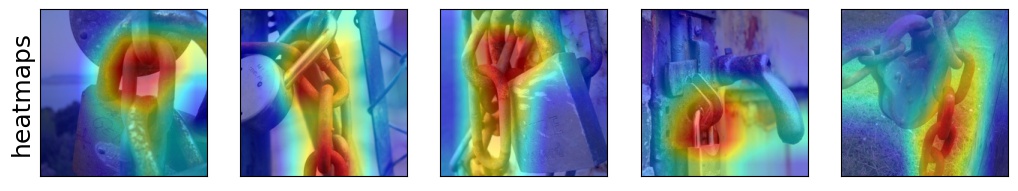}
\end{subfigure}\
\begin{subfigure}{\linewidth}
\centering
\includegraphics[trim=0cm 0cm 0cm 0cm, clip, width=0.9\linewidth]{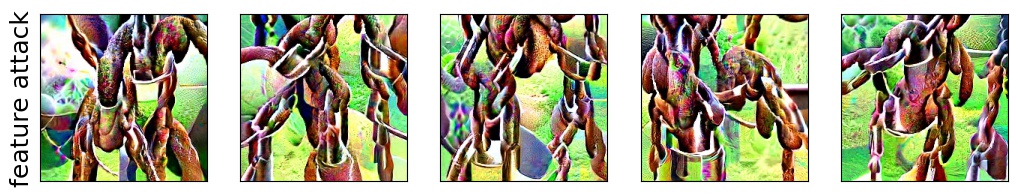}
\end{subfigure}
\caption{Visualization of feature \textbf{128} for class \textbf{padlock} (class index: \textbf{695}).\\ For \textbf{Resnet-50}, accuracy drop: \textcolor{red}{\textbf{-12.307\%}} (initial: \textbf{87.692\%}). For \textbf{Efficientnet-B7}, accuracy drop: \textcolor{red}{\textbf{-15.385}}\% (initial: \textbf{95.385\%}).\\ For \textbf{CLIP VIT-B32}, accuracy drop: \textcolor{red}{\textbf{-63.077\%}} (initial: \textbf{92.308\%}). For \textbf{VIT-B32}, accuracy drop: \textcolor{red}{\textbf{-4.616\%}} (initial: \textbf{86.154\%}).}
\label{fig:appendix_695_128}
\end{figure}

% \clearpage

\begin{figure}[h!]
\centering
\begin{subfigure}{\linewidth}
\centering
\includegraphics[trim=0cm 0cm 0cm 0cm, clip, width=0.9\linewidth]{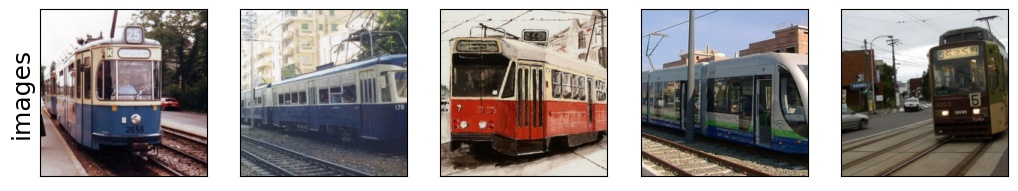}
\end{subfigure}\
\begin{subfigure}{\linewidth}
\centering
\includegraphics[trim=0cm 0cm 0cm 0cm, clip, width=0.9\linewidth]{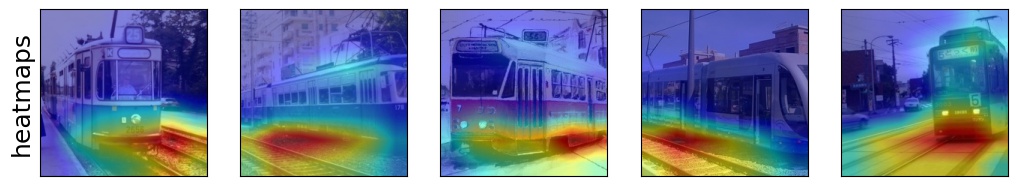}
\end{subfigure}\
\begin{subfigure}{\linewidth}
\centering
\includegraphics[trim=0cm 0cm 0cm 0cm, clip, width=0.9\linewidth]{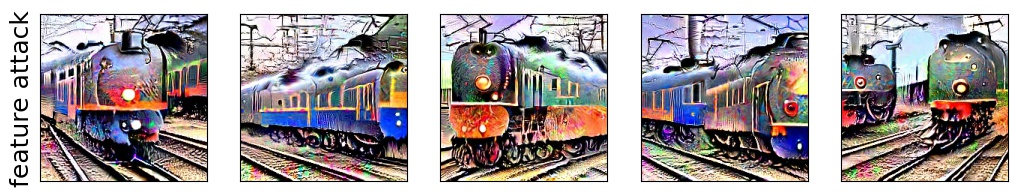}
\end{subfigure}
\caption{Visualization of feature \textbf{56} for class \textbf{streetcar} (class index: \textbf{829}).\\ For \textbf{Resnet-50}, accuracy drop: \textcolor{red}{\textbf{-12.307\%}} (initial: \textbf{87.692\%}). For \textbf{Efficientnet-B7}, accuracy drop: \textcolor{red}{\textbf{-15.384}}\% (initial: \textbf{93.846\%}).\\ For \textbf{CLIP VIT-B32}, accuracy drop: \textcolor{red}{\textbf{-29.231\%}} (initial: \textbf{92.308\%}). For \textbf{VIT-B32}, accuracy drop: \textcolor{red}{\textbf{-6.154\%}} (initial: \textbf{93.846\%}).}
\label{fig:appendix_829_56}
\end{figure}

\begin{figure}[h!]
\centering
\begin{subfigure}{\linewidth}
\centering
\includegraphics[trim=0cm 0cm 0cm 0cm, clip, width=0.9\linewidth]{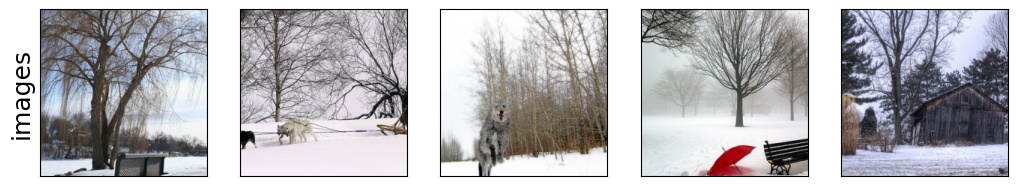}
\end{subfigure}\
\begin{subfigure}{\linewidth}
\centering
\includegraphics[trim=0cm 0cm 0cm 0cm, clip, width=0.9\linewidth]{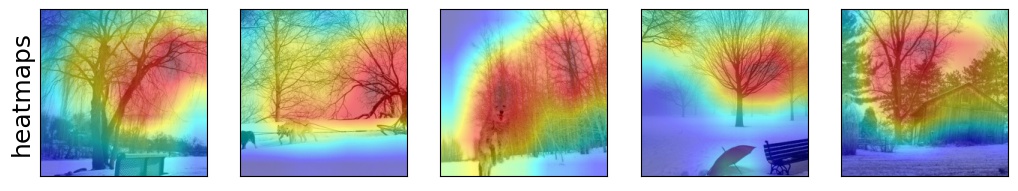}
\end{subfigure}\
\begin{subfigure}{\linewidth}
\centering
\includegraphics[trim=0cm 0cm 0cm 0cm, clip, width=0.9\linewidth]{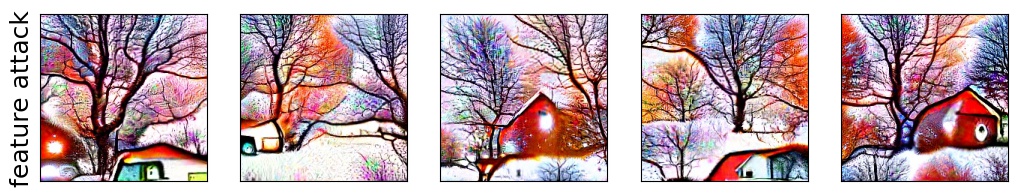}
\end{subfigure}
\caption{Visualization of feature \textbf{1623} for class \textbf{snowplow} (class index: \textbf{803}).\\ For \textbf{Resnet-50}, accuracy drop: \textcolor{red}{\textbf{-18.461\%}} (initial: \textbf{96.923\%}). For \textbf{Efficientnet-B7}, accuracy drop: \textcolor{red}{\textbf{-6.154}}\% (initial: \textbf{95.385\%}).\\ For \textbf{CLIP VIT-B32}, accuracy drop: \textcolor{red}{\textbf{-38.461\%}} (initial: \textbf{93.846\%}). For \textbf{VIT-B32}, accuracy drop: \textcolor{red}{\textbf{-4.615\%}} (initial: \textbf{96.923\%}).}
\label{fig:appendix_803_1623}
\end{figure}

% \clearpage

% \begin{figure}[h!]
% \centering
% \begin{subfigure}{\linewidth}
% \centering
% \includegraphics[trim=0cm 0cm 0cm 0cm, clip, width=0.9\linewidth]{visualizations/542_1165_images}
% \end{subfigure}\
% \begin{subfigure}{\linewidth}
% \centering
% \includegraphics[trim=0cm 0cm 0cm 0cm, clip, width=0.9\linewidth]{visualizations/542_1165_heatmaps}
% \end{subfigure}\
% \begin{subfigure}{\linewidth}
% \centering
% \includegraphics[trim=0cm 0cm 0cm 0cm, clip, width=0.9\linewidth]{visualizations/542_1165_attacks}
% \end{subfigure}
% \caption{Visualization of feature \textbf{1165} for class \textbf{drumstick} (class index: \textbf{542}).\\ For \textbf{Resnet-50}, accuracy drop: \textcolor{red}{\textbf{-24.615\%}} (initial: \textbf{61.538\%}). For \textbf{Efficientnet-B7}, accuracy drop: \textcolor{blue}{\textbf{+1.539}}\% (initial: \textbf{41.538\%}).\\ For \textbf{CLIP VIT-B32}, accuracy drop: \textcolor{red}{\textbf{-13.846\%}} (initial: \textbf{20.0\%}). For \textbf{VIT-B32}, accuracy drop: \textcolor{red}{\textbf{-9.231\%}} (initial: \textbf{66.154\%}).}
% \label{fig:appendix_542_1165}
% \end{figure}

\begin{figure}[h!]
\centering
\begin{subfigure}{\linewidth}
\centering
\includegraphics[trim=0cm 0cm 0cm 0cm, clip, width=0.9\linewidth]{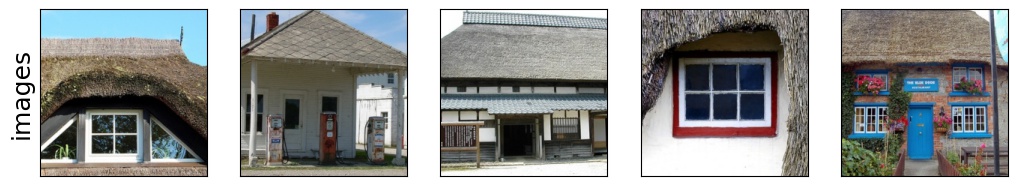}
\end{subfigure}\
\begin{subfigure}{\linewidth}
\centering
\includegraphics[trim=0cm 0cm 0cm 0cm, clip, width=0.9\linewidth]{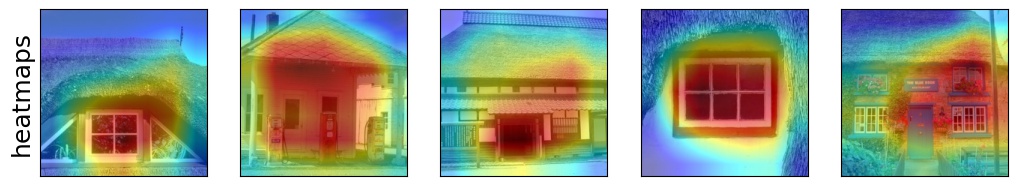}
\end{subfigure}\
\begin{subfigure}{\linewidth}
\centering
\includegraphics[trim=0cm 0cm 0cm 0cm, clip, width=0.9\linewidth]{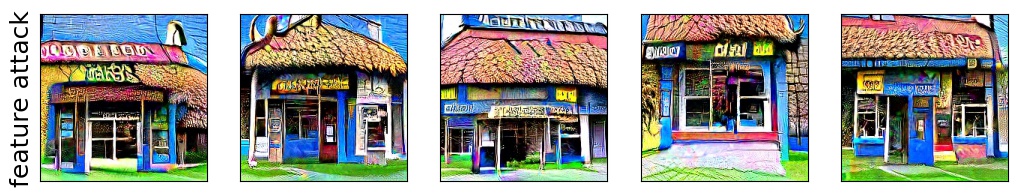}
\end{subfigure}
\caption{Visualization of feature \textbf{466} for class \textbf{thatch} (class index: \textbf{853}).\\ For \textbf{Resnet-50}, accuracy drop: \textcolor{red}{\textbf{-13.846\%}} (initial: \textbf{100.0\%}). For \textbf{Efficientnet-B7}, accuracy drop: \textcolor{red}{\textbf{-7.692}}\% (initial: \textbf{100.0\%}).\\ For \textbf{CLIP VIT-B32}, accuracy drop: \textcolor{red}{\textbf{-15.385\%}} (initial: \textbf{98.462\%}). For \textbf{VIT-B32}, accuracy drop: \textcolor{red}{\textbf{-4.615\%}} (initial: \textbf{96.923\%}).}
\label{fig:appendix_853_466}
\end{figure}

\clearpage

\begin{figure}[h!]
\centering
\begin{subfigure}{\linewidth}
\centering
\includegraphics[trim=0cm 0cm 0cm 0cm, clip, width=0.9\linewidth]{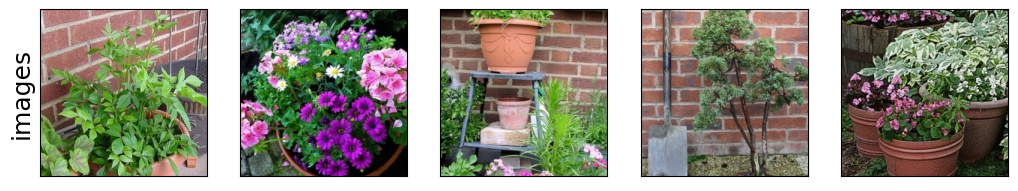}
\end{subfigure}\
\begin{subfigure}{\linewidth}
\centering
\includegraphics[trim=0cm 0cm 0cm 0cm, clip, width=0.9\linewidth]{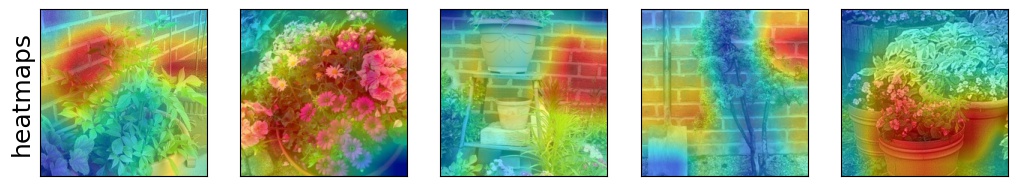}
\end{subfigure}\
\begin{subfigure}{\linewidth}
\centering
\includegraphics[trim=0cm 0cm 0cm 0cm, clip, width=0.9\linewidth]{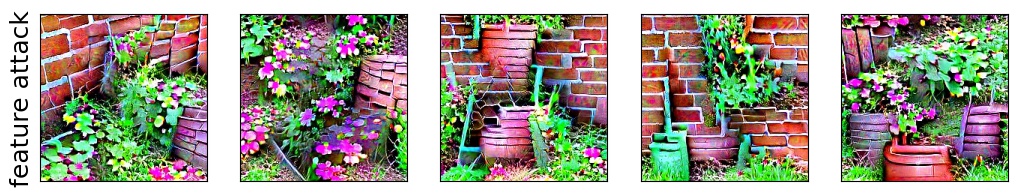}
\end{subfigure}
\caption{Visualization of feature \textbf{1829} for class \textbf{pot} (class index: \textbf{738}).\\ For \textbf{Resnet-50}, accuracy drop: \textcolor{red}{\textbf{-35.385\%}} (initial: \textbf{95.385\%}). For \textbf{Efficientnet-B7}, accuracy drop: \textcolor{red}{\textbf{-7.693}}\% (initial: \textbf{92.308\%}).\\ For \textbf{CLIP VIT-B32}, accuracy drop: \textcolor{blue}{\textbf{+1.538\%}} (initial: \textbf{60.0\%}). For \textbf{VIT-B32}, accuracy drop: \textcolor{red}{\textbf{-12.308\%}} (initial: \textbf{96.923\%}).}
\label{fig:appendix_738_1829}
\end{figure}

\begin{figure}[h!]
\centering
\begin{subfigure}{\linewidth}
\centering
\includegraphics[trim=0cm 0cm 0cm 0cm, clip, width=0.9\linewidth]{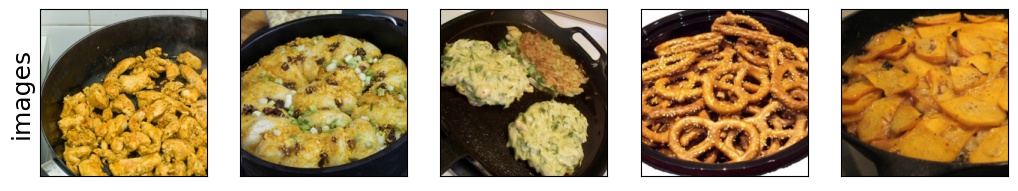}
\end{subfigure}\
\begin{subfigure}{\linewidth}
\centering
\includegraphics[trim=0cm 0cm 0cm 0cm, clip, width=0.9\linewidth]{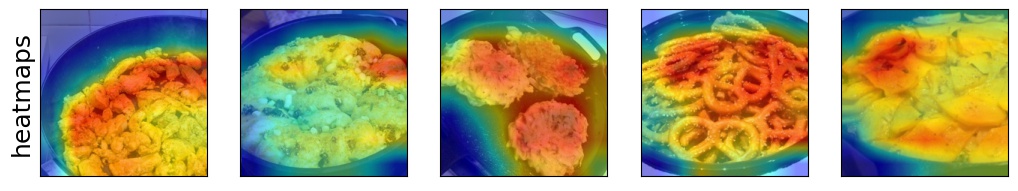}
\end{subfigure}\
\begin{subfigure}{\linewidth}
\centering
\includegraphics[trim=0cm 0cm 0cm 0cm, clip, width=0.9\linewidth]{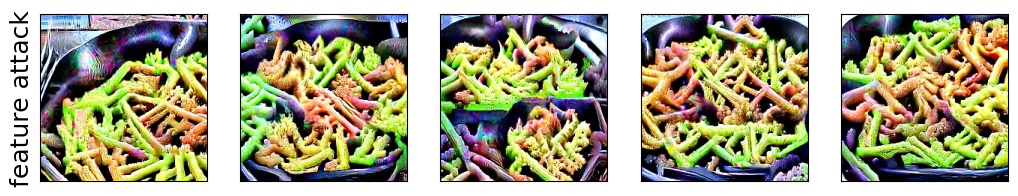}
\end{subfigure}
\caption{Visualization of feature \textbf{2000} for class \textbf{frying pan} (class index: \textbf{567}).\\ For \textbf{Resnet-50}, accuracy drop: \textcolor{red}{\textbf{-9.23\%}} (initial: \textbf{84.615\%}). For \textbf{Efficientnet-B7}, accuracy drop: \textcolor{red}{\textbf{-20.0}}\% (initial: \textbf{72.308\%}).\\ For \textbf{CLIP VIT-B32}, accuracy drop: \textcolor{red}{\textbf{-10.77\%}} (initial: \textbf{12.308\%}). For \textbf{VIT-B32}, accuracy drop: \textcolor{blue}{\textbf{+3.077\%}} (initial: \textbf{46.154\%}).}
\label{fig:appendix_567_2000}
\end{figure}

\clearpage

% \begin{figure}[h!]
% \centering
% \begin{subfigure}{\linewidth}
% \centering
% \includegraphics[trim=0cm 0cm 0cm 0cm, clip, width=0.9\linewidth]{visualizations/706_1633_images}
% \end{subfigure}\
% \begin{subfigure}{\linewidth}
% \centering
% \includegraphics[trim=0cm 0cm 0cm 0cm, clip, width=0.9\linewidth]{visualizations/706_1633_heatmaps}
% \end{subfigure}\
% \begin{subfigure}{\linewidth}
% \centering
% \includegraphics[trim=0cm 0cm 0cm 0cm, clip, width=0.9\linewidth]{visualizations/706_1633_attacks}
% \end{subfigure}
% \caption{Visualization of feature \textbf{1633} for class \textbf{patio} (class index: \textbf{706}).\\ For \textbf{Resnet-50}, accuracy drop: \textcolor{red}{\textbf{-13.846\%}} (initial: \textbf{87.692\%}). For \textbf{Efficientnet-B7}, accuracy drop: \textcolor{red}{\textbf{-6.153}}\% (initial: \textbf{81.538\%}).\\ For \textbf{CLIP VIT-B32}, accuracy drop: \textcolor{red}{\textbf{-26.153\%}} (initial: \textbf{84.615\%}). For \textbf{VIT-B32}, accuracy drop: \textcolor{red}{\textbf{-6.153\%}} (initial: \textbf{81.538\%}).}
% \label{fig:appendix_706_1633}
% \end{figure}

\begin{figure}[h!]
\centering
\begin{subfigure}{\linewidth}
\centering
\includegraphics[trim=0cm 0cm 0cm 0cm, clip, width=0.9\linewidth]{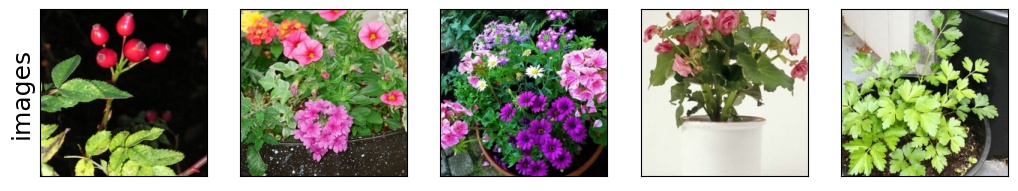}
\end{subfigure}\
\begin{subfigure}{\linewidth}
\centering
\includegraphics[trim=0cm 0cm 0cm 0cm, clip, width=0.9\linewidth]{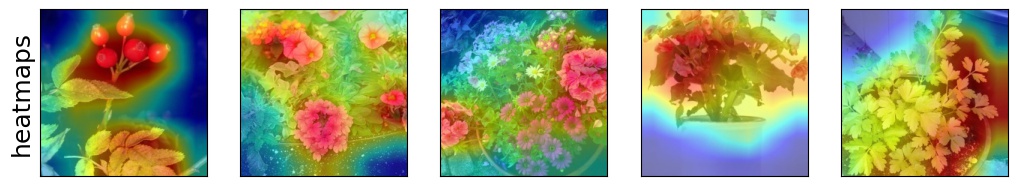}
\end{subfigure}\
\begin{subfigure}{\linewidth}
\centering
\includegraphics[trim=0cm 0cm 0cm 0cm, clip, width=0.9\linewidth]{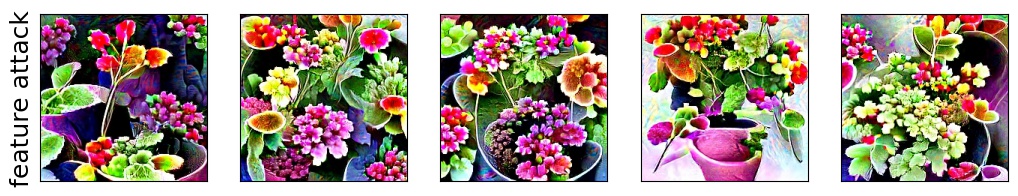}
\end{subfigure}
\caption{Visualization of feature \textbf{1392} for class \textbf{pot} (class index: \textbf{738}).\\ For \textbf{Resnet-50}, accuracy drop: \textcolor{red}{\textbf{-41.538\%}} (initial: \textbf{90.769\%}). For \textbf{Efficientnet-B7}, accuracy drop: \textcolor{red}{\textbf{-6.154}}\% (initial: \textbf{93.846\%}).\\ For \textbf{CLIP VIT-B32}, accuracy drop: \textcolor{red}{\textbf{-12.308\%}} (initial: \textbf{69.231\%}). For \textbf{VIT-B32}, accuracy drop: \textcolor{red}{\textbf{-6.154\%}} (initial: \textbf{87.692\%}).}
\label{fig:appendix_738_1392}
\end{figure}

% \clearpage

\begin{figure}[h!]
\centering
\begin{subfigure}{\linewidth}
\centering
\includegraphics[trim=0cm 0cm 0cm 0cm, clip, width=0.9\linewidth]{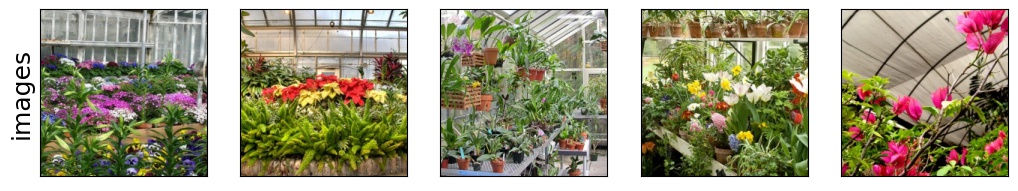}
\end{subfigure}\
\begin{subfigure}{\linewidth}
\centering
\includegraphics[trim=0cm 0cm 0cm 0cm, clip, width=0.9\linewidth]{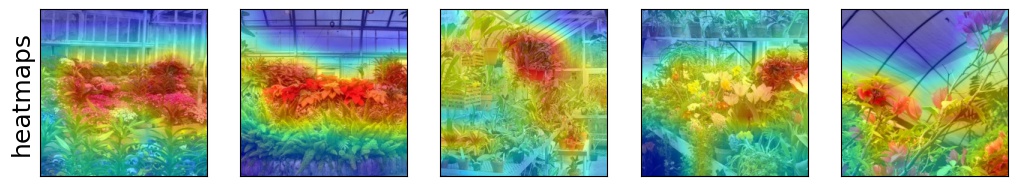}
\end{subfigure}\
\begin{subfigure}{\linewidth}
\centering
\includegraphics[trim=0cm 0cm 0cm 0cm, clip, width=0.9\linewidth]{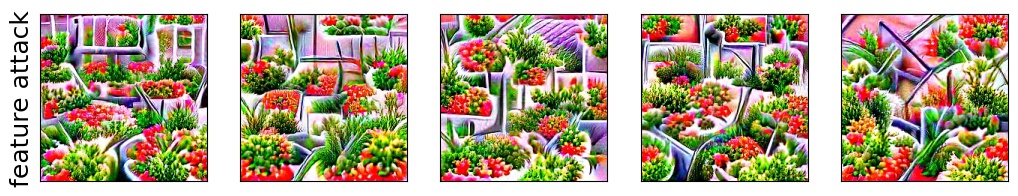}
\end{subfigure}
\caption{Visualization of feature \textbf{1933} for class \textbf{greenhouse} (class index: \textbf{580}).\\ For \textbf{Resnet-50}, accuracy drop: \textcolor{red}{\textbf{-9.231\%}} (initial: \textbf{93.846\%}). For \textbf{Efficientnet-B7}, accuracy drop: \textcolor{blue}{\textbf{+0.0}}\% (initial: \textbf{96.923\%}).\\ For \textbf{CLIP VIT-B32}, accuracy drop: \textcolor{red}{\textbf{-7.692\%}} (initial: \textbf{90.769\%}). For \textbf{VIT-B32}, accuracy drop: \textcolor{red}{\textbf{-7.693\%}} (initial: \textbf{89.231\%}).}
\label{fig:appendix_580_1933}
\end{figure}

% \begin{figure}[h!]
% \centering
% \begin{subfigure}{\linewidth}
% \centering
% \includegraphics[trim=0cm 0cm 0cm 0cm, clip, width=0.9\linewidth]{visualizations/738_1933_images}
% \end{subfigure}\
% \begin{subfigure}{\linewidth}
% \centering
% \includegraphics[trim=0cm 0cm 0cm 0cm, clip, width=0.9\linewidth]{visualizations/738_1933_heatmaps}
% \end{subfigure}\
% \begin{subfigure}{\linewidth}
% \centering
% \includegraphics[trim=0cm 0cm 0cm 0cm, clip, width=0.9\linewidth]{visualizations/738_1933_attacks}
% \end{subfigure}
% \caption{Visualization of feature \textbf{1933} for class \textbf{pot} (class index: \textbf{738}).\\ For \textbf{Resnet-50}, accuracy drop: \textcolor{red}{\textbf{-24.615\%}} (initial: \textbf{93.846\%}). For \textbf{Efficientnet-B7}, accuracy drop: \textcolor{red}{\textbf{-7.692}}\% (initial: \textbf{86.154\%}).\\ For \textbf{CLIP VIT-B32}, accuracy drop: \textcolor{red}{\textbf{-6.154\%}} (initial: \textbf{56.923\%}). For \textbf{VIT-B32}, accuracy drop: \textcolor{red}{\textbf{-10.77\%}} (initial: \textbf{95.385\%}).}
% \label{fig:appendix_738_1933}
% \end{figure}

\clearpage

\begin{figure}[h!]
\centering
\begin{subfigure}{\linewidth}
\centering
\includegraphics[trim=0cm 0cm 0cm 0cm, clip, width=0.9\linewidth]{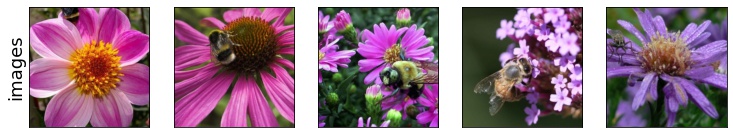}
\end{subfigure}\
\begin{subfigure}{\linewidth}
\centering
\includegraphics[trim=0cm 0cm 0cm 0cm, clip, width=0.9\linewidth]{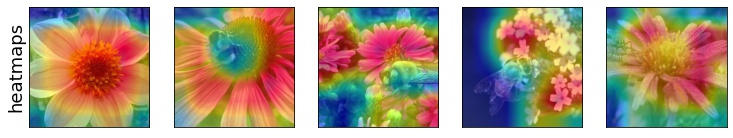}
\end{subfigure}\
\begin{subfigure}{\linewidth}
\centering
\includegraphics[trim=0cm 0cm 0cm 0cm, clip, width=0.9\linewidth]{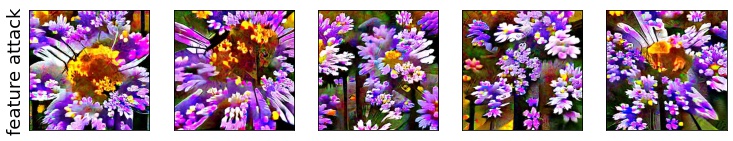}
\end{subfigure}
\caption{Visualization of feature \textbf{595} for class \textbf{bee} (class index: \textbf{309}).\\ For \textbf{Resnet-50}, accuracy drop: \textcolor{red}{\textbf{-41.539\%}} (initial: \textbf{86.154\%}). For \textbf{Efficientnet-B7}, accuracy drop: \textcolor{red}{\textbf{-21.538}}\% (initial: \textbf{76.923\%}).\\ For \textbf{CLIP VIT-B32}, accuracy drop: \textcolor{red}{\textbf{-58.461\%}} (initial: \textbf{93.846\%}). For \textbf{VIT-B32}, accuracy drop: \textcolor{red}{\textbf{-10.769\%}} (initial: \textbf{86.154\%}).}
\label{fig:appendix_309_595}
\end{figure}

\begin{figure}[h!]
\centering
\begin{subfigure}{\linewidth}
\centering
\includegraphics[trim=0cm 0cm 0cm 0cm, clip, width=0.9\linewidth]{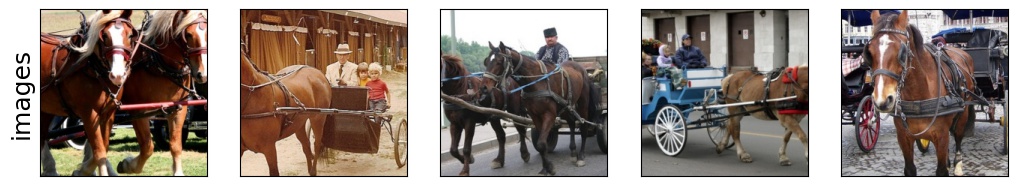}
\end{subfigure}\
\begin{subfigure}{\linewidth}
\centering
\includegraphics[trim=0cm 0cm 0cm 0cm, clip, width=0.9\linewidth]{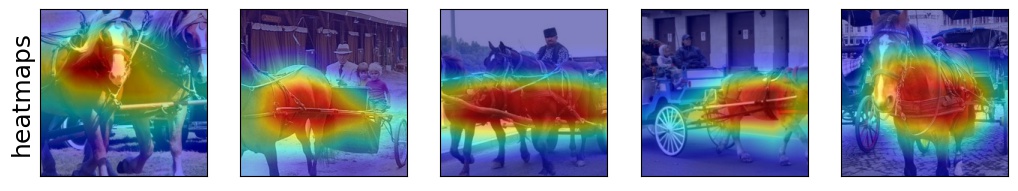}
\end{subfigure}\
\begin{subfigure}{\linewidth}
\centering
\includegraphics[trim=0cm 0cm 0cm 0cm, clip, width=0.9\linewidth]{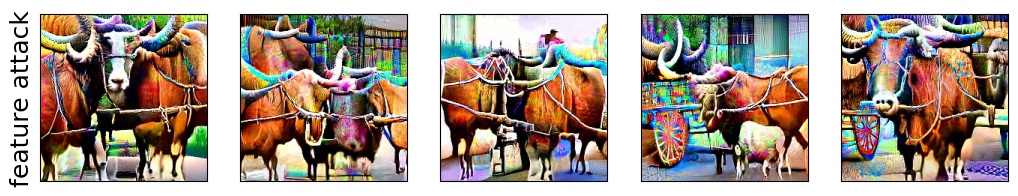}
\end{subfigure}
\caption{Visualization of feature \textbf{1526} for class \textbf{horse cart} (class index: \textbf{603}).\\ For \textbf{Resnet-50}, accuracy drop: \textcolor{red}{\textbf{-3.077\%}} (initial: \textbf{98.462\%}). For \textbf{Efficientnet-B7}, accuracy drop: \textcolor{red}{\textbf{-1.539}}\% (initial: \textbf{98.462\%}).\\ For \textbf{CLIP VIT-B32}, accuracy drop: \textcolor{red}{\textbf{-24.615\%}} (initial: \textbf{87.692\%}). For \textbf{VIT-B32}, accuracy drop: \textcolor{red}{\textbf{-9.231\%}} (initial: \textbf{96.923\%}).}
\label{fig:appendix_603_1526}
\end{figure}

\end{document}